%% file: main.tex
\documentclass[DIV=15, bibliography=totoc]{scrartcl}  

\usepackage[a4paper,margin=2.25cm,bottom=2.5cm]{geometry}
\usepackage{amsfonts}

\usepackage{xurl}

\input{configuration}
\input{00_frontpage}

\begin{document}     

\normalem \maketitle  
\normalfont\fontsize{11}{13}\selectfont

\vspace{-1.5cm} \hrule 

\section*{Abstract}
Neural networks have recently been employed as material discretizations within adjoint optimization frameworks for inverse problems and topology optimization. While advantageous regularization effects and better optima have been found for some inverse problems, the benefit for topology optimization has been limited --- where the focus of investigations has been the compliance problem. We demonstrate how neural network material discretizations can, under certain conditions, find better local optima in more challenging optimization problems, where we here specifically consider acoustic topology optimization. The chances of identifying a better optimum can significantly be improved by running multiple partial optimizations with different neural network initializations.\\
Furthermore, we show that the neural network material discretization's advantage comes from the interplay with the Adam optimizer and emphasize its current limitations when competing with constrained and higher-order optimization techniques. At the moment, this discretization has only been shown to be beneficial for unconstrained first-order optimization.

\vspace{0.25cm}
\noindent \textit{Keywords:} 
topology optimization, acoustics, transfer learning, deep learning, neural networks
\vspace{0.25cm}


\section{Introduction}\label{sec:introduction}

\subsection{Motivation}\label{ssec:motivation}
Although neural networks (NNs)~\cite{goodfellow_deep_2016} are prevalent in both topology optimization~\cite{woldseth_use_2022} and the broader field of computational mechanics~\cite{herrmann_deep_2024}, very few if any robust, yet efficient applications exist that outperform classical methods, as discussed in~\cite{woldseth_use_2022,herrmann_deep_2024}. \\

In topology optimization and inverse problems, the most popular approach is learning a direct mapping between the problem setup or measurements and the quantity of interest. Examples in topology optimization include~\cite{abueidda_topology_2020,zheng_generating_2021,wang_deep_2022,yan_deep_2022} (with extensions using specialized loss functions~\cite{ates_two-stage_2021,luo_improved_2021}, generative NN architectures~\cite{li_non-iterative_2019,yu_deep_2019,behzadi_gantl_2021,herath_topologically_2021}, moving morphable components~\cite{zheng_accurate_2021,hoang_data-driven_2022}, transfer learning~\cite{behzadi_real-time_2021}, and principal components analysis~\cite{ulu_data-driven_2016}). In inverse problems --- in which governing equations or their coefficients are identified from measurements --- notable works consist of~\cite{araya-polo_deep-learning_2018,das_convolutional_2019,mao_subsurface_2019,rao_quantitative_2023}. For various forward problems --- where governing equations are solved given a set of coefficients --- examples 
are~\cite{guo_convolutional_2016,thuerey_deep_2020,chen_towards_2023} for fluid mechanics,~\cite{khadilkar_deep_2019,nie_stress_2020} for solid mechanics, and~\cite{chen_deep_2021} for heat transfer --- where neural operators~\cite{lu_learning_2021,li_fourier_2021,lu_comprehensive_2022} is one of the most recent trends. The most promising situation are inverse problems, as data generation is not as expensive as for topology optimization, where each data point requires a full optimization instead of a single forward solve. Solving forward problems with NNs is less attractive due to competition from efficient forward solvers. However, even for inverse problems, the amount of required data is currently unviable, as, for example, seen in~\cite{bastek_inverse_2023} requiring $53\,007$ data points for a two-dimensional inverse design problem. Exceptions are possible when the problem is reducible, as demonstrated in combination with model order reduction techniques in~\cite{liang_deep_2018,muravleva_application_2018,liang_machine_2018,madani_bridging_2019,bhattacharya_model_2021,derouiche_data-driven_2021}. \\

However, even when the data requirements are neglected, reliability and accuracy of purely data-driven approaches are a major issue, as pointed out in~\cite{woldseth_use_2022,herrmann_deep_2024}. Unless no accurate and efficient alternative exists or accuracy is not required\footnote{An excellent application of NNs in computational mechanics is material modeling, where many constitutive laws are based on empirical fits, which require extensive tuning by the practitioner. NNs being excellent interpolants have therefore shown benefits~\cite{flaschel_unsupervised_2021,klein_polyconvex_2022,linka_automated_2023,linden_neural_2023,benady_nnmcre_2024,benady_unsupervised_2024}, see~\cite{dornheim_neural_2024} for a recent review.}, NNs have limited potential in replacing the entire simulation or optimization chain. However, as argued in~\cite{herrmann_deep_2024}, NNs might be beneficial in replacing part of the simulation or optimization chain. \\

\subsection{Neural Network Ansatz for Optimization Problems}

One such promising current trend is the discretization of the spatial material distribution for inverse problems~\cite{xu_neural_2019,berg_neural_2021,he_reparameterized_2021,zhu_integrating_2022,herrmann_use_2023,jiang_full_2024} and topology optimization~\cite{hoyer_neural_2019,deng_topology_2020,chen_new_2021,chandrasekhar_tounn_2021,chandrasekhar_multi-material_2021,chandrasekhar_approximate_2022,mallon_neural_2024}. Specifically an NN is used to represent the design variables, thus reparametrizing design variables as NN weights. Although the variables representing the material are the same, the number of adjustable parameters can be increased. The potential of this overparametrized discretization is three-fold. Firstly, NNs introduce an implicit regularization, potentially resulting in smoother designs than with classical approaches~\cite{xu_neural_2019,berg_neural_2021,herrmann_use_2023}. Secondly, an overparametrization may be less prone to get stuck in local optima as the chances of all gradient components becoming zero simultaneously decrease. This has, however, yet to be proven. Lastly, prior information can be introduced within the NN. For example, in full waveform inversion, mappings between previous measurements and material distributions are learned through a pretraining and exploited in a subsequent full waveform inversion optimization, yielding an accelerated convergence~\cite{muller_deep_2023,kollmannsberger_transfer_2023,singh_accelerating_2024}. This relies on the concept of transfer learning~\cite{pan_survey_2010,yosinski_how_2014,yan_comprehensive_2024}. However, in the context of topology optimization, the advantages of an NN material discretization have been limited~\cite{chandrasekhar_tounn_2021,mallon_neural_2024}. One reason for this lack of advantage is that the focus has been on the compliance problem~\cite{bendsoe_optimal_1989,sigmund_99_2001}, which is not a challenging optimization problem --- in which ending in premature local optima is not a critical issue and can, in turn, not be improved by an NN parametrization. To this end, we investigate a more challenging optimization problem, \emph{acoustic topology optimization}\footnote{with a direct connection to topology optimization for photonics, see for example~\cite{jensen_topology_2011,christiansen_compact_2021,christiansen_inverse_2021}}~\cite{wadbro_topology_2006,yoon_topology_2007,du_minimization_2007,lee_rigid_2009,kook_acoustical_2012,wadbro_analysis_2014}, which is prone to get stuck in extremely diverse local optima~\cite{duhring_acoustic_2008}. The research question under investigation is whether discretizing the density with an NN can benefit topology optimization. \\

This work will refer to the classical approach, where each design variable is optimized directly, as the \emph{linear ansatz}. When an NN predicts the design variables and the NN parameters, in turn, become the actual tunable parameters, we refer to it as the \emph{NN ansatz}. 

\section{Acoustic Topology Optimization}\label{sec:acoustictopOpt}
\begin{figure}[htb]
    \centering
    \begin{tikzpicture}
    \fill [thin, fill=lightgray] (0,0) rectangle (6,4);

    \fill [thin, fill=gray] (4.5,0.5) rectangle (5.5,1.5); 
    \fill [thin, pattern=crosshatch, pattern color=gray] (0,3.2) rectangle (6,4);

    \fill [red] (1,1) circle (0.1cm);
    \begin{scope}
   \clip (0,0) rectangle (6,4.2);
   \foreach {\r} in {0.3,0.5,...,1.1} {
         \draw [line width=0.2mm, red] (1,1) circle (\r cm);
    }
    \end{scope}

    \node at (5,1) {$\Omega_s$};
    \node at (3,3.6) {$\Omega_c$};
    \node at (3,2) {$\Omega$};

    \draw[thin, |-|, gray] (0,-0.35) -- (1,-0.35);
    \draw[thin, |-|, gray] (-0.35,0) -- (-0.35,1);
    \node [gray] at (0.5,-0.57) {$x_f$};
    \node [gray] at (-0.6,0.5) {$y_f$};

    \draw[thin, |-|, gray] (4.5,1.85) -- (5.5,1.85);
    \draw[thin, |-|, gray] (4.15,0.5) -- (4.15,1.5);
    \node [gray] at (5,2.1) {$a_s$};
    \node [gray] at (3.9,1) {$b_s$};

    \draw [thin, |-|, gray] (0,-0.75) -- (5,-0.75);
    \draw [thin, |-|, gray] (6.35,0) -- (6.35,1);
    \node [gray] at (2.5,-1) {$x_s$};
    \node [gray] at (6.6,0.5) {$y_s$};
    
    \draw[thin, |-|, gray] (-0.25,4) -- (-0.25,3.2);
    \node [gray] at (-0.6,3.6) {$h_c$};

    \draw[thin, |-|, gray] (0,-1.15) -- (6,-1.15);
    \node [gray] at (3,-1.4) {$a$};
    \draw[thin, |-|, gray] (6.85,0) -- (6.85,4);
    \node [gray] at (7.1,2) {$b$};

    \node [red] at (2,2) {$f$};

    \draw[thick, latex-latex] (0,0.8) -- (0,0) -- (0.8,0);
    \node at (-0.18,0.5) {\footnotesize{$y$}};
    \node at (0.5,-0.18) {\footnotesize{$x$}};
        
    \end{tikzpicture}
    \caption{Topology optimization of a ceiling $\Omega_c$ of height $h_c$ in order to suppress the sound pressure in the domain $\Omega_s$ from a harmonic source $s$ at $x_f, y_f$ with frequency $f$. The suppressed domain is rectangular with dimensions $a_s, b_s$ and centered at $x_s, y_s$. The domain $\Omega$ with dimensions $a\times b$, including $\Omega_c$ and $\Omega_s$, is governed by the acoustic wave equation.}\label{fig:setup}
\end{figure}
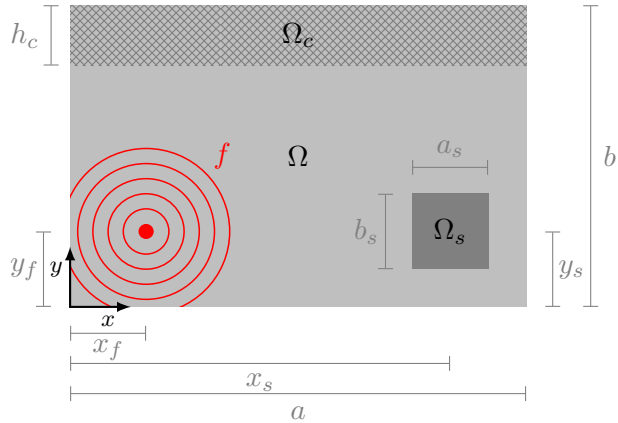

Throughout this treatise, we consider an example from~\cite{duhring_acoustic_2008}, illustrated in \Cref{fig:setup}. Given a harmonic point source $s$, at $(x_f, y_f)$ excited at frequency $f$, a solid material (aluminum) is to be distributed in the ceiling of the domain $\Omega_c$ with height $h_c$. The material is to be distributed such that the sound pressure level $L_p$ in the area $\Omega_s$ a rectangular domain of size $a_s\times b_s$ at $(x_s, y_s)$ is minimal. The sound pressure level $L_p=10 \log_{10}(|p|^2/p_0^2)$ depends on the wave pressures $p$, which are governed by the acoustic wave equation in the rectangular domain $\Omega$ of dimensions $a\times b$ filled by air with a reference pressure $p_0=2\cdot 10^{-6}$ Pa~\cite{jacobsen_fundamentals_2013}. \\

\begin{figure}[htb]
    \centering
    \begin{subfigure}[t]{0.49\textwidth}
		\includegraphics[width=\textwidth]{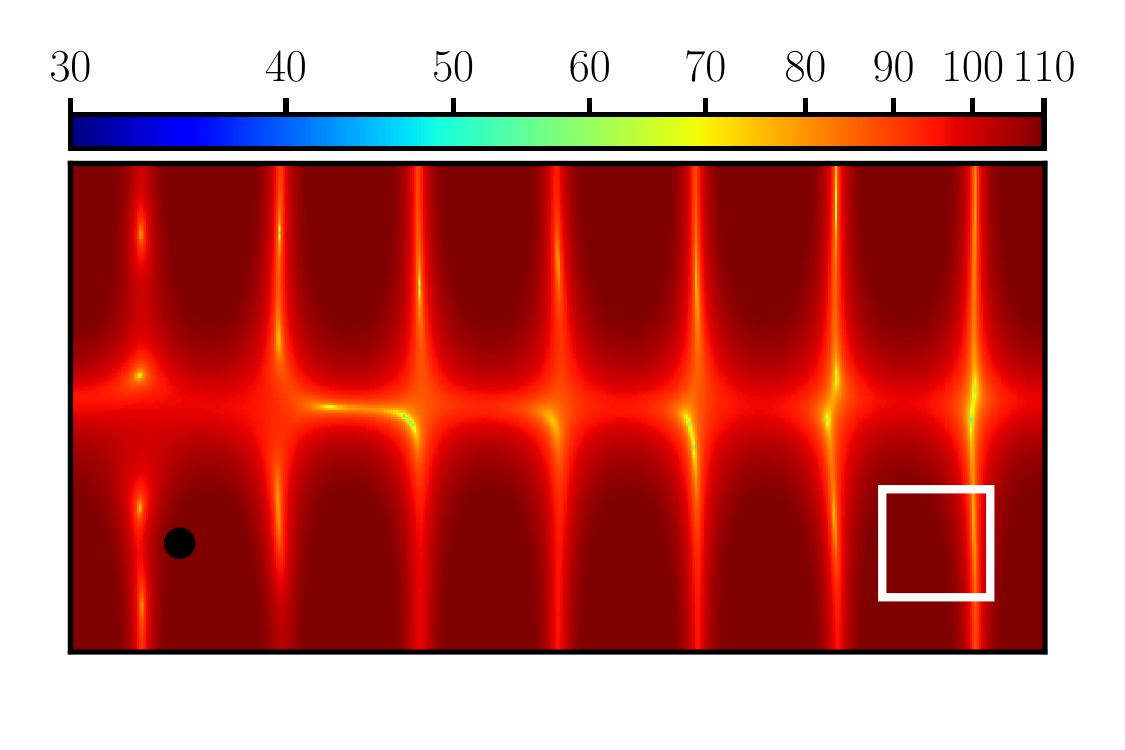}
        \caption{without optimized ceiling, $L_p=110.0$~dB}\label{fig:objective1}
	\end{subfigure}
    \hfill
        \begin{subfigure}[t]{0.49\textwidth}
		\includegraphics[width=\textwidth]{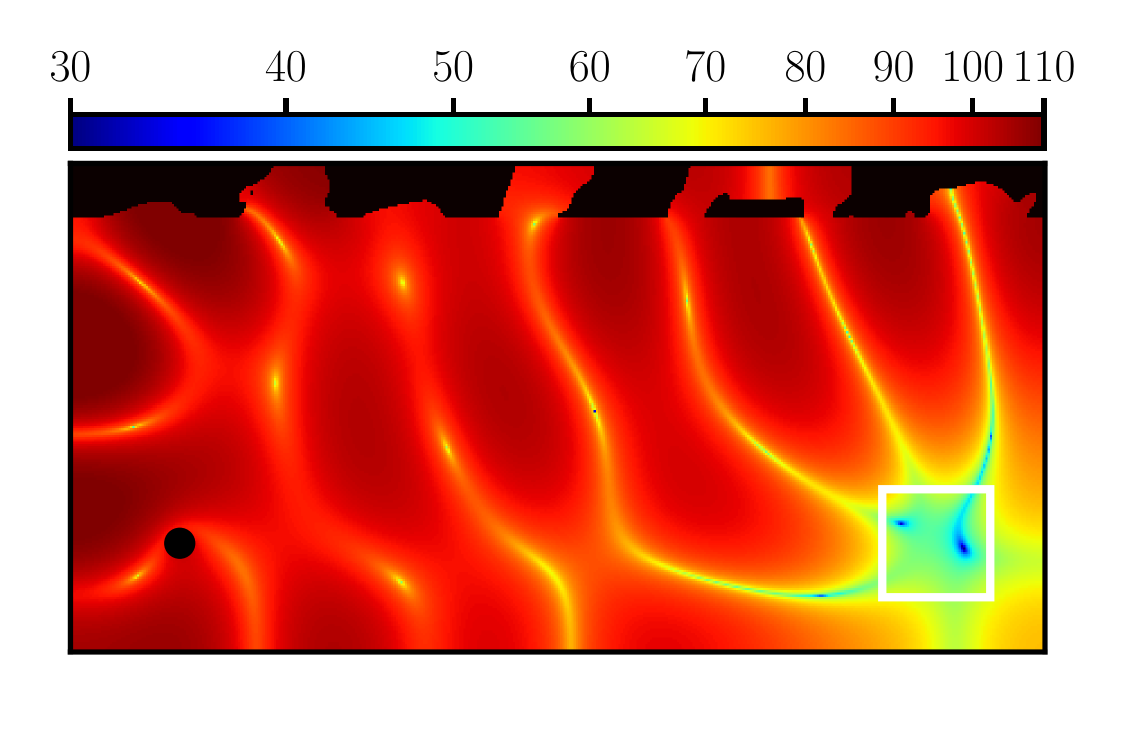}
        \caption{with optimized ceiling, $L_p=61.3$~dB}\label{fig:objective2}
	\end{subfigure}
    \caption{Sound pressure level distribution in dB at $f=69.43$~Hz without and with an optimized ceiling}\label{fig:objective}
\end{figure}

An exemplary optimized design is depicted in~\Cref{fig:objective}, where at a frequency of $f=69.43$~Hz, a sound pressure level of $L_p=110.0$~dB in the domain $\Omega_s$ is reduced to $L_p=61.3$~dB through an optimized ceiling. The dimensions of the problem are summarized in \Cref{tab:problemsetup} and will be used throughout the upcoming investigations unless stated otherwise.

\begin{table}[H]
    \centering
    \caption{Dimensions of the problem setup illustrated in \Cref{fig:setup} and simulated in \Cref{fig:objective}. Dimensions are provided in meters.}\label{tab:problemsetup}
    \begin{tabular}{ccccccccc}
    $a$ & $b$ & $h_c$ & $x_f$ & $y_f$ & $x_s$ & $y_s$ & $a_s$ & $b_s$ \\
    \hline
    18 & 9 & 1 & 2 & 2 & 16 & 2 & 2 & 2 \\
    \hline
    \end{tabular}
\end{table}

\subsection{Physical Model}
The wave pressure field $p$ is governed by the acoustic wave equation consisting of the mass density $\rho$, the bulk modulus $\kappa$, and a source term $s$. Specifically, we consider a normalized acoustic wave equation~\cite{duhring_acoustic_2008} with homogeneous Neumann boundary conditions, i.e., reflecting boundary conditions and mass-proportional damping with coefficient $\eta_d=0.01$~s$^{-1}$.
\begin{equation}
\nabla\cdot(\tilde{\rho}^{-1}\nabla p)-\eta_d\tilde{\kappa}^{-1}\frac{\partial p}{\partial t}-\tilde{\kappa}^{-1}\frac{\partial^2 p}{\partial t^2}=s\label{eq:acousticwaveequation}
\end{equation}
The normalized inverse mass density $\tilde{\rho}^{-1}$ and bulk modulus $\tilde{\kappa}^{-1}$ are interpolated by the indicator $\zeta$ and the corresponding material parameters of air $\rho_1=1.204$~kg/m$^3$, $\kappa_1=1.419 \cdot 10^{5}$~N/m$^{2}$, and the solid material (aluminum) $\rho_2=2643$~kg/m$^3$, $\kappa_2=6.87 \cdot 10^{10}$~N/m$^{2}$~\cite{duhring_acoustic_2008}.
\begin{align}
\tilde{\rho}^{-1}(\zeta)=1+\zeta\left(\frac{\rho_1}{\rho_2}-1\right)\label{eq:parametrization1}\\
\tilde{\kappa}^{-1}(\zeta)=1+\zeta\left(\frac{\kappa_1}{\kappa_2}-1\right)\label{eq:parametrization2}
\end{align}
Thus, air with an indicator of $\zeta=0$, yields $\tilde{\rho}^{-1}(0)=1, \tilde{\kappa}^{-1}(0)=1$, while the solid with an indicator of $\zeta=1$, yields $\tilde{\rho}^{-1}(1)=\rho_1/\rho_2, \tilde{\kappa}^{-1}(1)=\kappa_1/\kappa_2$. \\

Assuming a periodic excitation $s=\hat{s}e^{-i\tilde{\omega}t}$ with the normalized angular frequency $\tilde{\omega}=\omega/\sqrt{\kappa_1/\rho_1}$, \Cref{eq:acousticwaveequation} may be simplified to the Helmholtz equation. The transformation is carried out by assuming a solution of the form $p=\hat{p}e^{-i\tilde{\omega}t}$.
\begin{equation}
\nabla\cdot(\tilde{\rho}^{-1}\nabla\hat{p})+i\tilde{\omega}\eta_d\tilde{\kappa}^{-1}\hat{p}+\tilde{\omega}^2\tilde{\kappa}^{-1}\hat{p}=\hat{s}\label{eq:helmholtz}
\end{equation}
In the sequel, a fixed amplitude of $\hat{s}=10$~Pa$/$m$^2$ is chosen.

\subsection{Discretization}\label{ssec:discretization}
A finite element discretization of the Helmholtz equation from \Cref{eq:helmholtz} yields the system
\begin{equation}
\boldsymbol{S}\boldsymbol{\hat{p}}=\boldsymbol{f},
\end{equation}
with the solution vector $\boldsymbol{\hat{p}}$, the source vector $\boldsymbol{f}$, and the system matrix
\begin{equation}
\boldsymbol{S}=\boldsymbol{K}-i\tilde{\omega}\eta_d\boldsymbol{M}-\tilde{\omega}^2\boldsymbol{M} =\boldsymbol{K}-(i\tilde{\omega}\eta_d+\tilde{\omega}^2)\boldsymbol{M}\label{eq:systemmatrix}
\end{equation}
consisting of stiffness $\boldsymbol{K}$ and mass matrix $\boldsymbol{M}$. We employ high-order integrated Legendre polynomials $\Phi_i(\boldsymbol{\xi})$~\cite{babuska_p_1981,stein__2017} with degree $q$ as finite element ansatz for the scalar solution field $\hat{p}(\boldsymbol{\xi})$ within an individual element $e$.
\begin{equation}
\hat{p}_e(\boldsymbol{\xi})=\sum_{i=1}^{(q+1)^2} \Phi_i(\boldsymbol{\xi})\hat{p}_i    
\end{equation}
The indicator $\zeta$ and thereby the normalized inverse mass density $\tilde{\rho}^{-1}$ and bulk modulus $\tilde{\kappa}^{-1}$ are discretized with $n_v$ subvoxels per dimension within each element, as in the multi-resolution finite cell method~\cite{parvizian_finite_2007,stein__2017}.
Together with the voxel discretization of the indicator $\zeta(\boldsymbol{\xi})$ --- yielding the normalized inverse mass density and bulk modulus voxel values $\tilde{\rho}_v^{-1}, \tilde{\kappa}_v^{-1}$, the local stiffness matrix $\boldsymbol{k}_e$ and local mass matrix $\boldsymbol{m}_e$ can be constructed as
\begin{align}
\boldsymbol{k}_e=\sum_v \tilde{\rho}^{-1}_v \int_{\Omega_v}\boldsymbol{k}(\boldsymbol{\xi})d\Omega_v=\sum_v\tilde{\rho}^{-1}_v\boldsymbol{k}_v, \\
\boldsymbol{m}_e=\sum_v \tilde{\kappa}^{-1}_v \int_{\Omega_v} \boldsymbol{m}(\boldsymbol{\xi})d\Omega_v=\sum_v\tilde{\kappa}^{-1}_e \boldsymbol{m}_v.
\end{align}
This is achieved most efficiently through preintegration~\cite{yang_efficient_2012}, where the stiffness $\boldsymbol{k}(\boldsymbol{\xi})$ and mass matrix integrands $\boldsymbol{m}(\boldsymbol{\xi})$ are integrated separately on each subvoxel domain $\Omega_v$, yielding the stiffness $\boldsymbol{k}_v$ and mass matrix contribution $\boldsymbol{m}_v$ of each subvoxel $v$. Thus leading to only a single local integration valid for all equally sized elements and all subsequent optimization iterations. \\

In the subsequent analyses, we will always consider a material grid discretized by $432\times 216$ voxels. The number of elements (i.e., finite cells) thus depends on the number of subvoxels $n_v$ per element.

\subsection{Optimization}
With the governing physics and a corresponding discretization, the cost function $C$ can be formulated. As introduced in \Cref{fig:setup}, the aim is to minimize the sound pressure level $L_p$ in the domain $\Omega_s$. To this end, the wave pressure is integrated in $\Omega_s$ and normalized by the area of $\Omega_s$, yielding
\begin{equation}
C=\frac{1}{\int_{\Omega_s}d\Omega_s}\int_{\Omega_s}|\hat{p}|^2d\Omega_s.\label{eq:cost}
\end{equation}
No additional constraints are enforced on the optimization. For all cases, the optimization is performed with the first-order optimizer Adam~\cite{kingma_adam_2017} --- with only the learning rate $\alpha$ as a tuned hyperparameter. Adam is for the unconstrained acoustic topology optimization problem assessed to be superior over simple steepest descent for both the linear and NN ansatz, as similarly seen in the context of full waveform inversion~\cite{herrmann_use_2023}. Each gradient-based update is referred to as an epoch within this text, thus adopting the deep learning terminology~\cite{goodfellow_deep_2016}. \\

Furthermore, the high-order multi-resolution discretization poses an additional challenge within the optimization, as previously explored for the compliance problem in~\cite{groen_higherorder_2017}. In the compliance problem, optimized topologies with too low polynomial degrees $q$ given a number of subvoxels $n_v$ suffer from spurious artifacts, rendering the design's discretization dependent and thereby useless. By contrast, artifacts are limited in acoustic topology optimization, as explored in Appendix~\ref{appendix:mtop}. Nevertheless, errors occur, which can often be resolved by a few iterations of a second subsequent optimization with an increased polynomial degree $q$, i.e., acting as a correction of the obtained design. This scheme and its advantage in computational speed are explained in detail in Appendix~\ref{appendix:mtop}. This work performs the optimization with $q=2$, $n_v=4$ for 280 epochs, followed by $q=4, n_v=4$ until the cost function is at maximum $1.5$ larger than after the initial optimization\footnote{This criterion was determined empirically and can surely be improved. The intent is to prevent excessive iterations without improvement. The attained cost function values of the first optimization cannot always be reached by the second optimization. To this end, the cost function threshold is increased by a percentage, which in this work is $50\%$.}. The second optimization is capped by 100 epochs. The final design is evaluated with $q=2, n_v=1$. \\

Additional tools used for the optimization are a polynomial learning rate scheduler --- reducing the learning rate following the rule $(0.2 \cdot \text{epoch} + 1)^{-0.5}$ --- and gradient clipping~\cite{pascanu_difficulty_2013,zhang_why_2020}. Both are only employed when using NNs, as the linear ansatz does not benefit from them.

\subsection{Sensitivity Analysis}
In order to perform the gradient-based optimization, the gradients have to be computed, which can be achieved with an adjoint sensitivity approach, see~\cite{christiansen_compact_2021} for a short, but complete derivation of the adjoint sensitivity approach with the Helmholtz equation. As outlined for acoustic topology optimization~\cite{duhring_acoustic_2008,christiansen_creating_2015} or more generally~\cite{tortorelli_design_1994}, first the adjoint equation is considered
\begin{equation}
\boldsymbol{S}^{\intercal}\boldsymbol{\hat{p}}^{\dagger}=-\left(\frac{\partial C}{\partial \boldsymbol{\hat{p}}_R} -i\frac{\partial C}{\partial \boldsymbol{\hat{p}}_I} \right)^{\intercal}\label{eq:adjoint},
\end{equation}
where $\boldsymbol{\hat{p}}_R$ and $\boldsymbol{\hat{p}}_I$ are correspondingly the real and imaginary part of $\boldsymbol{\hat{p}}$ and $\boldsymbol{\hat{p}}^{\dagger}$ is the corresponding adjoint solution. The $k^{\textrm{th}}$ entry of the right-hand side of \Cref{eq:adjoint} is given by
\begin{equation}
\left(\frac{\partial C}{\partial \boldsymbol{\hat{p}}_R}-i\frac{\partial C}{\partial \boldsymbol{\hat{p}}_I} \right)_{k} = \frac{1}{\int_{\Omega_s}d\Omega_s}\int_{\Omega_s}2(\boldsymbol{\hat{p}}_R-i\boldsymbol{\hat{p}}_I)_{k}N_{k}d\Omega_s.
\end{equation}
By solving \Cref{eq:adjoint} with the same discretization as described in \Cref{ssec:discretization}, the adjoint solution $\boldsymbol{\hat{p}}^{\dagger}$ is obtained. With the adjoint solution, the sensitivity of the cost function $C$ with respect to the indicator voxel values $\zeta_v$ can be computed as
\begin{equation}
\frac{dC}{d\zeta_v}=\frac{\partial C}{\partial \zeta_v}+\Re\left(\boldsymbol{\hat{p}}^{\dagger}\frac{\partial \boldsymbol{S}}{\partial \zeta_v}\boldsymbol{\hat{p}}\right)=\Re\left(\boldsymbol{\hat{p}}^{\dagger}\frac{\partial \boldsymbol{S}}{\partial \zeta_v}\boldsymbol{\hat{p}}\right),\label{eq:intermediateSensitivity}
\end{equation}
as described in~\cite{duhring_acoustic_2008,christiansen_creating_2015}, where $\Re$ extracts the real part. \Cref{eq:intermediateSensitivity} requires the system matrix sensitivity with respect to the indicator voxel values $\partial \boldsymbol{S}/\partial \zeta_v$, which can be computed with the stiffness matrix derivative with respect to the normalized inverse mass density $\partial\boldsymbol{K}/\partial\tilde{\rho}^{-1}_v=\boldsymbol{k}_v$, and the mass matrix derivative with respect to the normalized inverse bulk modulus $\partial\boldsymbol{M}/\partial \tilde{\kappa}^{-1}_v=\boldsymbol{m}_v$. This yields\footnote{Note that the sensitivity computation becomes more involved if static condensation is performed on the internal modes of the higher-order shape functions, as seen in~\cite{groen_higherorder_2017}.} 
\begin{equation}
\frac{\partial\boldsymbol{S}}{\partial \zeta_v}=\frac{\partial \boldsymbol{K}}{\partial \tilde{\rho}_v^{-1}}\frac{\partial\tilde{\rho}_v^{-1}}{\partial\zeta_v}-(i\tilde{\omega}\eta_d+\tilde{\omega})\frac{\partial \boldsymbol{M}}{\partial \tilde{\kappa}_v^{-1}}\frac{\partial\tilde{\kappa}^{-1}_v}{\partial\zeta_v}. \label{eq:systemsensitivity}
\end{equation}
Inserting \Cref{eq:systemsensitivity} into \Cref{eq:intermediateSensitivity} results in the final expression of the cost function gradient with respect to the indicator voxel values.
\begin{equation}
\frac{d C}{d\zeta_v}=\Re\left(\boldsymbol{\hat{p}}^{\dagger}\left[\left(\frac{\rho_1}{\rho_2}-1\right)\boldsymbol{k}_v-(i\tilde{\omega}\eta_d+\tilde{\omega}^2)\left(\frac{\kappa_1}{\kappa_2}-1\right)\boldsymbol{m}_v\right]\boldsymbol{\hat{p}}\right)\label{eq:finalSensitivity}
\end{equation}

\subsection{Filtering and Projection}\label{ssec:filtering}
In order to avoid mesh-dependent solutions with too small and, thereby, unmanufacturable features, filtering is employed~\cite{sigmund_numerical_1998}. Specifically, a density filter with linear decay~\cite{bruns_topology_2001,bourdin_filters_2001} is employed\footnote{Note that acoustic topology optimization does not typically yield robust designs~\cite{christiansen_creating_2015}. This can be ensured with approaches discussed in~\cite{sigmund_morphology-based_2007,sigmund_manufacturing_2009,christiansen_creating_2015}. This is, however, neglected in this work.}. The filtered indicators are given as 
\begin{equation}
\tilde{\zeta}_v=\frac{\sum_{k\in N_v}w(\boldsymbol{x}_v-\boldsymbol{x}_k)\zeta_k}{\sum_{k\in N_v} w(\boldsymbol{x_e}-\boldsymbol{x}_k)},\label{eq:filter}
\end{equation}
where $N_v$ is the neigborhood of voxel $v$, i.e., $N_v=\{k\text{ }|\text{ }||\boldsymbol{x}_k-\boldsymbol{x}_v||<r_f\}$. The linear decay is enforced through the weighting function $w(\boldsymbol{x})$ with the filter radius $r_f$.
\begin{equation}
w(\boldsymbol{x})=
\begin{cases} 
r_f - ||\boldsymbol{x}|| & \text{if } ||\boldsymbol{x}|| \leq r_f \\
0 & \text{else}
\end{cases}
\end{equation}
A filter radius of two voxels is employed, i.e., $r_f=2\cdot b / 216\approx 0.083$~m. As the filtering smears the designs and sharp designs are sought-after, a subsequent projection ensures a near $0/1$- design~\cite{guest_achieving_2004,wang_projection_2011,li_volume_2015}. To this end, a smooth approximation of the Heaviside function $H(\tilde{\zeta})$ --- allowing differentiation --- is employed~\cite{wang_projection_2011}, yielding projected indicator values $\bar{\tilde{\zeta}}$ after thresholding $\tilde{\zeta}$.
\begin{equation}
\bar{\tilde{\zeta}}=H(\tilde{\zeta})=\frac{\tanh(\beta \eta)+\tanh(\beta (\tilde{\zeta}-\eta))}{\tanh(\beta \eta)+\tanh(\beta (1-\eta))}\label{eq:projection}
\end{equation}
Here $\eta$ shifts the thresholding, while $\beta$ controls the sharpness. To allow a proper gradient flow from the projected indicator values $\bar{\tilde{\zeta}}$ to the indicator values $\zeta$, $\beta$ is initially kept small and gradually increased~\cite{guest_achieving_2004,wang_projection_2011,christiansen_creating_2015}. Specifically, we initialize $\beta=1$, and increase it by $2\%$ per epoch and limit it to $\beta_{\text{max}}=75$ during the first optimization at $q=2, n_v=4$ and to $\beta_{\text{max}}=150$ for the second optimization at $q=4, n_v=4$. The threshold is set to $\eta=1/2$. \\

The mass density and bulk modulus parametrization from \Cref{eq:parametrization1,eq:parametrization2} thus rely on the filtered and projected indicator $\bar{\tilde{\zeta}}$, which in turn is also considered during the sensitivity computation with \Cref{eq:finalSensitivity}. As, the framework is implemented in PyTorch~\cite{paszke_pytorch_2019}, the sensitivity $\partial \bar{\tilde{\zeta}}/\partial \zeta$ can be obtained through automatic differentiation~\cite{baydin_automatic_2017}. The computational cost of this operation is negligible. 

\section{Neural Network Ansatz}

Before introducing the NN ansatz for acoustic topology optimization (\Cref{ssec:NNansatzTopOpt}), the motivation for using an NN ansatz is clarified on simple optimization benchmark problems (\Cref{ssec:motivationNN}). 

\subsection{Motivation for a Neural Network Ansatz}\label{ssec:motivationNN}
Consider the two-dimensional Rosenbrock function
\begin{equation}
f(x,y)=(1-x)^2 + 100 (y-x^2)^2\label{eq:rosenbrock}
\end{equation}
with $x$ and $y$ as design variables. The function's minimum is at $x=1, y=1$. Given an initial guess $(x_0, y_0)$, the design variables can be updated via gradient descent schemes to decrease $f$. For this, first-order optimization methods rely on the gradients $\frac{\partial f}{\partial x}$ and $\frac{\partial f}{\partial y}$. \\

An NN parametrization, i.e., the NN ansatz, can be introduced to represent $x$ and $y$. To be more precise, an NN $a(\boldsymbol{z};\boldsymbol{\varTheta})$ parametrized by the NN weights $\boldsymbol{\varTheta}$ predicts $\hat{x}$ and $\hat{y}$ from an input $\boldsymbol{z}$, where the hat symbol indicates an NN prediction.
\begin{equation}
    \begin{bmatrix} \hat{x}\\ \hat{y} \end{bmatrix} = a(\boldsymbol{z}; \boldsymbol{\varTheta})
\end{equation}
Instead of updating $\hat{x}$ and $\hat{y}$ directly during an optimization, the NN weights $\boldsymbol{\varTheta}$ are updated utilizing the gradients $\frac{\partial f}{\partial\boldsymbol{\varTheta}}=\frac{\partial f}{\partial \hat{x}}\cdot \frac{\partial \hat{x}}{\partial \boldsymbol{\varTheta}} + \frac{\partial f}{\partial \hat{y}}\cdot \frac{\partial \hat{y}}{\partial \boldsymbol{\varTheta}}$ obtained with the chain rule. This parametrization allows for the increase of the number of tunable parameters, i.e., an overparametrization. This can reduce the chances of getting stuck in a local optimum, as more gradients need to be zero than when directly optimizing $x, y$. \\

We demonstrate the potential of the NN ansatz through a comparison with the linear ansatz, i.e., treating $x, y$ directly as design variables on the benchmark problem from \Cref{eq:rosenbrock}. To this end, a fully connected NN predicts $\hat{x}, \hat{y}$ from 10 random inputs $\boldsymbol{z}$ sampled from a uniform distribution in the range $[-1,1]$. One hidden layer composed of 50 neurons is selected, determined by tuning the architecture. This yields 625 instead of two design variables. From an initial guess\footnote{Different initial guesses were tested with similar outcomes.} $x_0=3, y_0=3$ the two approaches are compared, where different NN instances are averaged --- possible due to different NN weight initializations. To guarantee the same start value, the first NN output is shifted such that $\hat{x}_0=3, \hat{y}_0=3$. In both cases, the first-order optimizer Adam~\cite{kingma_adam_2017} is employed for 300 epochs, similar to the works utilizing NN material discretizations~\cite{xu_neural_2019,berg_neural_2021,he_reparameterized_2021,zhu_integrating_2022,herrmann_use_2023,jiang_full_2024,hoyer_neural_2019,deng_topology_2020,chen_new_2021,chandrasekhar_tounn_2021,chandrasekhar_multi-material_2021,chandrasekhar_approximate_2022,mallon_neural_2024}. The corresponding learning rate $\alpha$ is tuned through a grid search, which only performs 20 epochs to assess the quality of an optimization. \\

\begin{figure}[htb]
    \centering
    \begin{subfigure}[t]{0.49\textwidth}
		\includegraphics[width=\textwidth]{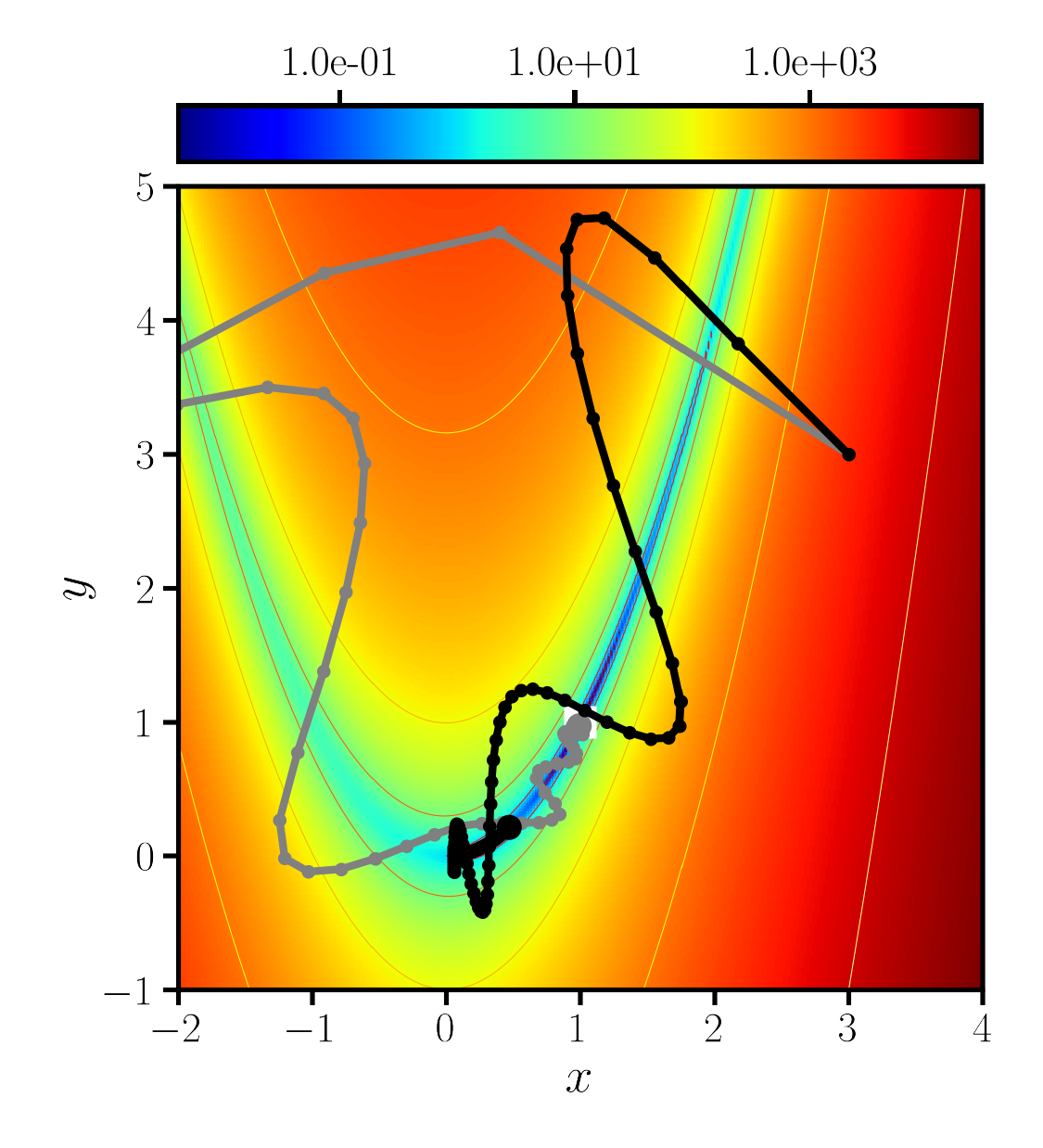}
		\caption{Optimization path}\label{fig:RosenbrockAdam1}
	\end{subfigure}
    \hfill
    \begin{subfigure}[t]{0.49\textwidth}
		\includegraphics[width=\textwidth]{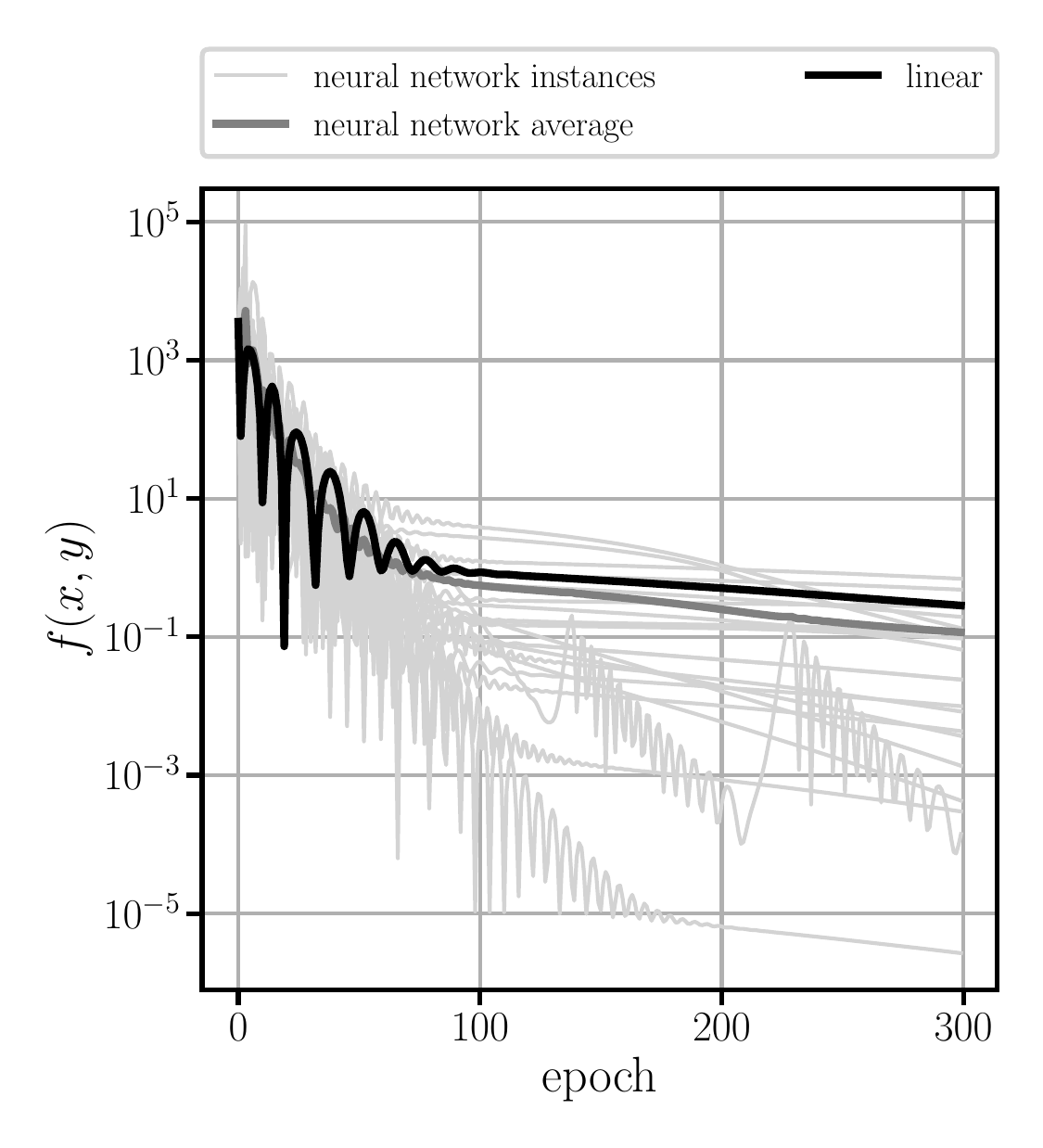}
		\caption{Cost histories}\label{fig:RosenbrockAdam2}
	\end{subfigure}
    \caption{Rosenbrock function optimized with NN ansatz and directly on the coordinates (linear ansatz) using \textbf{Adam}}\label{fig:RosenbrockAdam}
\end{figure}

\Cref{fig:RosenbrockAdam1} compares the optimization paths along the Rosenbrock landscape, in which the NN ansatz is indicated in gray and the linear ansatz in black. For this NN instance, an advantage is apparent. However, when considering 20 NN instances obtained with different NN weight initializations, the NN is not always superior, as seen in \Cref{fig:RosenbrockAdam2}. The average performance over the 20 instances is only slightly improved, although many NN instances are significantly better. This point is crucial because it demonstrates that better solutions can often be found using an indirect design variable parametrization rather than optimizing the design variables directly. \\

\begin{figure}[htb]
    \centering
    \begin{subfigure}[t]{0.49\textwidth}
		\includegraphics[width=\textwidth]{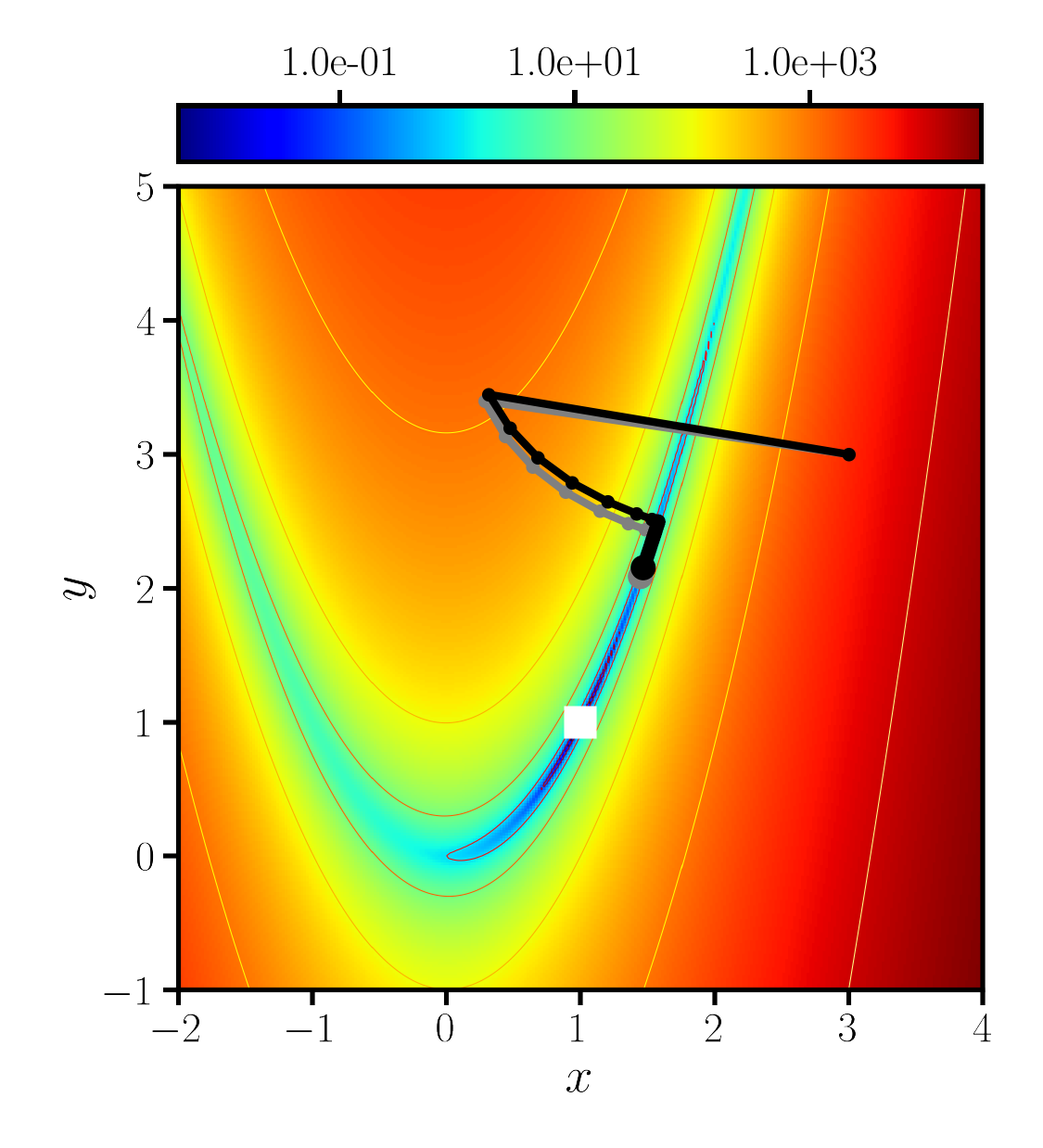}
		\caption{Optimization paths}\label{fig:RosenbrockSteepest1}
	\end{subfigure}
    \hfill
    \begin{subfigure}[t]{0.49\textwidth}
		\includegraphics[width=\textwidth]{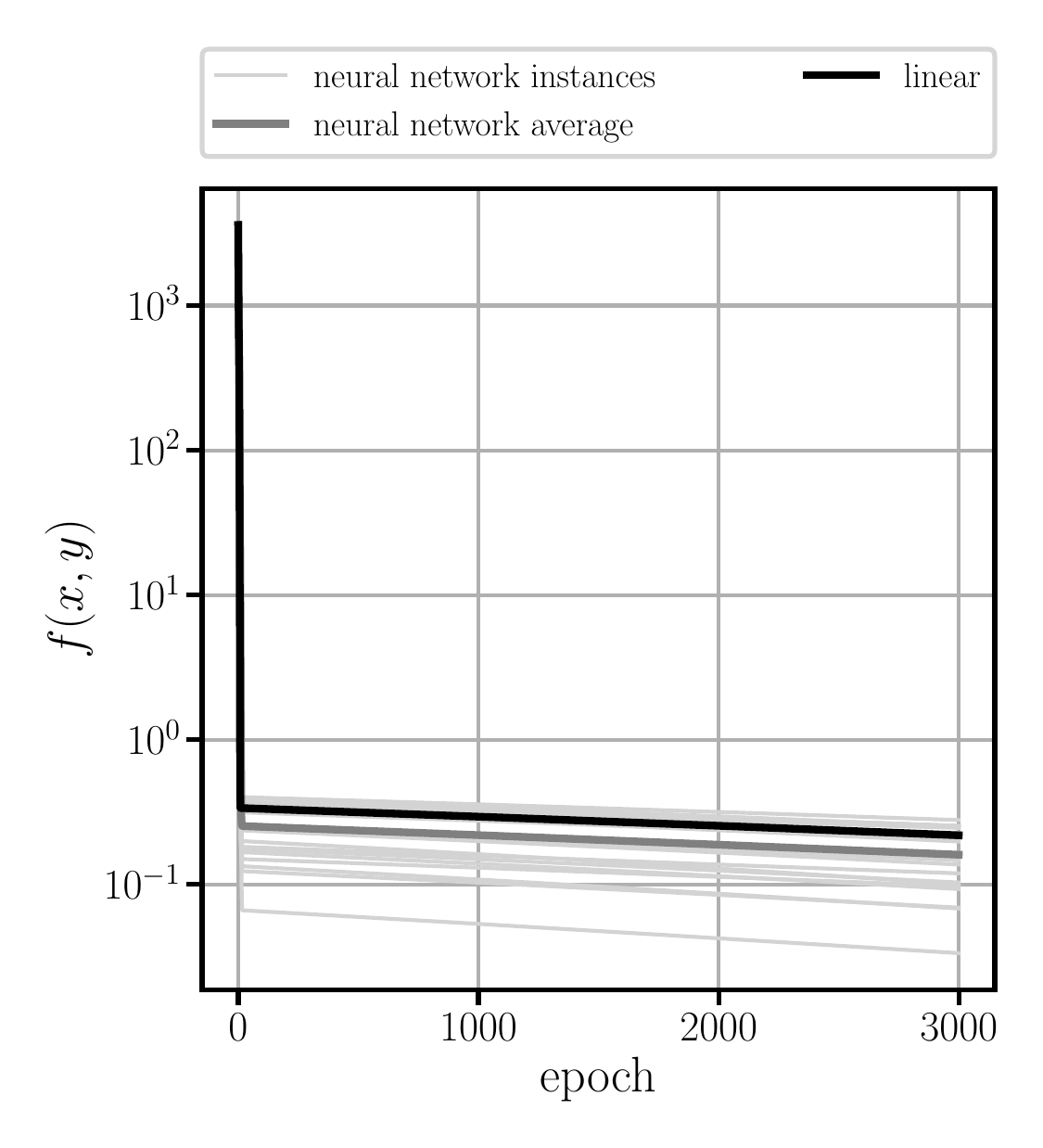}
		\caption{learning histories}\label{fig:RosenbrockSteepest2}
	\end{subfigure}
    \caption{Rosenbrock function optimized with NN ansatz and directly on the coordinates (linear ansatz) using \textbf{steepest descent}}\label{fig:RosenbrockSteepest}
\end{figure}

Changing the optimizer reduces the benefit. Steepest descent does not exhibit a clear advantage, as shown in \Cref{fig:RosenbrockSteepest}. Also, using second-order optimizers diminishes the differences, which was tested with L-BFGS~\cite{nocedal_updating_1980}. A further disadvantage of only working indirectly with the true quantities of interest, i.e., $x, y$, arises when enforcing constraints. Popular constrained optimizers in topology optimization, such as the optimality criterion methods (see, e.g.,~\cite{bendsoe_topology_2003}), the method of moving asymptotes~\cite{svanberg_method_1987}, or the globally convergent method of moving asymptotes~\cite{svanberg_class_2002}, do not provide an update direction of the design variables but an exact update of them. Currently, only directions can be transferred to the NN weights $\boldsymbol{\varTheta}$. Projecting updated design variables $x, y$ to the NN weights would require an additional non-linear optimization due to the non-linear nature of the NN. Thus, the potential advantage of an NN ansatz reduces for constrained optimization. For the compliance problem, this challenge has been tackled by enforcing the constraints via penalty terms~\cite{chandrasekhar_tounn_2021,chandrasekhar_multi-material_2021,chandrasekhar_approximate_2022,mallon_neural_2024}. \\

Another limitation is that the NN ansatz's benefit is highly problem-dependent. To illustrate this, we consider three additional optimization benchmarks, the Ratrigrin, Ackley, and Lévi functions (see Appendix~\ref{appendix:optimization} for details). The optimization's success is most sensitive to the choice of initial guess. In order to gain a statistical understanding of the method's benefit, $4\,000$ optimizations with different initial guesses are performed for each of the benchmarks. The outcome is quantified by considering the percentage of the NN ansatz being superior over the linear ansatz for a given problem. The results are summarized in \Cref{tab:NNansatzstatistics} for five different NN architectures. Only the number of hidden neurons is varied from 25 to 400 neurons, yielding 327 (for 25 neurons), 652 (for 50 neurons), $1\,302$ (for 100 neurons), $2\,602$ (for 200 neurons), and $5\,202$  (for 400 neurons) trainable parameters. The best performances are indicated with gray highlights in \Cref{tab:NNansatzstatistics} and are between 50 and 100 neurons. 

\begin{table}[htb]
    \centering
    \caption{Percentage of NN ansatz successfully outperforming the linear ansatz --- averaged over $4\,000$ initial guesses with a single NN initialization and optimized learning rates $\alpha$ over 10 tuning epochs. The choice of initial guess is much more sensitive than the NN initialization.}\label{tab:NNansatzstatistics}
    \begin{tabular}{lccccc}
    & \textbf{25 neurons} & \textbf{50 neurons} & \textbf{100 neurons} & \textbf{200 neurons} & \textbf{400 neurons} \\
    \hline
    \textbf{Rosenbrock} & $74.8 \%$ & $77.5 \%$ & \cellcolor{lightgray}$79.5 \%$ & $78.2 \%$ & $70.9 \%$ \\
    \hline
    \textbf{Rastrigrin} & $53.2 \%$ & $ 52.0 \%$ & \cellcolor{lightgray}$56.6 \%$ & $52.7 \%$ & $47.7 \%$ \\
    \hline
    \textbf{Ackley} & $41.3 \%$ & $43.4 \%$ & \cellcolor{lightgray}$44.0 \%$ & $34.0 \%$ & $26.8 \%$ \\
    \hline
    \textbf{Lévi} & $48.5 \%$ & \cellcolor{lightgray}$49.1 \%$ & $47.9 \%$ & $42.6 \%$ & $37.1 \%$ \\
    \hline
    \end{tabular}
\end{table}

The exact reason for the deterioration of the NN ansatz success observed in \Cref{tab:NNansatzstatistics} for the Rastrigrin, Ackley, and Lévi function is unclear. However, the question arises of whether an improved, problem-specific NN weight initialization can improve performance. Currently, a generic He initialization~\cite{he_delving_2015} is employed, whose direct aim is to ensure a good gradient flow through the NN. Transfer learning~\cite{pan_survey_2010,yosinski_how_2014,yan_comprehensive_2024} can be seen as a problem-specific initialization, as employed in many computer vision tasks~\cite{simonyan_very_2015,he_deep_2016}, which are pretrained on state-of-the-art NNs such as ImageNet~\cite{krizhevsky_imagenet_2012}. However, transfer learning has also successfully been applied to full waveform inversion~\cite{muller_deep_2023,kollmannsberger_transfer_2023,singh_accelerating_2024}. In transfer learning, before applying the NN to a problem, it is pretrained on similar problems, thus improving convergence. As this task is highly problem-dependent, the effect of transfer learning will be studied on the acoustic topology optimization problem in the following sections. 

\subsection{Neural Network Ansatz for Acoustic Topology Optimization}\label{ssec:NNansatzTopOpt}
Acoustic topology optimization will be performed on the setup described in~\Cref{fig:setup} with the dimensions from \Cref{tab:problemsetup}. The optimization is performed for distinct individual frequencies. Similarly to~\cite{duhring_acoustic_2008,christiansen_creating_2015}, resonance frequencies are considered, where eight frequencies: 21.33~Hz, 34.39~Hz, 57.22~Hz, 69.43~Hz, 76.30~Hz, 95.37~Hz, 141.78~Hz, and 166.56~Hz were selected. The frequencies were determined as described in Appendix~\ref{appendix:frequencies}. As an NN architecture, we employ a U-net with $148\,806$ trainable parameters (see Appendix~\ref{appendix:NN} for details). To put this number into context, the number of design variables, i.e., voxels in the ceiling $\Omega_c$, is $432\times 24=10\,368$.\\

Before proceeding, the motivation of acoustic topology optimization to investigate the possibilities of an NN ansatz is clarified (\Cref{ssec:motivationAcousticTopOpt}) --- namely, the potential of NNs to identify better designs than classically obtainable. This is followed by highlighting how the NN struggles with filtering and projection (\Cref{ssec:outsmarting}) and how this can be overcome by transfer learning (\Cref{sssec:pretraining}).

\subsubsection{Acoustic Topology Optimization and Neural Networks}\label{ssec:motivationAcousticTopOpt}
Using NNs in acoustic topology optimization is a favorable combination mainly for two reasons. Firstly, an unconstrained optimization problem can be considered (\Cref{eq:cost}) --- and secondly, the optimized topologies are highly sensitive to the initial guess. Thus, different optimization runs yield very different local optima, i.e., different final designs, which can potentially be improved through smart initial guesses by the NN in order to consistently reach good local optima. \\

\begin{figure}[htb]
	\centering
	\begin{subfigure}[t]{0.49\textwidth}
		\includegraphics[width=\textwidth]{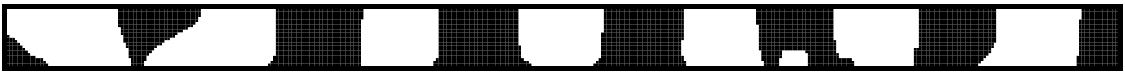}
		\caption{$f=69.43$~Hz and $\boldsymbol{\zeta}=0$ yields $L_p=82.2$~dB}\label{fig:sensitiveDesigns1}
	\end{subfigure}
	\hfill
	\begin{subfigure}[t]{0.49\textwidth}
		\includegraphics[width=\textwidth]{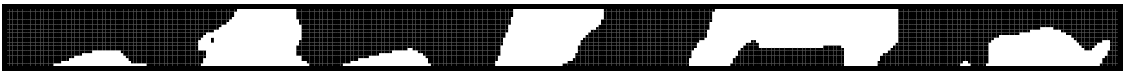}
		\caption{$f=69.43$~Hz and $\boldsymbol{\zeta}=1$ yields $L_p=61.3$~dB}\label{fig:sensitiveDesigns2}
	\end{subfigure}
    \\
	\begin{subfigure}[t]{0.49\textwidth}
		\includegraphics[ width=\textwidth]{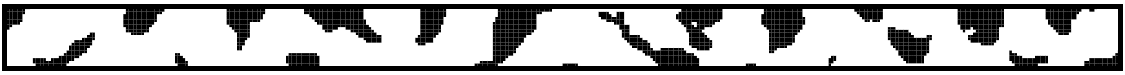}
		\caption{$f=166.56$~Hz and $\boldsymbol{\zeta}=0$ yields $L_p=77.87$~dB}\label{fig:sensitiveDesigns3}
	\end{subfigure}
	\hfill
	\begin{subfigure}[t]{0.49\textwidth}
		\includegraphics[width=\textwidth]{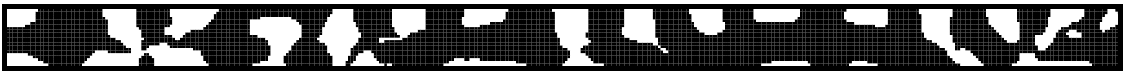}
		\caption{$f=166.56$~Hz and $\boldsymbol{\zeta}=1$ yields $L_p=80.52$~dB}\label{fig:sensitiveDesign4}
	\end{subfigure}
    \caption{Optimized structure after 300 epochs (280 with $q=2, n_v=4$ and 20 with $q=4, n_v=4$) with the corresponding sound pressure level $L_p$ in the to-be-suppressed area}\label{fig:sensitiveDesigns}
 \end{figure}

This sensitivity to the initial guess is illustrated in \Cref{fig:sensitiveDesigns}, where the final design exhibits a completely different performance when changing the initial indicator guess from $\boldsymbol{\zeta}=0$ to $\boldsymbol{\zeta}=1$, as seen for $f=69.43$~Hz and $f=206.31$~Hz.

\subsubsection{Bypassing of Filtering and Projection}\label{ssec:outsmarting}
\begin{figure}[htb]
	\centering
     \begin{subfigure}[t]{0.49\textwidth}
		\includegraphics[width=\textwidth]{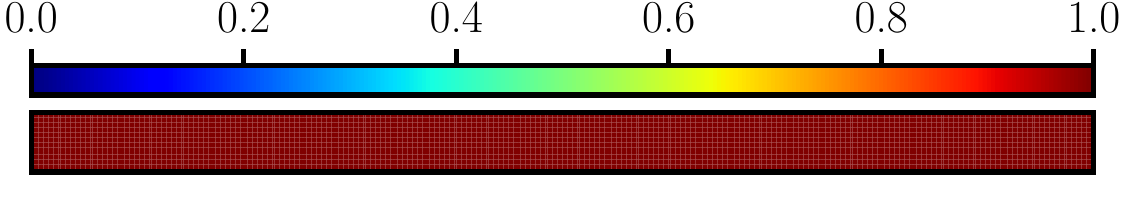}
		\caption{Linear ansatz initial guess}\label{fig:outsmarting1}
	\end{subfigure}
	\hfill
    \begin{subfigure}[t]{0.49\textwidth}
		\includegraphics[width=\textwidth]{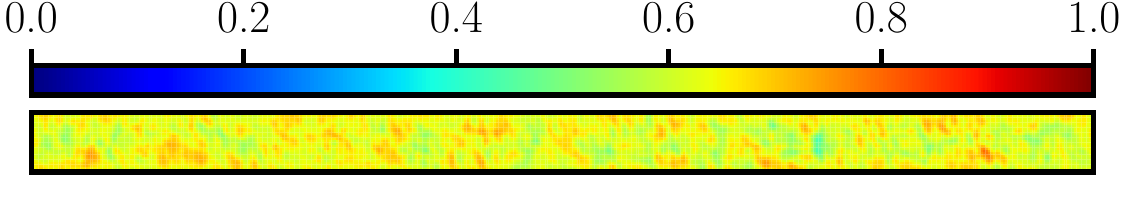}
		\caption{NN initial guess}\label{fig:outsmarting2}
	\end{subfigure}
	\begin{subfigure}[t]{0.49\textwidth}
		\includegraphics[width=\textwidth]{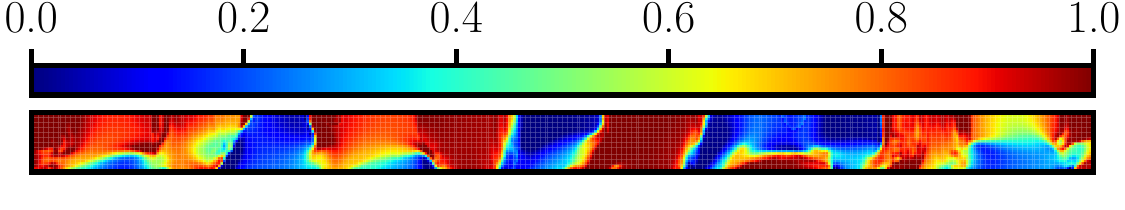}
		\caption{Linear ansatz before filtering and projection}\label{fig:outsmarting3}
	\end{subfigure}
	\hfill
    \begin{subfigure}[t]{0.49\textwidth}
		\includegraphics[width=\textwidth]{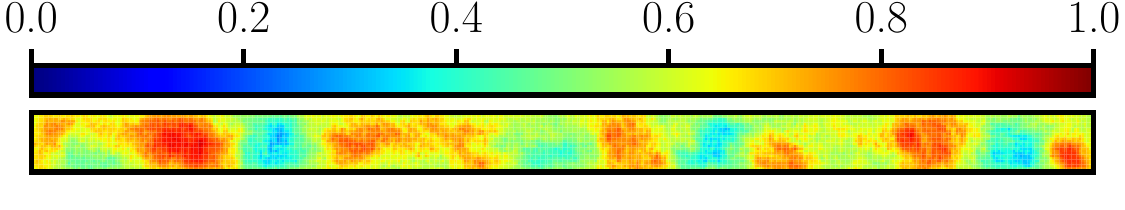}
		\caption{NN before filtering and projection}\label{fig:outsmarting4}
	\end{subfigure}
	\\
 	\begin{subfigure}[t]{0.49\textwidth}
		\includegraphics[width=\textwidth]{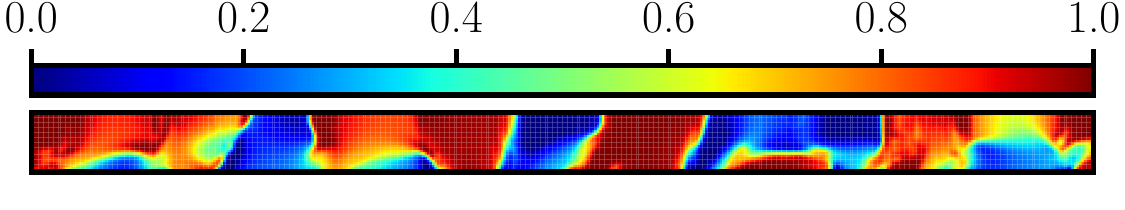}
		\caption{Linear ansatz after filtering, before projection}\label{fig:outsmarting5}
	\end{subfigure}
	\hfill
    \begin{subfigure}[t]{0.49\textwidth}
		\includegraphics[width=\textwidth]{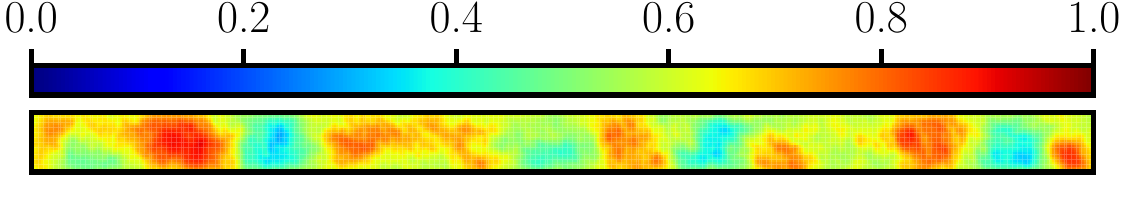}
		\caption{NN after filtering, before projection}\label{fig:outsmarting6}
	\end{subfigure}
    \\
    \begin{subfigure}[t]{0.49\textwidth}
		\includegraphics[width=\textwidth]{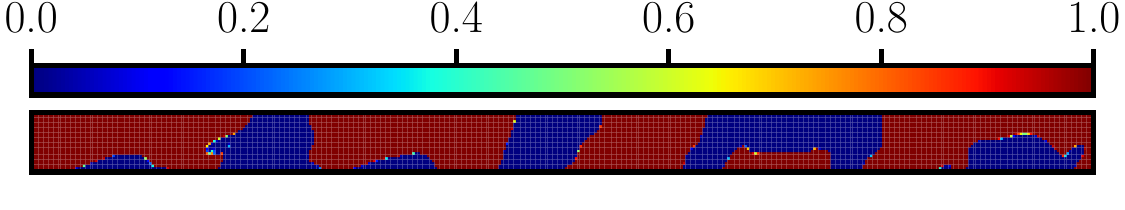}
		\caption{Linear ansatz after filtering and projection}\label{fig:outsmarting7}
	\end{subfigure}
	\hfill
    \begin{subfigure}[t]{0.49\textwidth}
		\includegraphics[width=\textwidth]{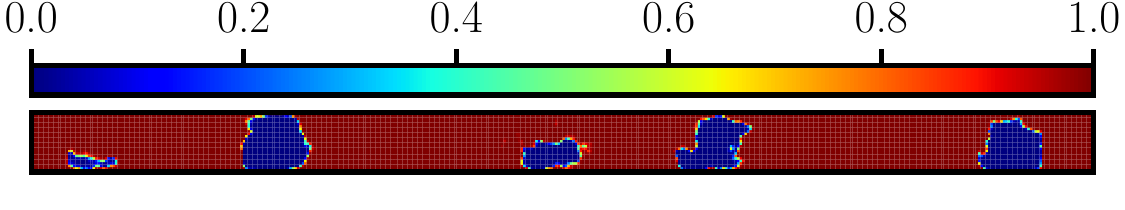}
		\caption{NN after filtering and projection}\label{fig:outsmarting8}
	\end{subfigure}
    \caption{Illustration of the bypassing of the filtering and projection. The sound pressure level at $f=69.43$~Hz before and after thresholding for the linear ansatz: $L_p^{\text{before}}=60.22$~dB, $L_p^{\text{after}}=60.33$~dB; and for the NN ansatz: $L_p^{\text{before}}=59.88$~dB, $L_p^{\text{after}}=70.09$~dB}\label{fig:outsmarting}
\end{figure}

However, simply exchanging the linear ansatz with the NN ansatz in the described topology optimization framework (\Cref{sec:acoustictopOpt}) yields unsatisfactory results. \Cref{fig:outsmarting} compares an optimization between the two approaches for $f=69.43$~Hz. For simplicity, the optimization is conducted only with $q=2, n_v=4$, i.e., without the second optimization, acting as correction, described in Appendix~\ref{appendix:mtop}. Manufacturing restrictions require filtering and projection as described in~\Cref{ssec:filtering}. However, projecting the final design with the NN ansatz, seen in~\Cref{fig:outsmarting8}, yields a design that is not sharp and does not obey the filtering length scale $r_f$. The problem is that the filtering and projection are intended to be applied to unfiltered designs already spanning the full $[0,1]$ range and resembling the final design, as in \Cref{fig:outsmarting3,fig:outsmarting5,fig:outsmarting7}. This is not the case for the NN ansatz in \Cref{fig:outsmarting4,fig:outsmarting6,fig:outsmarting8}. The cause is linked to the initialization, which consists of random noise centered around 0.5 (see \Cref{fig:outsmarting2}). However, shifting the centering closer to 0 or 1 does not help. The linear ansatz does not suffer from this phenomenon independent of its initialization. \\

The positive side is that the NN reaches a slightly better optimum, i.e., $L_p^{\text{before}}=59.88$~dB, compared to $L_p^{\text{before}}=60.22$~dB with the linear ansatz. However, this phenomenon of bypassing the filtering and projection diminishes any advantage gained after thresholding the final design to the discrete $0/1$ space. The design using the NN ansatz goes from $L_p^{\text{before}}=59.88$~dB to $L_p^{\text{after}}=70.09$~dB after thresholding, while with the linear ansatz, the thresholded design results in $L_p^{\text{after}}=60.33$~dB. To overcome this problem, the NN initialization has to be improved, which will be achieved through transfer learning.

\subsubsection{Transfer learning}\label{sssec:pretraining}
\begin{figure}[htb]
	\centering
     \begin{subfigure}[t]{0.49\textwidth}
		\includegraphics[width=\textwidth]{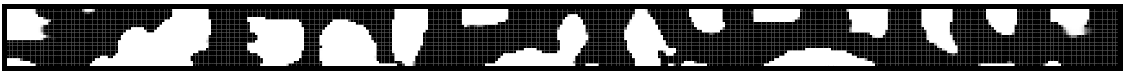}
		\caption{Linear ansatz after 200 epochs, $f=166.56$~Hz, $L_p=78.6$~dB}\label{fig:restarting1}
	\end{subfigure}
	\\
    \begin{subfigure}[t]{0.49\textwidth}
		\includegraphics[width=\textwidth]{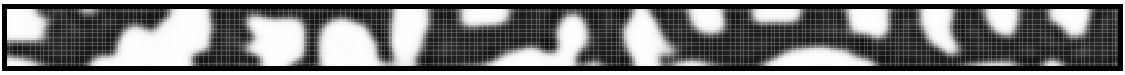}
		\caption{NN ansatz initialization, $f=166.56$~Hz}\label{fig:restarting2}
	\end{subfigure}
    \\
    \begin{subfigure}[t]{0.49\textwidth}
		\includegraphics[width=\textwidth]{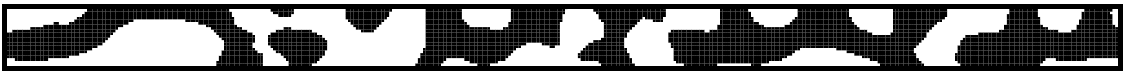}
		\caption{NN ansatz after optimization, $f=166.56$~Hz, $L_p=71.5$~dB}\label{fig:restarting3}
	\end{subfigure}
    \caption{Restarting scheme in which an optimization with linear ansatz is first performed (a), then learned by the NN in a supervised manner (b), and used as an initial guess in a subsequent optimization with the NN ansatz (c)}\label{fig:restarting}
\end{figure}

To overcome the aforementioned filtering and projection bypassing problem (\Cref{ssec:outsmarting}) and still exploit the potential identification of better local optima with an NN ansatz (\Cref{ssec:motivationNN}), restarting and transfer learning are introduced. The main goal is to change the initial guess from \Cref{fig:outsmarting2} to a material distribution resembling \Cref{fig:outsmarting3} or \Cref{fig:outsmarting4}. To this end, two schemes are proposed:
\begin{itemize}
    \item a restarting scheme,
    \item transfer learning.
\end{itemize}
In the restarting scheme, an optimization with the linear ansatz is performed, followed by an optimization with an NN ansatz, in which the final design from the first optimization (see \Cref{fig:restarting1} is used as an initial guess (see~\Cref{fig:restarting2}). Importantly, the projection parameter $\beta$ from \Cref{eq:projection} is reset to one for the second optimization --- similarly to how the learning rate is reset, i.e., increased in cosine annealing~\cite{loshchilov_sgdr_2017}. The idea of the scheme is summarized by \Cref{fig:restarting}. \\

\begin{figure}[htb]
    \centering
    \begin{subfigure}[t]{0.49\textwidth}
		\includegraphics[width=\textwidth]{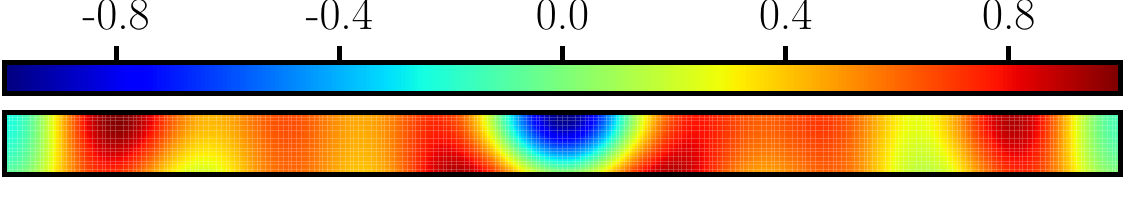}
		\caption{Input for a training example: the scaled sensitivity $dC/d\zeta_v$ assuming $\zeta_v=0$}\label{fig:NNpretrainingpredictions1}
	\end{subfigure}
    \hfill
    \begin{subfigure}[t]{0.49\textwidth}
		\includegraphics[width=\textwidth]{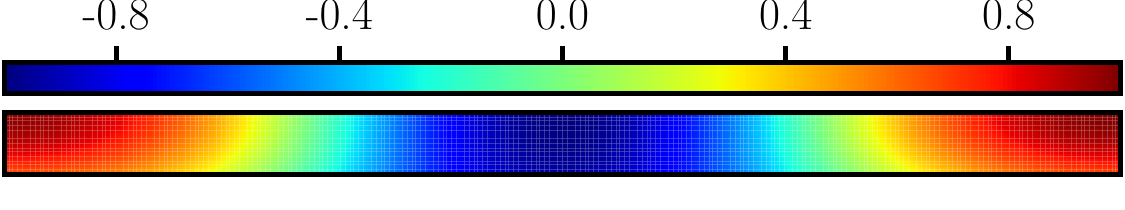}
		\caption{Input for a validation example: the scaled sensitivity $dC/d\zeta_v$ assuming $\zeta_v=0$}\label{fig:NNpretrainingpredictions2}
	\end{subfigure}
    \\
    \begin{subfigure}[t]{0.49\textwidth}
		\includegraphics[width=\textwidth]{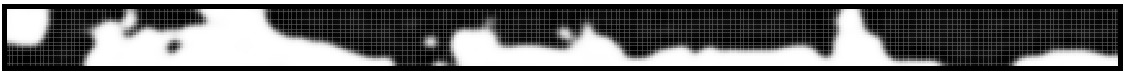}
		\caption{Prediction from training set}\label{fig:NNpretrainingpredictions3}
	\end{subfigure}
    \hfill
    \begin{subfigure}[t]{0.49\textwidth}
		\includegraphics[width=\textwidth]{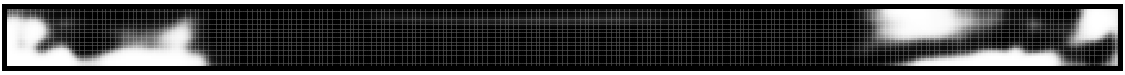}
		\caption{Prediction from validation set}\label{fig:NNpretrainingpredictions4}
	\end{subfigure}
    \\
    \begin{subfigure}[t]{0.49\textwidth}
		\includegraphics[width=\textwidth]{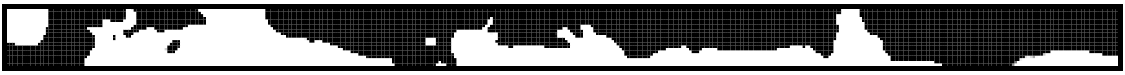}
		\caption{Label from training set}\label{fig:NNpretrainingpredictions5}
	\end{subfigure}
    \hfill
 	\begin{subfigure}[t]{0.49\textwidth}
		\includegraphics[width=\textwidth]{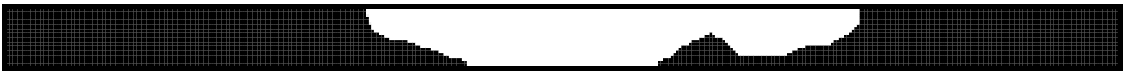}
		\caption{Label from validation set}\label{fig:NNpretrainingpredictions6}
	\end{subfigure}
    \caption{Labeled data for performing the NN pretraining and its corresponding predictions with an NN trained on 4 samples}\label{fig:NNpretrainingpredictions}
\end{figure}

For transfer learning, a dataset of optimized designs obtained with the linear ansatz is generated. Specifically, the dataset is composed of the initial cost function gradients with respect to the indicator $\partial C/\partial \boldsymbol{\zeta}^{\text{initial}}$, assuming $\boldsymbol{\zeta}^{\text{initial}}=0$ (see \Cref{fig:NNpretrainingpredictions1,fig:NNpretrainingpredictions2}), and the final designs $\boldsymbol{\zeta}^{\text{final}}$ (see \Cref{fig:NNpretrainingpredictions5,fig:NNpretrainingpredictions6}). The dataset is generated by sticking to the setup from \Cref{fig:setup} and randomly sampling the frequency $f$ in the range $[10,100]$~Hz. Next, the NN is trained in a supervised manner, such that a mapping between $\partial C/\partial \boldsymbol{\zeta}^{\text{initial}}$ and $\boldsymbol{\zeta}^{\text{final}}$ is established. This step is referred to as pretraining, where the goal is not to obtain an ideal mapping to be used in an online scenario, as in~\cite{abueidda_topology_2020,zheng_generating_2021,wang_deep_2022,yan_deep_2022}, but simply a closer qualitative resemblance to material distributions such as those seen in \Cref{fig:outsmarting3,fig:outsmarting5,fig:outsmarting7}. This is illustrated by the prediction of a validation data point in \Cref{fig:NNpretrainingpredictions3}, which demonstrates poor generalization performance compared to the corresponding label in \Cref{fig:NNpretrainingpredictions6}, but a qualitative match with $0/1$ designs. This behavior is, therefore, desired, as the target --- of the prediction qualitatively matching a $0/1$ design --- is fulfilled. Only a few data points are required to achieve this, as illustrated with \Cref{fig:NNpretrainingpredictions} relying on only 4 samples. We experienced the greatest benefit with sample sizes between 4 and 8.\\

\begin{figure}[htb]
	\centering
    \begin{subfigure}[t]{0.49\textwidth}
		\includegraphics[width=\textwidth]{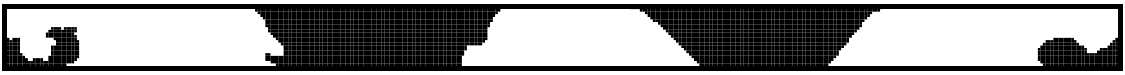}
		\caption{linear ansatz after optimization, $f=34.39$~Hz, $L_p=74.9$~dB}\label{fig:conceptPretraining1}
	\end{subfigure}
    \hfill
    \begin{subfigure}[t]{0.49\textwidth}
		\includegraphics[width=\textwidth]{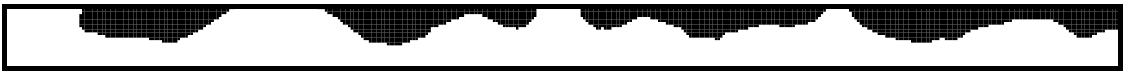}
		\caption{linear ansatz after optimization, $f=76.30$~Hz, $L_p=80.1$~dB}\label{fig:conceptPretraining2}
	\end{subfigure}
	\\
    \begin{subfigure}[t]{0.49\textwidth}
		\includegraphics[width=\textwidth]{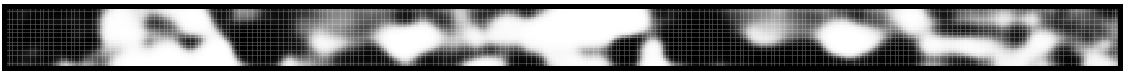}
		\caption{NN ansatz initialization, $f=34.39$~Hz}\label{fig:conceptPretraining3}
	\end{subfigure}
    \hfill
    \begin{subfigure}[t]{0.49\textwidth}
		\includegraphics[width=\textwidth]{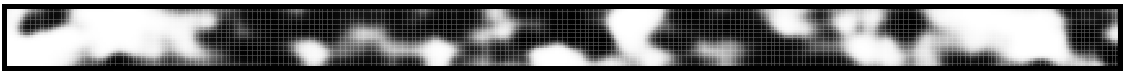}
		\caption{NN ansatz initialization, $f=76.30$~Hz}\label{fig:conceptPretraining4}
	\end{subfigure}
    \\
    \begin{subfigure}[t]{0.49\textwidth}
		\includegraphics[width=\textwidth]{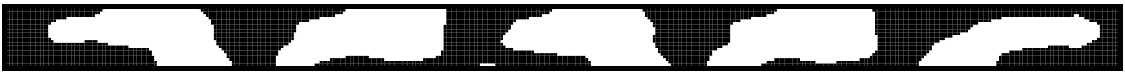}
		\caption{NN ansatz after optimization, $f=34.39$~Hz, $L_p=69.8$~dB}\label{fig:conceptPretraining5}
	\end{subfigure}
    \hfill
    \begin{subfigure}[t]{0.49\textwidth}
		\includegraphics[width=\textwidth]{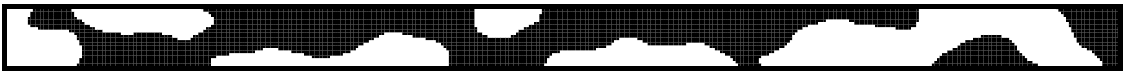}
		\caption{NN ansatz after optimization, $f=76.30$~Hz, $L_p=67.9$~dB}\label{fig:conceptPretraining6}
	\end{subfigure}
    \caption{Transfer learning scheme. An NN is pretrained on a number of optimized designs yielding initial guesses (c) and (d). A subsequent optimization leads to the optimized topologies (e) and (f). For reference, the results of a conventional linear ansatz optimization without pretaining is depicted in (a) and (b).}\label{fig:conceptPretraining}
\end{figure}

The scheme is demonstrated on two examples with correspondingly $f=34.39$~Hz and $f=76.30$~Hz in~\Cref{fig:conceptPretraining}. From the NN initialization obtained through transfer learning (see \Cref{fig:conceptPretraining3,fig:conceptPretraining4}), an optimization is performed with the NN ansatz, yielding the designs depicted in \Cref{fig:conceptPretraining5,fig:conceptPretraining6}. As a comparison, optimized designs obtained with a conventional linear ansatz are provided in \Cref{fig:conceptPretraining1,fig:conceptPretraining2}. \\

Both schemes were tested extensively, but the transfer learning outperformed the restarting scheme slightly, which is briefly outlined in Appendix~\ref{appendix:restarting}. Therefore, only the pretraining results will be discussed in the following section (\Cref{sec:results}). Both schemes require the additional overhead of pretraining to the linear ansatz and can only be deemed beneficial if better designs are discovered, which would not have been found conventionally. 

\section{Results}\label{sec:results}

The optimized sound pressure levels will be evaluated on the three stages of optimization described in Appendix~\ref{appendix:optimization}. The first is the sound pressure level after the first 280 epochs obtained with the discretization $q=2, n_v=4$, while the second is the sound pressure level obtained from the second optimization with $q=4, n_v=4$. The last value stems from the evaluation based on $q=2, n_v=1$. This will be presented according to \Cref{tab:NNlabels}. \\

\begin{table}[htb]
    \centering
    \caption{Labels of sound pressure levels in upcoming investigations (\Cref{tab:Reference,tab:NNansatz4}). The first is the sound pressure level after the first 280 epochs with $q=2, n_v=4$, and the second is the sound pressure level after the subsequent optimization with a higher polynomial degree of $q=4$. Lastly, the optimized design is evaluated on a discretization with $q=2, n_v=1$. All designs are thresholded before evaluation.}\label{tab:NNlabels}
    \begin{tabular}{c|c|c}
        \begin{tabular}{c}sound pressure level\\after first optimization\\$q=2, n_v=4$\end{tabular} 
        & \begin{tabular}{c}sound pressure level\\after second optimization\\$q=4, n_v=4$\end{tabular} 
        & \begin{tabular}{c}sound pressure level\\after second optimization\\$q=2, n_v=1$\end{tabular} \\
        \hline
        \hline
         -- & -- & -- \\
        \hline
    \end{tabular}
\end{table}

\begin{figure}[htb]
	\centering
     \begin{subfigure}[t]{0.49\textwidth}
		\includegraphics[width=\textwidth]{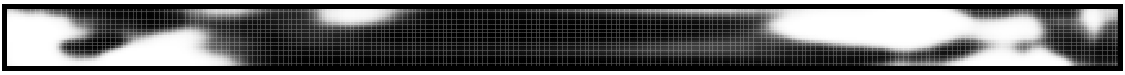}
		\caption{Seed 1}\label{fig:seed1}
	\end{subfigure}
	\\
    \begin{subfigure}[t]{0.49\textwidth}
		\includegraphics[width=\textwidth]{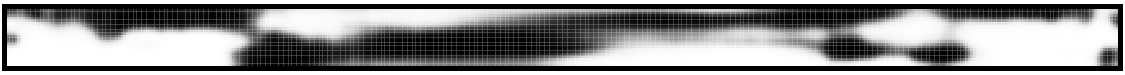}
		\caption{Seed 2}\label{fig:seed2}
	\end{subfigure}
    \\
    \begin{subfigure}[t]{0.49\textwidth}
		\includegraphics[width=\textwidth]{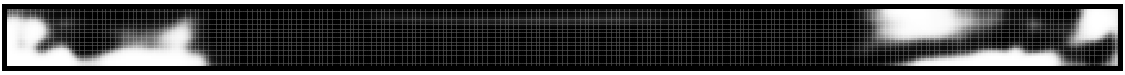}
		\caption{Seed 3}\label{fig:seed3}
	\end{subfigure}
   \caption{Three different NN initial guesses for the same optimization problem obtained by changing the NN weight initialization during pretraining (accomplished by modifying the seed)}\label{fig:seed}
\end{figure}

Additionally, the computational effort varies from case to case, captured by the required epochs for tuning initial guesses and learning rates for the first and second optimization. The first optimization always consists of 280 epochs, while the second optimization ranges between 1 and 101 epochs, depending on the number of needed iterations to correct the design due to the initial under-discretization. In addition, a tuning is performed prior to the first optimization. In the linear ansatz, three initial guesses: $\boldsymbol{\zeta}=0, \boldsymbol{\zeta}=0.5, \boldsymbol{\zeta}=1$ are compared for 50 epochs of which the best is selected\footnote{See \Cref{ssec:statisticalEvaluation} for a statistical evaluation of the initialization's impact.}. Similarly, for the NN ansatz, three different initializations are tested by performing three different pretrainings with the same training data but with variations of the NN weight initialization. Three exemplary initial guesses after pretraining are illustrated in \Cref{fig:seed} reusing the validation data point from \Cref{fig:NNpretrainingpredictions4,fig:NNpretrainingpredictions6}. However, the optimization's success is also more sensitive to the learning rate when working with the NN ansatz. To this end, also three learning rates: $\alpha=10^{-5}$, $\alpha=2\cdot 10^{-5}$, $\alpha=4\cdot 10^{-5}$ were compared. This results in $3\times 3\times 50=450$ tuning epochs for the NN ansatz, yielding an increase in computational effort of about ${\sim}1.7$ per optimization in comparison to the linear ansatz with $150$ tuning epochs. Note that the data generation required for the pretraining also comes with a computational overhead, which, relatively seen, is reduced when multiple subsequent optimizations are performed.

\subsection{Reference Optimization}
To assess if the NN ansatz provides an improvement, a baseline is established with the conventional linear ansatz. The corresponding results are summarized in \Cref{tab:Reference} for the eight chosen frequencies.
\begin{table}[H]
    \centering
    \caption{Reference optimized sound pressure levels with \textbf{linear ansatz} using the best result from three initial guesses $\boldsymbol{\zeta}=0, \boldsymbol{\zeta}=0.5, \boldsymbol{\zeta}=1$. Sound pressure levels $L_p$ are given according to the two stages of training and a final evaluation, see \Cref{tab:NNlabels}. The epochs consist of the tuning iterations, the first optimization with $q=2, n_v=4$, and the second optimization $q=4, n_v=4$. The epochs of the second optimization are marked in bold as they vary from case to case and are more expensive.}\label{tab:Reference}
    \setlength{\tabcolsep}{4pt}
    \begin{tabular}{ccc}
        $f$ & \begin{tabular}{c}sound pressure level\end{tabular} & \begin{tabular}{c}epochs\end{tabular} \\ 
        \hline
        $21.33$~Hz & \begin{tabular}{c|c|c} 78.5~dB & 79.8~dB & 82.2~dB \end{tabular} & 150 \& 280 \& \textbf{1}\\
        \hline
        $34.39$~Hz & \begin{tabular}{c|c|c} 75.0~dB & 75.2~dB & 74.9~dB \end{tabular} & 150 \& 280 \& \textbf{11}\\
        \hline
        $57.22$~Hz & \begin{tabular}{c|c|c} 64.3~dB & 62.4~dB & 69.1~dB \end{tabular} & 150 \& 280 \& \textbf{1}\\
        \hline
        $69.43$~Hz & \begin{tabular}{c|c|c} 60.4~dB & 61.3~dB & 60.5~dB \end{tabular} & 150 \& 280 \& \textbf{11}\\
        \hline
        $76.30$~Hz & \begin{tabular}{c|c|c} 78.1~dB & 78.8~dB & 80.1~dB \end{tabular} & 150 \& 280 \& \textbf{1}\\
        \hline
        $95.37$~Hz & \begin{tabular}{c|c|c} 77.3~dB & 79.5~dB & 82.9~dB \end{tabular} & 150 \& 280 \& \textbf{11}\\
        \hline
        $141.78$~Hz & \begin{tabular}{c|c|c} 75.8~dB & 78.0~dB & 80.5~dB \end{tabular} & 150 \& 280 \& \textbf{101}\\
        \hline
        $166.56$~Hz & \begin{tabular}{c|c|c} 74.8~dB & 75.0~dB & 77.8~dB \end{tabular} & 150 \& 280 \& \textbf{21}\\
        \hline
    \end{tabular}
\end{table}

\subsection{Neural Network Ansatz with Transfer Learning}\label{ssec:resultpretrainedNN}

The same problem is now optimized for the eight frequencies utilizing the transfer learning scheme described in \Cref{sssec:pretraining} relying on 4 samples during pretraining. The results are summarized by \Cref{tab:NNansatz4}, where gray highlighting indicates a superiority over the reference obtained in \Cref{tab:Reference}. The NN yields better optima in all but one case in the optimization phase where $f=69.43$~Hz poses an exception. During the evaluation, the case with $f=166.56$~Hz also turns out to be worse. In six out of eight cases, the NN ansatz yields an improvement of correspondingly $12.8$~dB, $1.0$~dB, $6.4$~dB, $11.9$~dB, $14.7$~dB, and $4.6$~dB. Thus, the NN ansatz provides a benefit for determining better optima with a ${~\sim}$1.7 higher computational effort. \\

Interestingly, the designs at higher frequencies --- beyond the $f=100$~Hz threshold employed during pretraining --- yield less or no improvement over the reference in \Cref{tab:Reference}. By modifying the pretraining to incorporate higher frequencies up to $f=200$~Hz, the designs improve. Thus, the frequency dependency of the designs is relevant for the proposed transfer learning scheme.
Further generalization properties are studied in \Cref{ssec:generalization}.

\begin{table}[H]
    \centering
    \caption{Optimized sound pressure levels with \textbf{NN ansatz} based on \textbf{transfer learning} with \textbf{4 samples} in the range $10-100$~Hz. The three sound pressure levels are provided according to \Cref{tab:NNlabels}. The gray highlights indicate that the reference from \Cref{tab:Reference} is outperformed. Underlining indicates a better design than achieved with transfer learning utilizing a linear ansatz (see \Cref{tab:Linearansatz4}). Additionally, the improvement over the reference designs from \Cref{tab:Reference} is quantified.}\label{tab:NNansatz4}
    \setlength{\tabcolsep}{4pt}
    \begin{tabular}{cccc}
        $f$ & \begin{tabular}{c}sound pressure level\end{tabular} & improvement & \begin{tabular}{c}epochs\end{tabular} \\ 
        \hline
        $21.33$~Hz & \begin{tabular}{c|c|c} \cellcolor{lightgray}\underline{63.2~dB} & \cellcolor{lightgray}\underline{66.8~dB} & \cellcolor{lightgray}\underline{69.4~dB} \end{tabular} & 12.8~dB & 450 \& 280 \& \textbf{1}\\
        \hline
        $34.39$~Hz & \begin{tabular}{c|c|c} \cellcolor{lightgray}\underline{73.1~dB} & \cellcolor{lightgray}\underline{73.5~dB} & \cellcolor{lightgray}\underline{73.9~dB} \end{tabular} & 1.0~dB & 450 \& 280 \& \textbf{1}\\
        \hline
        $57.22$~Hz & \begin{tabular}{c|c|c} \cellcolor{lightgray}\underline{54.6~dB} & \cellcolor{lightgray}\underline{60.0~dB} & \cellcolor{lightgray}\underline{62.7~dB} \end{tabular} & 6.4~dB & 450 \& 280 \& \textbf{1}\\
        \hline
        $69.43$~Hz & \begin{tabular}{c|c|c} \underline{64.7~dB} & \underline{64.1~dB} & \underline{66.6~dB} \end{tabular} & -6.1~dB & 450 \& 280 \& \textbf{31}\\
        \hline
        $76.30$~Hz & \begin{tabular}{c|c|c} \cellcolor{lightgray}\underline{74.3~dB} & \cellcolor{lightgray}\underline{75.2~dB} & \cellcolor{lightgray}\underline{68.2~dB} \end{tabular} & 11.9~dB & 450 \& 280 \& \textbf{1}\\
        \hline
        $95.37$~Hz & \begin{tabular}{c|c|c} \cellcolor{lightgray}\underline{72.4~dB} & \cellcolor{lightgray}\underline{72.2~dB} & \cellcolor{lightgray}\underline{68.2~dB} \end{tabular} & 14.7~dB & 450 \& 280 \& \textbf{11}\\
        \hline
        $141.78$~Hz & \begin{tabular}{c|c|c} \cellcolor{lightgray}69.4~dB & \cellcolor{lightgray}69.8~dB & \cellcolor{lightgray}\underline{75.9~dB} \end{tabular} & 4.6~dB & 450 \& 280 \& \textbf{41}\\
        \hline
        $166.56$~Hz & \begin{tabular}{c|c|c} \cellcolor{lightgray}\underline{73.1~dB} & \cellcolor{lightgray}\underline{74.2~dB} & \underline{78.5~dB} \end{tabular} & -0.7~dB & 450 \& 280 \& \textbf{21}\\
        \hline
    \end{tabular}
\end{table}

\subsection{Neural Network Prediction as Initial Guess for Linear Ansatz}\label{ssec:initialGuessLinear}
Naturally, the question arises of whether it is truly the NN that is beneficial or only the initial guess provided through the pretraining. To this end, we isolate the effect of the pretraining. Instead of relying on the NN ansatz during the optimization, a linear ansatz is employed but initialized with the initial guess of the pretrained NN. With the same pretraining as employed for \Cref{tab:NNansatz4}, an optimization with the linear ansatz and initial NN guess results in \Cref{tab:Linearansatz4}. As seen by the gray highlights, only half of the investigated cases provide an improvement over the reference from \Cref{tab:Reference}. Furthermore, the underlined sound pressure level values in \Cref{tab:NNansatz4} and \Cref{tab:Linearansatz4} indicate a better attained optimum with the respective method. The linear ansatz with an NN initial guess only yields a better design for one case, $f=141.78$~Hz, which, however, turns out to be worse during the final evaluation with $q=2, n_v=1$. \\

\begin{table}[htb]
    \centering
    \caption{Optimized sound pressure levels with \textbf{linear ansatz} using \textbf{transfer learning} with \textbf{4 samples} in the range $10-100$~Hz to provide an initial guess. The three sound pressure levels are provided according to \Cref{tab:NNlabels}. The gray highlights indicate that the reference from \Cref{tab:Reference} is outperformed. Underlining indicates a better design than achieved with transfer learning utilizing a linear ansatz (see \Cref{tab:NNansatz4}). Additionally, the improvement over the reference designs from \Cref{tab:Reference} is quantified.}\label{tab:Linearansatz4}
    \setlength{\tabcolsep}{4pt}
    \begin{tabular}{cccc}
        $f$ & \begin{tabular}{c}sound pressure level\end{tabular} & improvement & \begin{tabular}{c}epochs\end{tabular} \\ 
        \hline
        $21.33$~Hz & \begin{tabular}{c|c|c} 82.4~dB & 84.2~dB & 85.0~dB \end{tabular} & -2.8~dB & 450 \& 280 \& \textbf{1}\\
        \hline
        $34.39$~Hz & \begin{tabular}{c|c|c} \cellcolor{lightgray}74.6~dB & 75.8~dB & 76.6~dB \end{tabular} & -1.7~dB & 450 \& 280 \& \textbf{1}\\
        \hline
        $57.22$~Hz & \begin{tabular}{c|c|c} \cellcolor{lightgray}62.3~dB & \cellcolor{lightgray}60.2~dB & \cellcolor{lightgray}68.9~dB \end{tabular} & 0.2~dB & 450 \& 280 \& \textbf{11}\\
        \hline
        $69.43$~Hz & \begin{tabular}{c|c|c} 71.9~dB & 72.0~dB & 73.7~dB \end{tabular} & -13.2~dB & 450 \& 280 \& \textbf{11}\\
        \hline
        $76.30$~Hz & \begin{tabular}{c|c|c} \cellcolor{lightgray}75.2~dB & \cellcolor{lightgray}75.4~dB & \cellcolor{lightgray}80.1~dB \end{tabular} & 0.0~dB & 450 \& 280 \& \textbf{21}\\
        \hline
        $95.37$~Hz & \begin{tabular}{c|c|c} \cellcolor{lightgray}71.1~dB & \cellcolor{lightgray}65.1~dB & \cellcolor{lightgray}81.4~dB \end{tabular} & 1.5~dB & 450 \& 280 \& \textbf{101}\\
        \hline
        $141.78$~Hz & \begin{tabular}{c|c|c} \cellcolor{lightgray}\underline{59.9~dB} & \cellcolor{lightgray}\underline{67.3~dB} & \cellcolor{lightgray}76.8~dB \end{tabular} & 3.7~dB & 450 \& 280 \& \textbf{101}\\
        \hline
        $166.56$~Hz & \begin{tabular}{c|c|c} \cellcolor{lightgray}73.4~dB & 75.3~dB & 82.7~dB \end{tabular} & -4.9~dB & 450 \& 280 \& \textbf{101}\\
        \hline
    \end{tabular}
\end{table}


\begin{figure}[H]
	\centering
    \begin{subfigure}[t]{0.49\textwidth}
		\includegraphics[width=\textwidth]{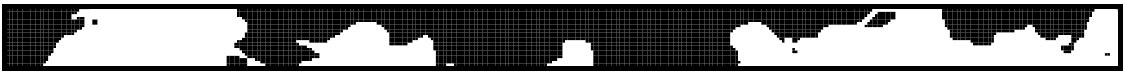}
		\caption{Linear ansatz after optimization, $f=57.22$~Hz, $L_p=68.9$~dB}\label{fig:NNvsLinear1}
	\end{subfigure}
    \hfill
    \begin{subfigure}[t]{0.49\textwidth}
		\includegraphics[width=\textwidth]{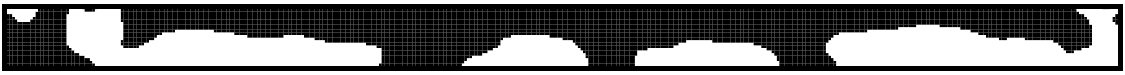}
		\caption{NN ansatz after optimization, $f=57.22$~Hz, $L_p=62.7$~dB}\label{fig:NNvsLinear2}
	\end{subfigure}
    \\
    \begin{subfigure}[t]{0.49\textwidth}
		\includegraphics[width=\textwidth]{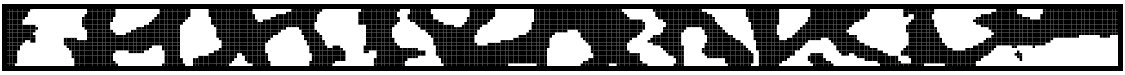}
		\caption{Linear ansatz after optimization, $f=166.56$~Hz, $L_p=82.7$~dB}\label{fig:NNvsLinear3}
	\end{subfigure}
    \hfill
    \begin{subfigure}[t]{0.49\textwidth}
		\includegraphics[width=\textwidth]{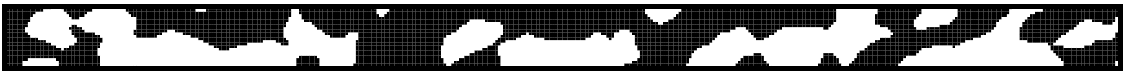}
		\caption{NN ansatz after optimization, $f=166.56$~Hz, $L_p=78.5$~dB}\label{fig:NNvsLinear4}
	\end{subfigure}
    \caption{Optimized topologies using transfer learning based on 8 samples with the linear ansatz on the left and the NN ansatz on the right}\label{fig:NNvsLinear}
\end{figure}

\begin{figure}[H]
    \centering
    \begin{subfigure}[t]{0.49\textwidth}
		\includegraphics[width=\textwidth]{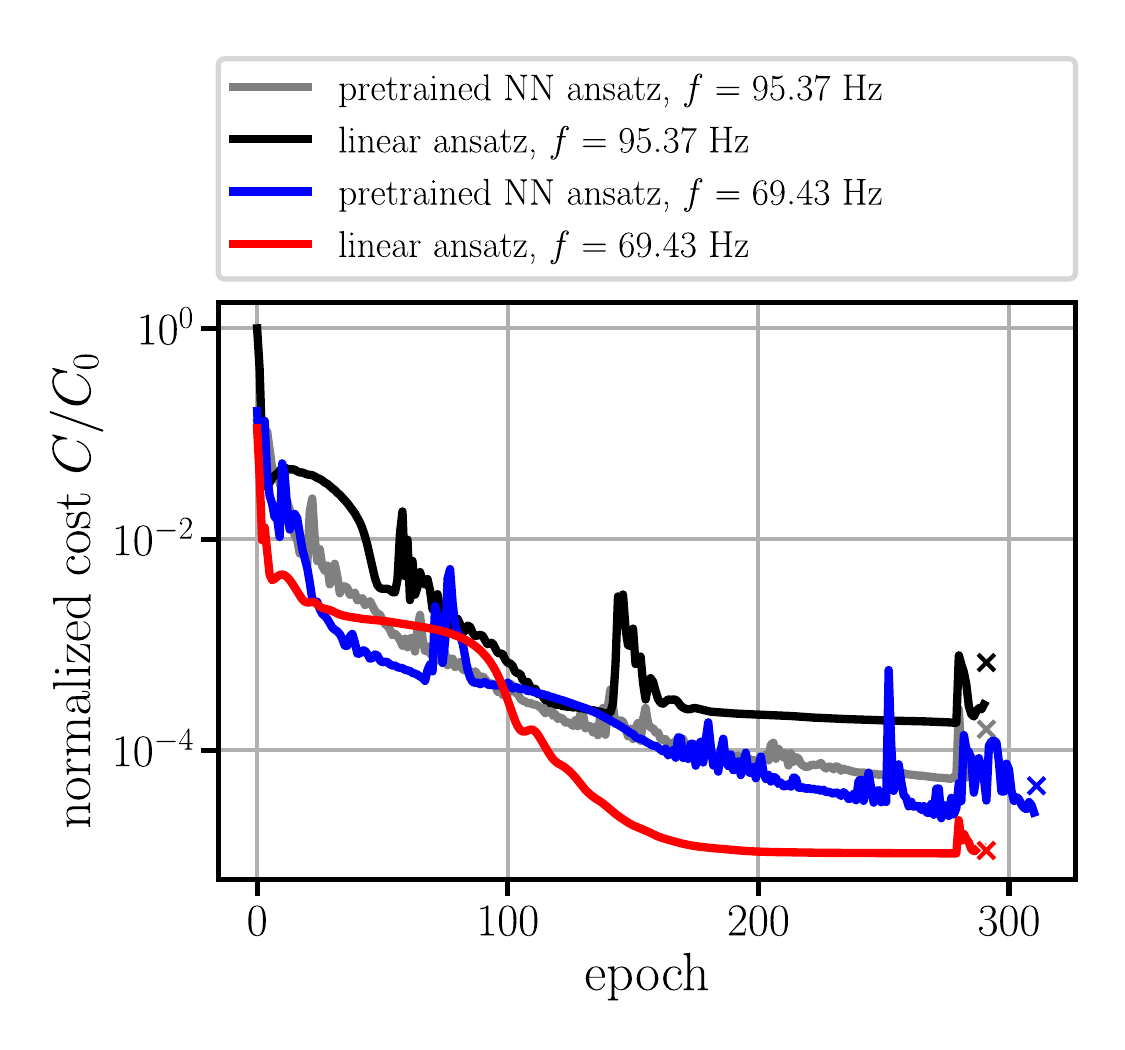}
		\caption{Pretrained NN ansatz}\label{fig:NNcostHistories41}
	\end{subfigure}
    \hfill
    \begin{subfigure}[t]{0.49\textwidth}
		\includegraphics[width=\textwidth]{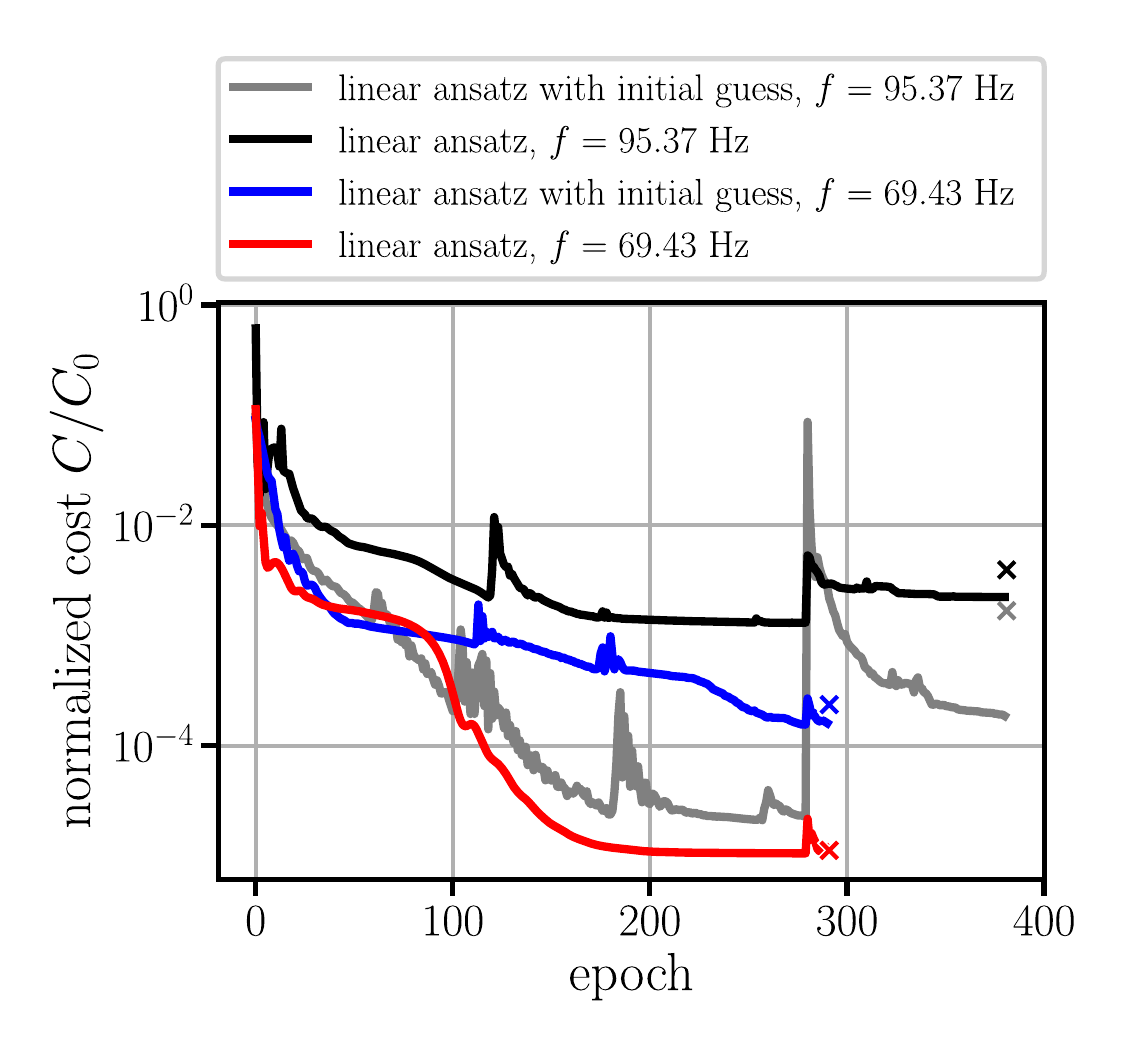}
		\caption{Linear ansatz with initial guess}\label{fig:NNcostHistories42}
	\end{subfigure}
 \caption{Cost histories of the best and worst cases of the transfer learning scheme based on 4 samples for correspondingly the NN and linear ansatz. The final correctly evaluated cost functions are indicated by the crosses. The curves with the linear ansatz (without initial guess) are provided as reference and are identical in (a) and (b).}\label{fig:NNcostHistories4}
\end{figure}

The cost histories over the epochs are illustrated in \Cref{fig:NNcostHistories4} for correspondingly the best performance achieved with an NN, i.e., $f=95.37$~Hz, and the worst performance, i.e., $f=69.43$~Hz (compare \Cref{tab:NNansatz4} with \Cref{tab:Reference}). \Cref{fig:NNcostHistories41} shows the histories of the pretrained NN ansatz compared to the reference trajectories with a linear ansatz, while \Cref{fig:NNcostHistories41} depicts the trajectories of the linear ansatz with and without initial guess provided by the NN.  

\subsection{Generalization}\label{ssec:generalization}

As a final investigation, the generalization capabilities of transfer learning are investigated --- in order to evaluate if a new pretraining is necessary for a change in problem setup (see \Cref{fig:setup}) beyond the change in frequency. To this end, the same pretraining is employed for a novel setup characterized by the changes in \Cref{tab:problemsetup2}. The source is moved closer to the ceiling and relocated to the right side of the domain, while the domain to be suppressed is reduced in size and shifted to the left. \\

\begin{table}[htb]
    \centering
    \caption{Modified dimensions of the problem. Compare to \Cref{tab:problemsetup} and the problem setup in \Cref{fig:setup}. Dimensions are provided in meters.}\label{tab:problemsetup2}
    \begin{tabular}{cccccc}
    $x_f$ & $y_f$ & $x_s$ & $y_s$ & $a_s$ & $b_s$ \\
    \hline
    17 & 6 & 5.5 & 6.5 & 1 & 1 \\
    \hline
    \end{tabular}
\end{table}

\begin{figure}[H]
	\centering
    \begin{subfigure}[t]{0.49\textwidth}
		\includegraphics[width=\textwidth]{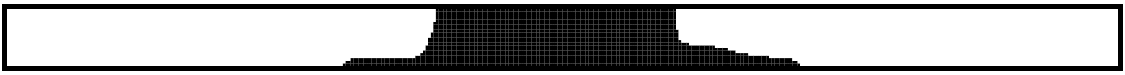}
		\caption{Linear ansatz at $f=21.33$~Hz yielding $L_p=95.8$~dB}
	\end{subfigure}
    \hfill
    \begin{subfigure}[t]{0.49\textwidth}
		\includegraphics[width=\textwidth]{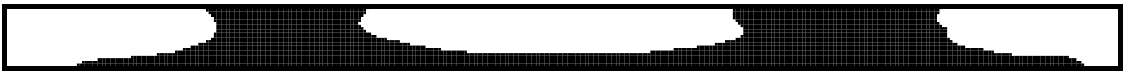}
		\caption{Pretrained NN ansatz at $f=21.33$~Hz yielding $L_p=93.6$~dB with 4 samples and $L_p=82.2$~dB with \textbf{8 samples}}
	\end{subfigure}
    \\
    \begin{subfigure}[t]{0.49\textwidth}
		\includegraphics[width=\textwidth]{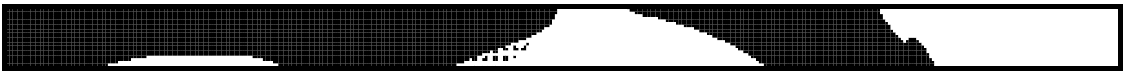}
		\caption{Linear ansatz at $f=34.39$~Hz yielding $L_p=79.1$~dB}
	\end{subfigure}
    \hfill
    \begin{subfigure}[t]{0.49\textwidth}
		\includegraphics[width=\textwidth]{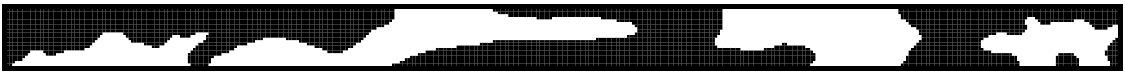}
		\caption{Pretrained NN ansatz at $f=34.39$~Hz yielding $L_p=71.1$~dB with 4 samples and $L_p=55.1$~dB with \textbf{8 samples}}
	\end{subfigure}
    \\
    \begin{subfigure}[t]{0.49\textwidth}
		\includegraphics[width=\textwidth]{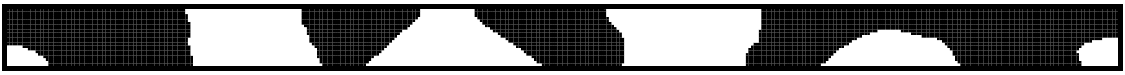}
		\caption{Linear ansatz at $f=57.22$~Hz yielding $L_p=68.0$~dB}
	\end{subfigure}
    \hfill
    \begin{subfigure}[t]{0.49\textwidth}
		\includegraphics[width=\textwidth]{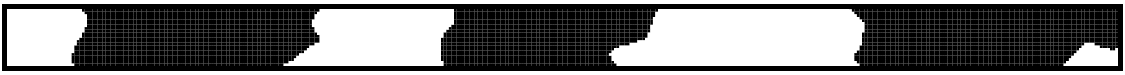}
		\caption{Pretrained NN ansatz at $f=57.22$~Hz yielding $L_p=57.9$~dB with \textbf{4 samples} and $L_p=58.0$~dB with 8 samples}
	\end{subfigure}
    \\
    \begin{subfigure}[t]{0.49\textwidth}
		\includegraphics[width=\textwidth]{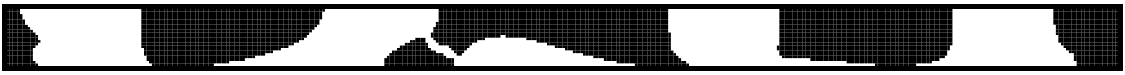}
		\caption{Linear ansatz at $f=69.43$~Hz yielding $L_p=47.8$~dB}
	\end{subfigure}
    \hfill
    \begin{subfigure}[t]{0.49\textwidth}
		\includegraphics[width=\textwidth]{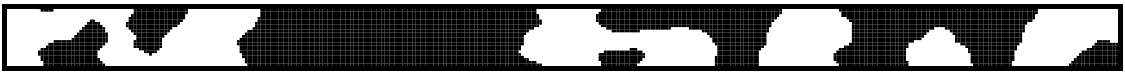}
		\caption{Pretrained NN ansatz at $f=69.43$~Hz yielding $L_p=45.7$~dB with \textbf{4 samples} and $L_p=54.7$~dB with 8 samples}
	\end{subfigure}
    \\
    \begin{subfigure}[t]{0.49\textwidth}
		\includegraphics[width=\textwidth]{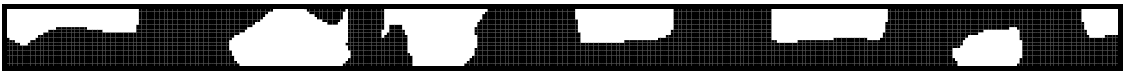}
		\caption{Linear ansatz at $f=76.30$~Hz yielding $L_p=69.5$~dB}
	\end{subfigure}
    \hfill
    \begin{subfigure}[t]{0.49\textwidth}
		\includegraphics[width=\textwidth]{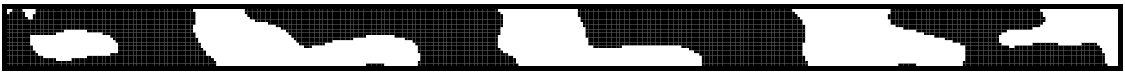}
		\caption{Pretrained NN ansatz at $f=76.30$~Hz yielding $L_p=45.2$~dB with 4 samples and $L_p=39.9$~dB with \textbf{8 samples}}
	\end{subfigure}
    \\
    \begin{subfigure}[t]{0.49\textwidth}
		\includegraphics[width=\textwidth]{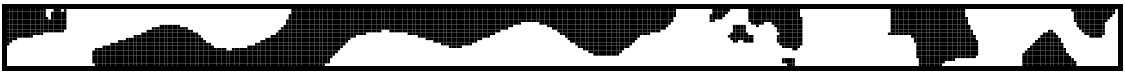}
		\caption{Linear ansatz at $f=95.37$~Hz yielding $L_p=65.0$~dB}
	\end{subfigure}
    \hfill
    \begin{subfigure}[t]{0.49\textwidth}
		\includegraphics[width=\textwidth]{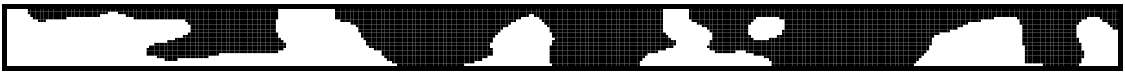}
		\caption{Pretrained NN ansatz at $f=95.37$~Hz yielding $L_p=42.2$~dB with \textbf{4 samples} and $L_p=46.3$~dB with 8 samples}
	\end{subfigure}
    \\
    \begin{subfigure}[t]{0.49\textwidth}
		\includegraphics[width=\textwidth]{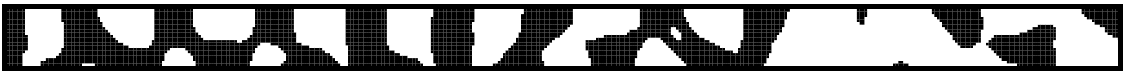}
		\caption{Linear ansatz at $f=141.78$~Hz yielding $L_p=64.1$~dB}
	\end{subfigure}
    \hfill
    \begin{subfigure}[t]{0.49\textwidth}
		\includegraphics[width=\textwidth]{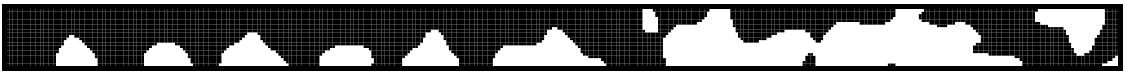}
		\caption{Pretrained NN ansatz at $f=141.78$~Hz yielding $L_p=62.7$~dB with \textbf{4 samples} and $L_p=68.6$~dB with 8 samples}
	\end{subfigure}
    \\
    \begin{subfigure}[t]{0.49\textwidth}
		\includegraphics[width=\textwidth]{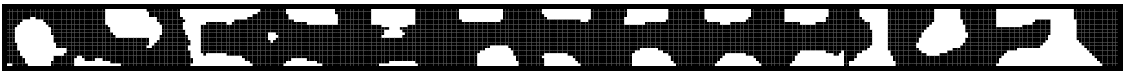}
		\caption{Linear ansatz at $f=166.56$~Hz yielding $L_p=68.7$~dB}
	\end{subfigure}
    \hfill
    \begin{subfigure}[t]{0.49\textwidth}
		\includegraphics[width=\textwidth]{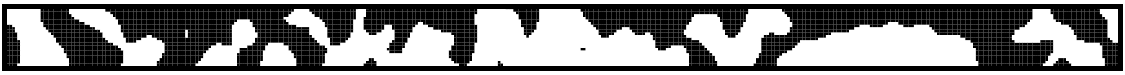}
		\caption{Pretrained NN ansatz at $f=166.56$~Hz yielding $L_p=66.8$~dB with \textbf{4 samples} and $L_p=73.3$~dB with 8 samples}
	\end{subfigure}   
    \caption{Optimized topologies for the adapted problem using transfer learning based on 4 or 8 samples with the NN ansatz on the right. The best possible design is depicted and indicated in bold. The left depicts optimized topologies without pretraining using a linear ansatz, i.e., the conventional approach.}\label{fig:generalizationCase}
\end{figure}

The optimization is performed with a pretraining using 4 and 8 samples. A comparison with designs obtained with the conventional linear ansatz is provided in \Cref{fig:generalizationCase}. For the NN ansatz, the best of the two pretrainings is illustrated. The pretraining with 4 samples consistently outperforms the linear ansatz for every tested frequency, while with 8 samples, the linear ansatz remains superior for $f=69.43$~Hz, $f=141.78$~Hz, and $f=166.56$~Hz. However, the extent to which the linear ansatz designs are surpassed is impressive. When choosing the best of the two NN pretrainings, the improvements are 13.6~dB, 24~dB, 10.1~dB, 2.1~dB, 29.6~dB, 22.8~dB, 1.4~dB, and 1.9~dB. Thus, a single pretraining can be reused for different problem setups, rendering the relative cost of the pretraining negligible. 

\subsection{Statistical Evaluation}\label{ssec:statisticalEvaluation}
\begin{figure}[htb]
	\centering
    \begin{subfigure}[t]{0.49\textwidth}
		\includegraphics[width=\textwidth]{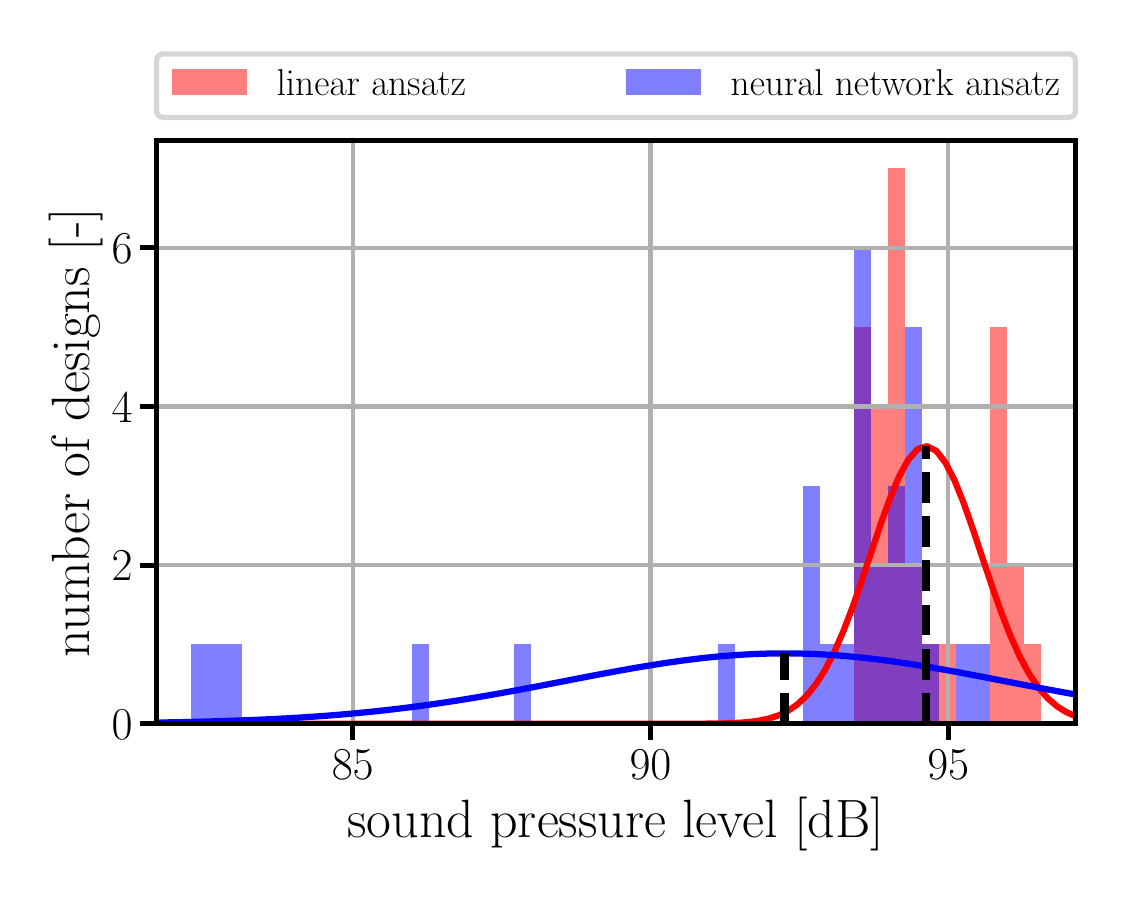}
		\caption{$f=21.33$~Hz}\label{fig:histograms1}
	\end{subfigure}
    \hfill
    \begin{subfigure}[t]{0.49\textwidth}
		\includegraphics[width=\textwidth]{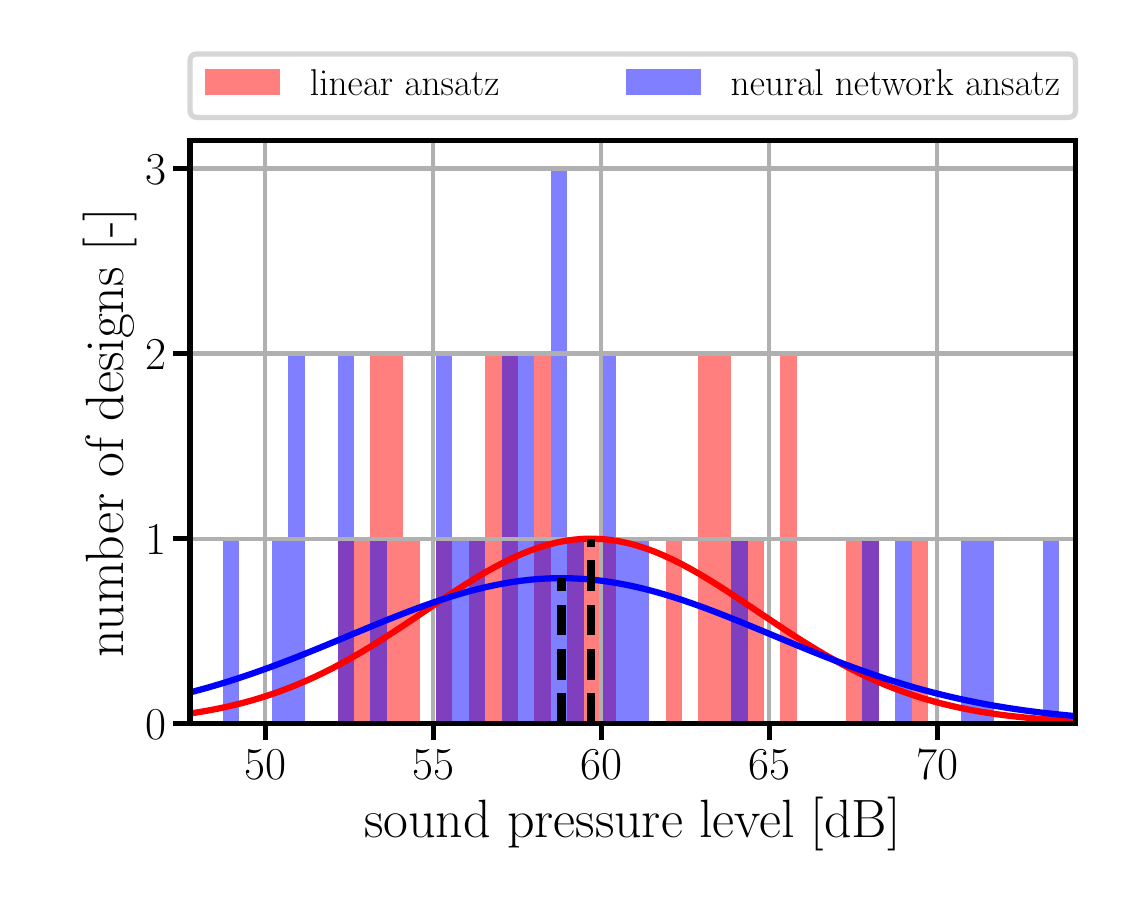}
		\caption{$f=57.22$~Hz}\label{fig:histograms2}
	\end{subfigure}
    \\
    \begin{subfigure}[t]{0.49\textwidth}
		\includegraphics[width=\textwidth]{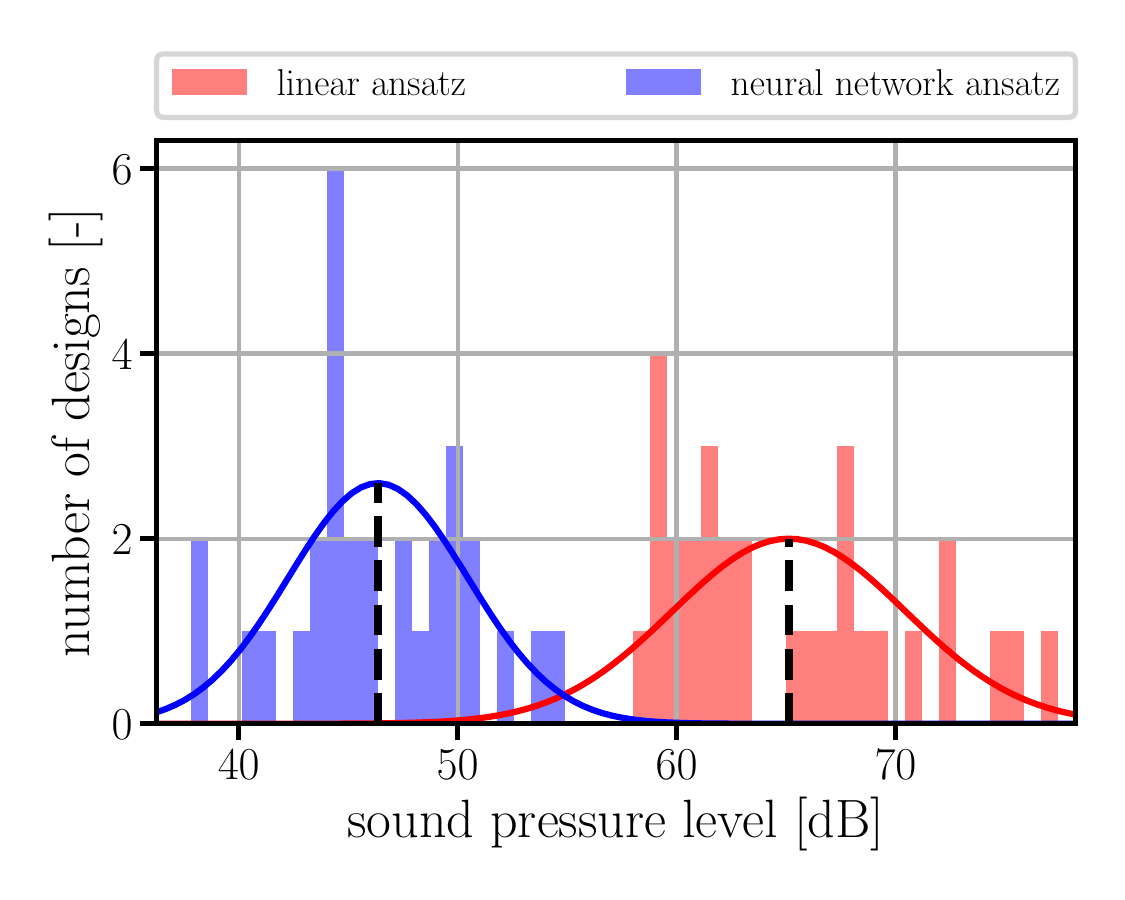}
		\caption{$f=76.30$~Hz}\label{fig:histograms3}
	\end{subfigure}
    \hfill
    \begin{subfigure}[t]{0.49\textwidth}
		\includegraphics[width=\textwidth]{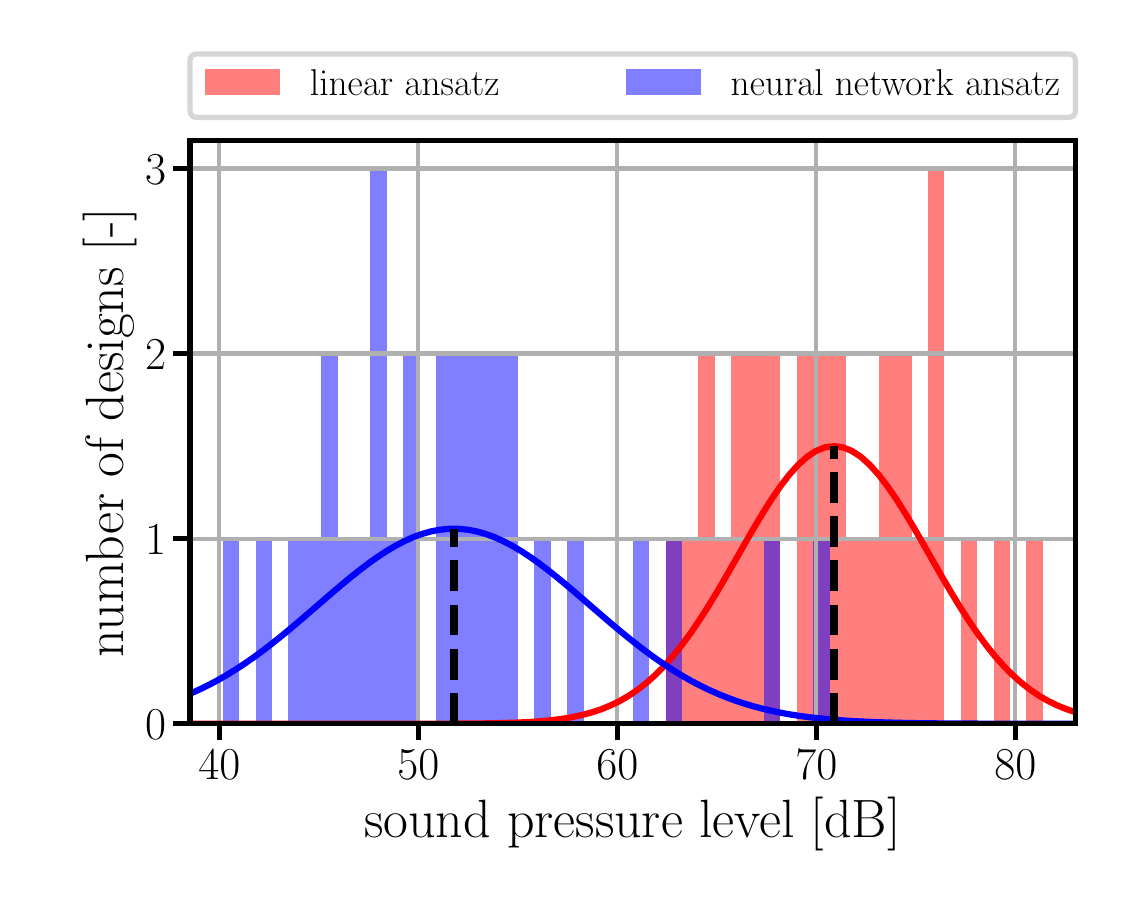}
		\caption{$f=95.37$~Hz}\label{fig:histograms4}
	\end{subfigure} 
    \caption{Statistical evaluation of the local optima quality with 30 samples for the linear (with homogeneous initial guesses) and the pretrained NN ansatz --- relying on different initializations. In the linear case, we used 30 homogeneous initial guesses equally spaced between 0 and 1, while the NN relies on 30 different pretrainings.}\label{fig:histograms}
\end{figure}

To complete the previous analysis, a statistical assessment of the initialization's effect on the two ansatz' is provided. The question is: does the NN consistently find better optima or are the observed advantages singular results discovered by chance? To this end, the generalization case from \Cref{ssec:generalization} is reconsidered. Instead of searching for the best initialization from three possibilities, separate optimizations are performed with 30 different initializations for each frequency. In the linear case, 30 homogeneous initial guesses are considered in the range $[0,1]$, while for the NN ansatz, 30 different pretrainings with the same four data samples are performed. The resulting distributions are represented as histograms in \Cref{fig:histograms} for the frequencies $f=21.33$~Hz, $f=57.22$~Hz, $f=76.30$~Hz, and $f=95.37$~Hz. The remaining histograms are provided in Appendix~\ref{appendix:statistics}. The sound pressure levels achieved with the NN ansatz are indicated in blue, while red marks those obtained using the linear ansatz. For $f=21.33$~Hz and $f=57.22$~Hz, given in \Cref{fig:histograms1,fig:histograms2}, the distributions mostly overlap. However, in both cases, the best designs are obtained with the NN --- and in \Cref{fig:histograms1}, outliers in the NN distribution represent an improvement of ${\sim}10$ dB over the linear ansatz. \Cref{fig:histograms3,fig:histograms4} with $f=76.30$~Hz, and $f=95.37$~Hz demonstrate a clear separation in design quality distribution. This also corresponds to the cases in \Cref{fig:generalizationCase} with the greatest improvement through an NN ansatz. The overlaps of the distributions are minimal, and in these scenarios, the NN ansatz consistently outperforms the linear ansatz. Thus, it can be concluded that the obtainable improvement through an NN ansatz is highly problem-dependent. Furthermore, initializations superior to the homogeneous initialization might also mitigate the NN's advantage. However, using the same NN initializations for the linear ansatz was not beneficial, as seen in \Cref{ssec:initialGuessLinear}, strengthening the potential of an NN ansatz.

\section{Conclusion}
This work investigates the usefulness of material parameterizations with NNs in the context of topology optimization on the specific problem of acoustic topology optimization. It is found that a pretrained NN relying on the concept of transfer learning is often able to identify better local optima, i.e., better designs than classical parameterizations. The origin of the benefit is currently not clear and has to be investigated further, as recently initiated by~\cite{sanu_neural_2024}. A possible explanation is that the overparametrization of the design variables decreases the chances of all gradient components becoming zero, i.e., getting stuck in a local optimum prematurely. Furthermore, similar results could possibly be reached by classical methods, such as the use of larger filters in the initial design stages to convexify the design problem~\cite{abdelhamid_revisiting_2021} and thereby avoid premature convergence. Considering the adjustment of parameters of the first NN layers as comparable to larger filters could be another explanation as to why the neural network ansatz is beneficial. \\

However, without convexification or other classical tuning efforts, finding a better optimum is often possible in the investigated acoustic topology optimization cases --- but still not guaranteed. The chances of finding one are increased by performing multiple optimizations with different NN initializations.\\

The NN ansatz is only found to be advantageous in combination with the first-order optimizer Adam, which was demonstrated with the Rosenbrock function, as a benchmark. It was also found that the presented method is currently incompatible with constrained optimizers, such as optimality criteria methods or the method of moving asymptotes. The obstacle is that concrete design variable updates can not directly be transferred to the NN parametrization but only update directions in terms of gradients. Until now, the constraints in the compliance problem have been addressed through penalty constraints~\cite{chandrasekhar_tounn_2021,chandrasekhar_multi-material_2021,chandrasekhar_approximate_2022,mallon_neural_2024}. To make the NN ansatz truly viable in topology optimization, the incorporation of constraints would have to be improved. \\

However, NN material parametrizations show some potential for acoustic topology optimization and could benefit other unconstrained optimization problems\footnote{where~\cite{sanu_neural_2024} emphasize that the potential lies in non-convex optimization problems. Applying neural network reparametrizations to convex optimization slows convergence.}. Significantly better optima were identified with comparable computational efforts --- which were less than twice as large. 

\section*{Acknowledgements}
We gratefully acknowledge the funds received by the Georg Nemetschek Institut (GNI) under the project Deep-Monitor. Furthermore, Ole Sigmund was supported by the Villum Investigator Project AMSTRAD (VIL54487) from VILLUM FONDEN.

\section*{Declarations}
\textbf{Conflict of interest} No potential conflict of interest was reported by
the authors.

\section*{Data Availability}
We provide a PyTorch~\cite{paszke_pytorch_2019} implementation of all methods in~\cite{herrmann_neural_2024}.

\renewcommand\thesection{\Alph{section}}
\setcounter{section}{0}
\section{High-Order Multi-Resolution Acoustic Topology Optimization}\label{appendix:mtop}
\begin{figure}[htb]
    \centering
    \begin{subfigure}[t]{0.49\textwidth}
        \centering
            \begin{tikzpicture}
                \fill [thin, fill=lightgray] (0,0) rectangle (6,4);
            
                \foreach \x in {0.3,0.9,...,5.7} {
                    \fill [black] (\x,2) circle (0.2cm);
                }
                \draw [thin, |-|, gray] (0.7,2.4) -- (1.1,2.4);
                \node [gray] at (0.9,2.75) {$0.3$ m};
            
                \fill [thin, fill=gray] (4.5,0.5) rectangle (5.5,1.5); 
                
                \fill [red] (1,1) circle (0.1cm);
            
                \node at (5,1) {$\Omega_s$};
                \node at (3,3) {$\Omega$};
            
                \draw[thin, |-|, gray] (0,-0.35) -- (1,-0.35);
                \draw[thin, |-|, gray] (-0.35,0) -- (-0.35,1);
                \node [gray] at (0.5,-0.57) {$2$ m};
                \node [gray] at (-0.7,0.5) {$2$ m};
            
                \draw[thin, |-|, gray] (4.5,0.15) -- (5.5,0.15);
                \draw[thin, |-|, gray] (4.15,0.5) -- (4.15,1.5);
                \node [gray] at (5,-0.2) {$2$ m};
                \node [gray] at (3.8,1) {$2$ m};
            
                \draw [thin, |-|, gray] (0,-0.75) -- (5,-0.75);
                \draw [thin, |-|, gray] (6.35,0) -- (6.35,1);
                \node [gray] at (2.5,-1) {$16$ m};
                \node [gray] at (6.7,0.5) {$2$ m};
                
                \node [red] at (1.3,1.3) {$f$};
            
                \draw[thick, latex-latex] (0,0.8) -- (0,0) -- (0.8,0);
                \node at (-0.18,0.5) {\footnotesize{$x$}};
                \node at (0.5,-0.18) {\footnotesize{$y$}};
                    
            \end{tikzpicture}
        \caption{Problem setup within the $18$ m$\times 9$ m domain with $\zeta=0$ in $\Omega$ and $\zeta=1$ in the centered black circles}\label{fig:convergence1}
	\end{subfigure}
    \hfill
    \begin{subfigure}[t]{0.49\textwidth}
		\includegraphics[width=\textwidth]{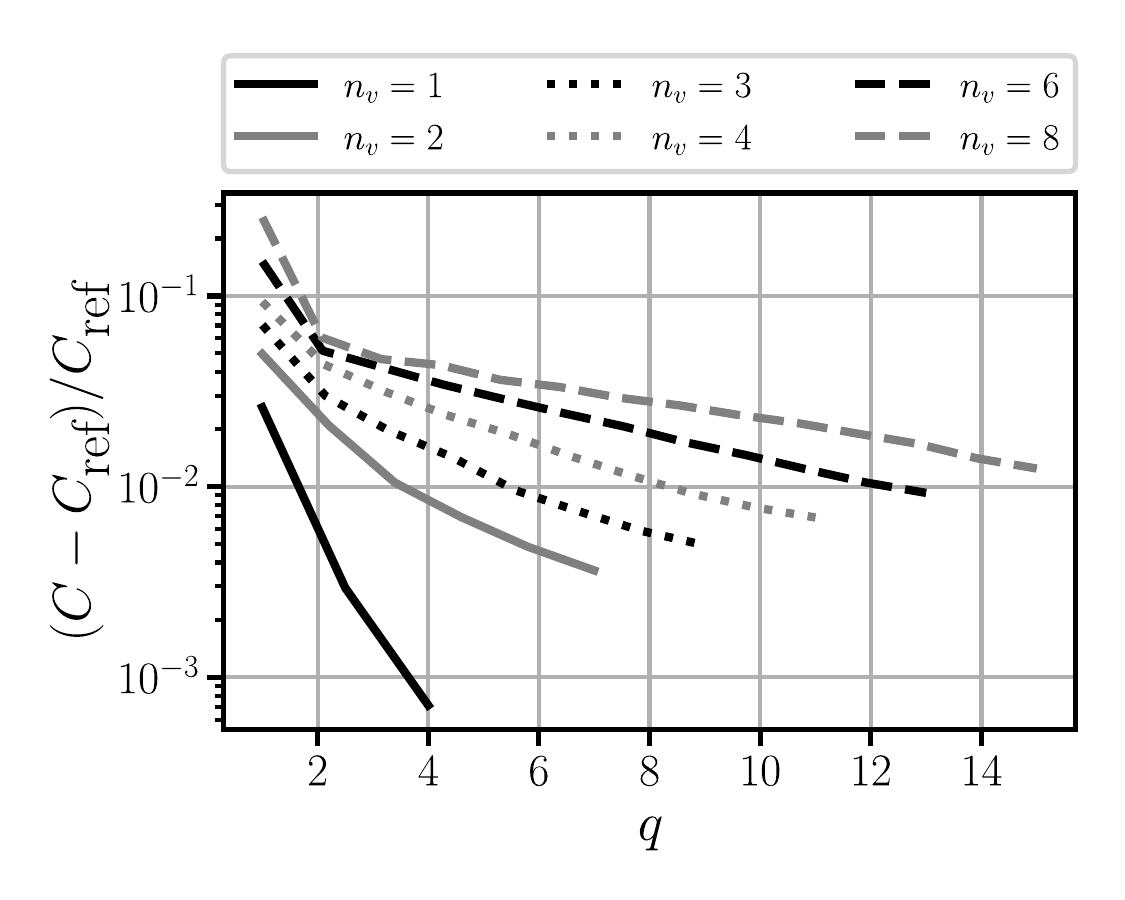}
		\caption{Convergence for $f=34.39$~Hz}\label{fig:convergence2}
	\end{subfigure}
    \caption{High-order multi-resolution convergence study}\label{fig:convergence}
\end{figure}
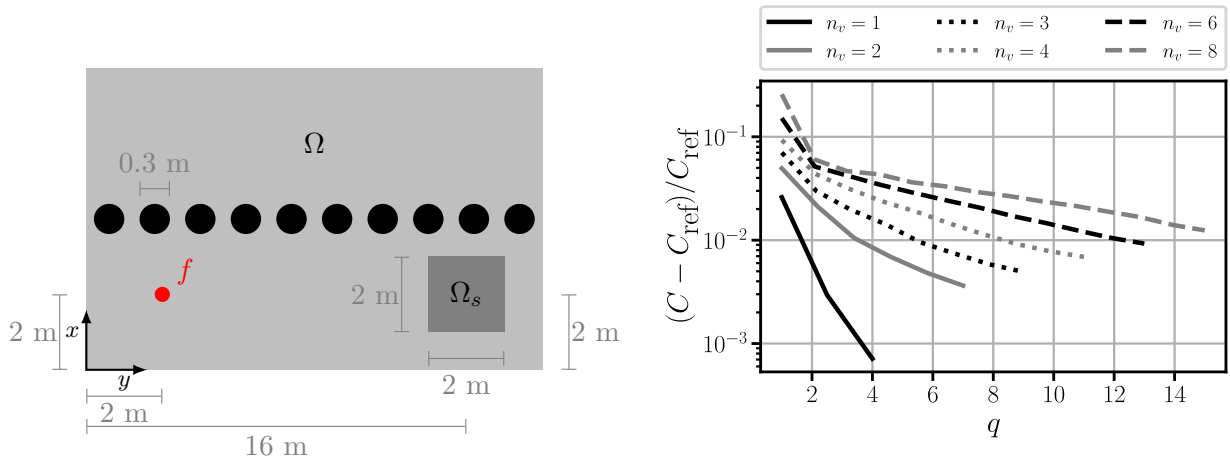

High-order multi-resolution topology optimization has previously been performed for the compliance problem in~\cite{groen_higherorder_2017}, where speedups of ${\sim}3$ in two dimensions and of ${\sim}32$ in three dimensions were obtained. Investigations for other problems have been limited. To reduce the computational effort of the acoustic topology optimization, we employ a high-order multi-resolution discretization. To this end, a brief investigation is performed. Before delving into an optimization, a convergence study for different polynomial degrees $q$ and number of subvoxels $n_v$ is performed. For this, the problem depicted in \Cref{fig:convergence1} is considered with air throughout the entire domain $\Omega$, except the circular inclusions, indicated in black. The number of total voxels is kept fixed with $432\times 216$ voxels as described in \Cref{ssec:discretization}. The considered frequency is $f=34.39$~Hz. The quantity of interest is the cost function $C$ defined as in \Cref{eq:cost}. A high-fidelity solution with $q=4, n_v=1$ is used as reference $C_{\text{ref}}$. From \Cref{fig:convergence2}, it becomes apparent that more subvoxels decrease the solution quality. However, keep in mind that more subvoxels also decrease the number of degrees of freedom drastically without decreasing the number of design variables. \\

To quantify the impact of potentially under-discretizing the problem, optimizations are to be conducted for different discretizations. To this end, we consider two frequencies $f=69.43$~Hz and $f=34.39$~Hz on the setup from \Cref{fig:setup} with dimensions from \Cref{tab:problemsetup}. For each discretization, the final design is to be tested on a more accurate discretization, which is chosen as $q=2, n_v=1$, thus incurring a relative error of less than $1 \%$ according to \Cref{fig:convergence2}. Three quantities of interest are established: 
\begin{itemize}
    \item the normalized speed-up of the simulation, where standard linear finite element, i.e., $q=1, n_v=1$ is taken as reference,
    \item the cost increase after adapting the discretization to $q=2, n_v=1$,
    \item and the final sound pressure level in dB evaluated with $q=2, n_v=1$.
\end{itemize}
Note that the design is not thresholded to compute the final sound pressure level to isolate the effect of the change in discretization. 
As these quantities are obtained for two frequencies, six values are computed for each discretization and arranged according to \Cref{tab:mtoplabels}. The optimization is conducted for 300 epochs using a filter radius twice the voxel size.

\begin{table}[htb]
    \centering
    \caption{Labels of \Cref{tab:mtop1,tab:mtop2}. Normalization with respect to time and cost is obtained with $q=1, n_v=1$.}\label{tab:mtoplabels}
    \begin{tabular}{c||c|c|c}
        & \begin{tabular}{c}normalized\\speed-up\end{tabular} 
        & \begin{tabular}{c}cost increase\\after the\\discretization change \end{tabular} 
        & \begin{tabular}{c}corrected sound\\pressure level in dB\end{tabular} \\
        \hline
        \hline
        $f=69.43$~Hz & -- & -- & -- \\
        $f=34.39$~Hz & -- & -- & -- \\
        \hline
    \end{tabular}
\end{table}

The corresponding results are presented in \Cref{tab:mtop1}, where equal performances and improvements in comparison to the discretization $q=1, n_v=1$ are highlighted in gray. Thus, any normalized speed-up $\geq 1$, any cost increase $\leq 1.5$, and any design $\leq 65$~dB and $\leq 78$~dB (within $5 \%$ of the design obtained with $q=1, n_v=1$) is highlighted. Considering this, only minor improvements can be expected with maximal speed-ups of ${\sim}3$ using $q=2, n_v=3$. \\

\begin{table}[htb]
    \centering
    \caption{Normalized time, cost increase after discretization change, and corrected sound pressure level (see \Cref{tab:mtoplabels}) for different discretizations relying on $432\times 216$ voxels at $f=69.43$~Hz and $f=34.39$~Hz}\label{tab:mtop1}
    \setlength{\tabcolsep}{4pt}
    \begin{tabular}{ccccccc}
              & $n_v=1$ & $n_v=2$ & $n_v=3$ & $n_v=4$ & $n_v=6$ & $n_v=8$ \\ 
        \hline
        $q=1$ & \begin{tabular}{c|c|c} \cellcolor{lightgray}$1$ & \cellcolor{lightgray}$1.5$ & \cellcolor{lightgray}$62$ \\ \cellcolor{lightgray}$1$ & \cellcolor{lightgray}$1$ & \cellcolor{lightgray}$75$ \end{tabular}
              & \begin{tabular}{c|c|c} \cellcolor{lightgray}$4.7$ & $2.6$ & $69$ \\ \cellcolor{lightgray}$4.8$ & \cellcolor{lightgray}$1.3$ & \cellcolor{lightgray}$76$ \end{tabular}
              & -- & -- & -- & -- \\ 
        \hline 
        $q=2$ & --
              & \begin{tabular}{c|c|c} \cellcolor{lightgray}$1.1$ & \cellcolor{lightgray}$1.4$ & \cellcolor{lightgray}$62$ \\ \cellcolor{lightgray}$1.1$ & \cellcolor{lightgray}$1.1$ & \cellcolor{lightgray}$75$ \end{tabular}
              & \begin{tabular}{c|c|c} \cellcolor{lightgray}$3.3$ & \cellcolor{lightgray}$1.5$ & \cellcolor{lightgray}$62$ \\ \cellcolor{lightgray}$3.3$ & \cellcolor{lightgray}$1.2$ & \cellcolor{lightgray}$75$ \end{tabular}
              & \begin{tabular}{c|c|c} \cellcolor{lightgray}$6.9$ & $3.6$ & $66$ \\ \cellcolor{lightgray}$7.3$ & $11.0$ & $85$ \end{tabular}
              & \begin{tabular}{c|c|c} \cellcolor{lightgray}$17.1$ & $40.9$ & $81$ \\ \cellcolor{lightgray}$17.4$ & $15.2$ & $87$ \end{tabular}
              & \begin{tabular}{c|c|c} \cellcolor{lightgray}$29.9$ & $41.1$ & $81$ \\ \cellcolor{lightgray}$30.4$ & $7.0$ & $83$ \end{tabular} \\
        \hline 
        $q=3$ & --
              & --
              & \begin{tabular}{c|c|c} \cellcolor{lightgray}$1.4$ & \cellcolor{lightgray}$1.5$ & \cellcolor{lightgray}$62$ \\ \cellcolor{lightgray}$1.4$ & \cellcolor{lightgray}$1.1$ & \cellcolor{lightgray}$75$ \end{tabular}
              & \begin{tabular}{c|c|c} \cellcolor{lightgray}$2.9$ & $1.7$ & \cellcolor{lightgray}$63$ \\ \cellcolor{lightgray}$2.9$ & \cellcolor{lightgray}$1.5$ & \cellcolor{lightgray}$77$ \end{tabular}
              & \begin{tabular}{c|c|c} \cellcolor{lightgray}$3.7$ & $6.4$ & $72$ \\ \cellcolor{lightgray}$7.2$ & $14.7$ & $87$ \end{tabular}
              & \begin{tabular}{c|c|c} \cellcolor{lightgray}$13.5$ & $6.9$ & $73$ \\ \cellcolor{lightgray}$13.6$ & $9.1$ & $84$ \end{tabular} \\
        \hline
        $q=4$ & -- & --
              & -- 
              & \begin{tabular}{c|c|c} \cellcolor{lightgray}$1.3$ & \cellcolor{lightgray}$1.5$ & \cellcolor{lightgray}$62$ \\ \cellcolor{lightgray}$1.3$ & \cellcolor{lightgray}$1$ & \cellcolor{lightgray}$75$ \end{tabular} 
              & \begin{tabular}{c|c|c} \cellcolor{lightgray}$3.2$ & $1.6$ & $66$ \\ \cellcolor{lightgray}$3.2$ & $2.4$ & $79$ \end{tabular}
              & \begin{tabular}{c|c|c} \cellcolor{lightgray}$6.1$ & $6.0$ & $74$ \\ \cellcolor{lightgray}$6.0$ & $20.1$ & $88$ \end{tabular} \\
              \hline
        $q=5$ & -- & -- & --
              & --
              & \begin{tabular}{c|c|c} \cellcolor{lightgray}$1.7$ & \cellcolor{lightgray}$1.5$ & $66$ \\  \cellcolor{lightgray}$1.7$ & $3.2$ & $80$ \end{tabular}
              & \begin{tabular}{c|c|c} \cellcolor{lightgray}$3.1$ & $1.8$ & $67$ \\ \cellcolor{lightgray}$3.0$ & $2.8$ & $79$ \end{tabular} \\
        \hline
        $q=6$ & -- & -- & -- & --
              & --
              & \begin{tabular}{c|c|c} \cellcolor{lightgray}$1.6$ & $6.3$ & $73$ \\ \cellcolor{lightgray}$1.6$ & $2.7$ & $79$ \end{tabular} \\
        \hline
    \end{tabular}
\end{table}

\begin{figure}[htb]
    \centering
    \begin{subfigure}[t]{0.49\textwidth}
		\includegraphics[width=\textwidth]{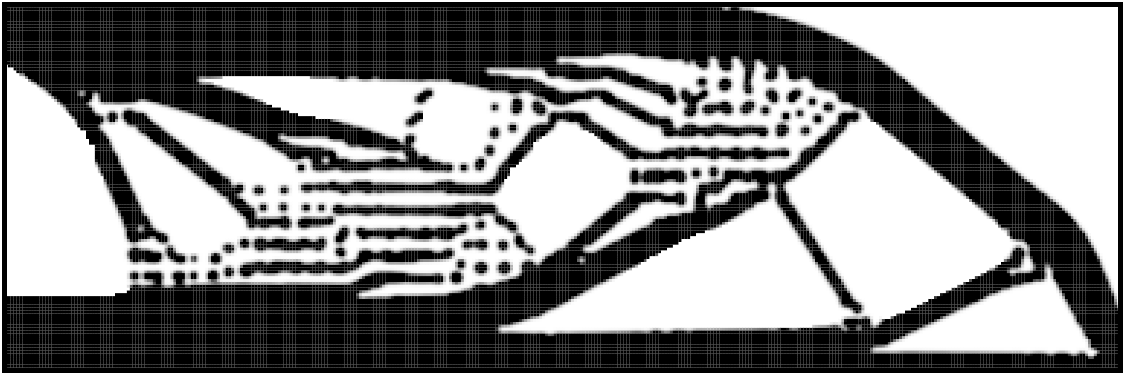}
		\caption{Artifacts in compliance problem}\label{fig:artifact1}
	\end{subfigure}
    \hfill
    \begin{subfigure}[t]{0.49\textwidth}
        \vspace*{-1.5cm}
		\includegraphics[width=\textwidth]{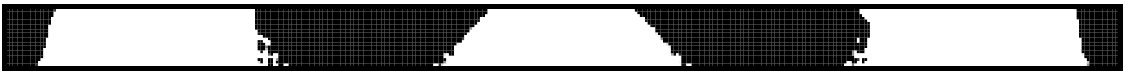}
        \caption{Artifacts in acoustic problem at $f=34.39$~Hz}\label{fig:artifact2}
	\end{subfigure}
    \caption{Numerical artifacts due to under-discretization in high-order multi-resolution topology optimization achieved with $q=1, n_v=4$, and a filter radius twice the voxel size}\label{fig:artifact}
\end{figure}

However, in contrast to the compliance problem, in which severe artifacts occur when under-discretizing~\cite{groen_higherorder_2017} (see \Cref{fig:artifact1}), such artifacts seem rare and localized in acoustic topology optimization. One of these rare examples, hand-selected from many other under-discretized optimizations that did not show these artifacts, is provided in \Cref{fig:artifact2}. \\

Due to the localization and rarity of these artifacts, the question arises of how far the designs obtained with under-discretizations are from useful designs. To answer this question, a second optimization is to be performed subsequent to the under-discretized optimization. Specifically, the first discretization is performed for 280 epochs, followed by a second for 20 epochs where the polynomial degree has been increased by two. Similar to before, the quantities of interest from \Cref{tab:mtoplabels} are evaluated for the adapted optimization scheme in \Cref{tab:mtop2}. Again, equal performances and improvements in comparison to the discretization $q=1, n_v=1$ are highlighted in gray. Thus, any normalized speed-up $\geq 1$, any cost increase $\leq 1.5$, and any design $\leq 65$~dB and $\leq 78$~dB (considering the same baseline as for \Cref{tab:mtop1}) is highlighted. With this modification, the advantages of a high-order multi-resolution discretization become more prominent. The discretization $q=2, n_v=4$ yields consistent results with a speed-up of ${\sim}5$ and is therefore employed throughout the remainder of this work. \\

\begin{table}[htb]
    \centering
    \caption{Normalized time, cost increase after discretization change, and corrected sound pressure level (see \Cref{tab:mtoplabels}) for different discretizations relying on $432\times 216$ voxels at $f=69.43$~Hz and $f=34.39$~Hz utilizing the proposed correction scheme with a second higher-order optimization}\label{tab:mtop2}
    \setlength{\tabcolsep}{4pt}
    \begin{tabular}{ccccccc}
              & $n_v=1$ & $n_v=2$ & $n_v=3$ & $n_v=4$ & $n_v=6$ & $n_v=8$ \\ 
        \hline
        $q=1$ & \begin{tabular}{c|c|c} $0.9$ & \cellcolor{lightgray}$1.2$ & \cellcolor{lightgray}$61$ \\ \cellcolor{lightgray}$1.0$ & \cellcolor{lightgray}$1.1$ & \cellcolor{lightgray}$75$ \end{tabular}
              & \begin{tabular}{c|c|c} \cellcolor{lightgray}$3.4$ & \cellcolor{lightgray}$1.1$ & $70$ \\ \cellcolor{lightgray}$4.0$ & \cellcolor{lightgray}$1.2$ & \cellcolor{lightgray}$76$ \end{tabular}
              & \begin{tabular}{c|c|c} \cellcolor{lightgray}$6.2$ & \cellcolor{lightgray}$1.0$ & $83$ \\ \cellcolor{lightgray}$5.4$ & \cellcolor{lightgray}$1.2$ & \cellcolor{lightgray}$76$ \end{tabular} 
              & \begin{tabular}{c|c|c} \cellcolor{lightgray}$9.8$ & \cellcolor{lightgray}$1.0$ & $83$ \\ \cellcolor{lightgray}$8.5$ & \cellcolor{lightgray}$1.1$ & \cellcolor{lightgray}$75$ \end{tabular} 
              & \begin{tabular}{c|c|c} \cellcolor{lightgray}$11.6$ & \cellcolor{lightgray}$1.0$ & $82$ \\ \cellcolor{lightgray}$12.2$ & \cellcolor{lightgray}$1.0$ & \cellcolor{lightgray}$75$ \end{tabular} 
              & \begin{tabular}{c|c|c} \cellcolor{lightgray}$15.2$ & \cellcolor{lightgray}$1.0$ & $82$ \\ \cellcolor{lightgray}$13.0$ & \cellcolor{lightgray}$1.0$ & $85$ \end{tabular} \\
        \hline 
        $q=2$ & --
              & \begin{tabular}{c|c|c} $0.9$ & \cellcolor{lightgray}$1.3$ & \cellcolor{lightgray}$61$ \\ \cellcolor{lightgray}$1.1$ & \cellcolor{lightgray}$1.3$ & \cellcolor{lightgray}$76$ \end{tabular}
              & \begin{tabular}{c|c|c} \cellcolor{lightgray}$2.5$ & \cellcolor{lightgray}$1.3$ & \cellcolor{lightgray}$61$ \\ \cellcolor{lightgray}$2.9$ & \cellcolor{lightgray}$1.1$ & \cellcolor{lightgray}$75$ \end{tabular}
              & \begin{tabular}{c|c|c} \cellcolor{lightgray}$4.6$ & \cellcolor{lightgray}$1.1$ & \cellcolor{lightgray}$61$ \\ \cellcolor{lightgray}$5.2$ & \cellcolor{lightgray}$1.1$ & \cellcolor{lightgray}$75$ \end{tabular}
              & \begin{tabular}{c|c|c} \cellcolor{lightgray}$8.1$ & $1.6$ & $67$ \\ \cellcolor{lightgray}$9.2$ & \cellcolor{lightgray}$1.3$ & \cellcolor{lightgray}$76$ \end{tabular}
              & \begin{tabular}{c|c|c} \cellcolor{lightgray}$11.9$ & \cellcolor{lightgray}$1.2$ & $69$ \\ \cellcolor{lightgray}$11.7$ & \cellcolor{lightgray}$1.0$ & \cellcolor{lightgray}$75$ \end{tabular} \\
        \hline 
        $q=3$ & --
              & --
              & \begin{tabular}{c|c|c} \cellcolor{lightgray}$1.2$ & \cellcolor{lightgray}$1.3$ & \cellcolor{lightgray}$62$ \\ \cellcolor{lightgray}$1.3$ & \cellcolor{lightgray}$1.3$ & \cellcolor{lightgray}$76$ \end{tabular}
              & \begin{tabular}{c|c|c} \cellcolor{lightgray}$2.2$ & \cellcolor{lightgray}$1.2$ &\cellcolor{lightgray}$61$ \\ \cellcolor{lightgray}$2.6$ & \cellcolor{lightgray}$1.2$ & \cellcolor{lightgray}$76$ \end{tabular}
              & \begin{tabular}{c|c|c} \cellcolor{lightgray}$4.9$ & \cellcolor{lightgray}$1.1$ & $68$ \\ \cellcolor{lightgray}$5.4$ & \cellcolor{lightgray}$1.1$ & \cellcolor{lightgray}$75$ \end{tabular}
              & \begin{tabular}{c|c|c} \cellcolor{lightgray}$8.1$ & $1.7$ & $67$ \\ \cellcolor{lightgray}$8.1$ & \cellcolor{lightgray}$1.0$ & \cellcolor{lightgray}$76$ \end{tabular} \\
        \hline
        $q=4$ & -- & --
              & --
              & \begin{tabular}{c|c|c} \cellcolor{lightgray}$1.1$ & \cellcolor{lightgray}$1.4$ & \cellcolor{lightgray}$62$ \\ \cellcolor{lightgray}$1.2$ & \cellcolor{lightgray}$1.0$ & \cellcolor{lightgray}$75$ \end{tabular}
              & \begin{tabular}{c|c|c} \cellcolor{lightgray}$2.4$ & \cellcolor{lightgray}$1.3$ & \cellcolor{lightgray}$65$ \\ \cellcolor{lightgray}$2.8$ & \cellcolor{lightgray}$1.2$ & \cellcolor{lightgray}$76$ \end{tabular}
              & \begin{tabular}{c|c|c} \cellcolor{lightgray}$4.8$ & \cellcolor{lightgray}$1.3$ & $68$ \\ \cellcolor{lightgray}$4.7$ & $1.9$ & \cellcolor{lightgray}$78$ \end{tabular} \\
              \hline
        $q=5$ & -- & -- & --
              & --
              & \begin{tabular}{c|c|c} \cellcolor{lightgray}$1.4$ & \cellcolor{lightgray}$1.3$ & \cellcolor{lightgray}$65$ \\ \cellcolor{lightgray}$1.6$ & $2.0$ & \cellcolor{lightgray}$78$ \end{tabular}
              & \begin{tabular}{c|c|c} \cellcolor{lightgray}$2.7$ & \cellcolor{lightgray}$1.4$ & $66$ \\ \cellcolor{lightgray}$2.7$ & \cellcolor{lightgray}$0.9$ & \cellcolor{lightgray}$75$ \end{tabular} \\
        \hline
        $q=6$ & -- & -- & -- & --
              & --
              & \begin{tabular}{c|c|c} \cellcolor{lightgray}$1.3$ & $3.1$ & $70$ \\ \cellcolor{lightgray}$1.5$ & \cellcolor{lightgray}$1.2$ & \cellcolor{lightgray}$76$ \end{tabular} \\
        \hline
    \end{tabular}
\end{table}

\begin{figure}[H]
\centering
    \begin{subfigure}[t]{0.49\textwidth}
		\includegraphics[width=\textwidth]{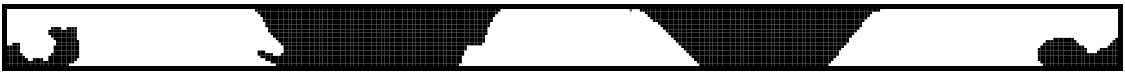}
		\caption{$q=2, n_v=4$ at $f=34.39$~Hz yielding $L_p=85.95$~dB in \textbf{198.53 s} after $280$ epochs}\label{fig:visualizationOfpcontinuation1}
	\end{subfigure}
    \hfill
    \begin{subfigure}[t]{0.49\textwidth}
		\includegraphics[width=\textwidth]{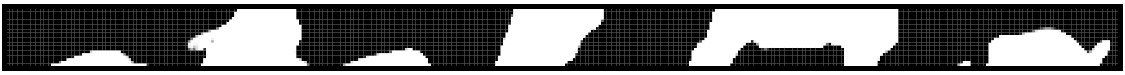}
		\caption{$q=2, n_v=4$ at $f=69.43$~Hz yielding $L_p=66.40$~dB in \textbf{200.35 s} after $280$ epochs}\label{fig:visualizationOfpcontinuation2}
	\end{subfigure}
    \\
    \begin{subfigure}[t]{0.49\textwidth}
		\includegraphics[width=\textwidth]{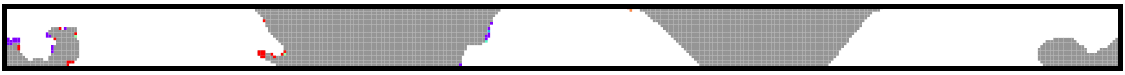}
		\caption{Adjustment through second optimization, where red is removing, gray is prior, and blue is adding material}\label{fig:visualizationOfpcontinuation3}
	\end{subfigure}
    \hfill
    \begin{subfigure}[t]{0.49\textwidth}
		\includegraphics[width=\textwidth]{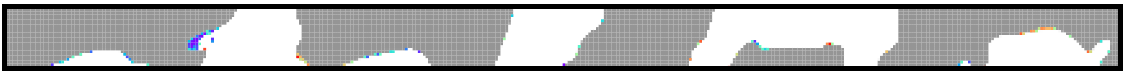}
		\caption{Adjustment through second optimization, where red is removing, gray is prior, and blue is adding material}\label{fig:visualizationOfpcontinuation4}
	\end{subfigure}
    \\
    \begin{subfigure}[t]{0.49\textwidth}
		\includegraphics[width=\textwidth]{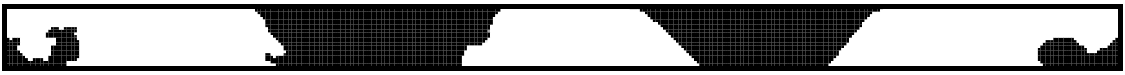}
		\caption{$q=4, n_v=4$ at $f=34.39$~Hz yielding $L_p=75.19$~dB in \textbf{83.65 s} with $20$ additional epochs}\label{fig:visualizationOfpcontinuation5}
	\end{subfigure}
    \hfill
    \begin{subfigure}[t]{0.49\textwidth}
		\includegraphics[width=\textwidth]{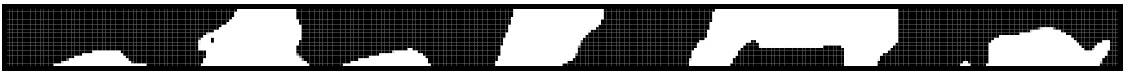}
		\caption{$q=4, n_v=4$ at $f=69.43$~Hz yielding $L_p=61.34$~dB in \textbf{81.00 s} with $20$ additional epochs}\label{fig:visualizationOfpcontinuation6}
	\end{subfigure}
    \\
    \begin{subfigure}[t]{0.49\textwidth}
		\includegraphics[width=\textwidth]{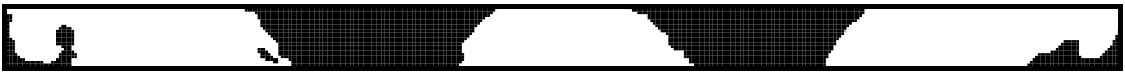}
		\caption{$q=4, n_v=4$ at $f=34.39$~Hz yielding $L_p=77.13$~dB in \textbf{1046 s} after $300$ epochs}\label{fig:visualizationOfpcontinuation7}
	\end{subfigure}
    \hfill
    \begin{subfigure}[t]{0.49\textwidth}
		\includegraphics[width=\textwidth]{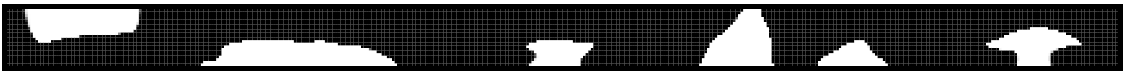}
		\caption{$q=4, n_v=4$ at $f=69.43$~Hz yielding $L_p=69.81$~dB in \textbf{1065 s} after $300$ epochs}\label{fig:visualizationOfpcontinuation8}
	\end{subfigure}
    \caption{Ensuring good optima despite under-discretization going from $n_v=4, q=2$ to $n_v=4, q=4$}\label{fig:visualizationOfpcontinuation}
\end{figure}

\begin{figure}[htb]
    \centering
    \begin{subfigure}[t]{0.49\textwidth}
        \includegraphics[width=\textwidth]{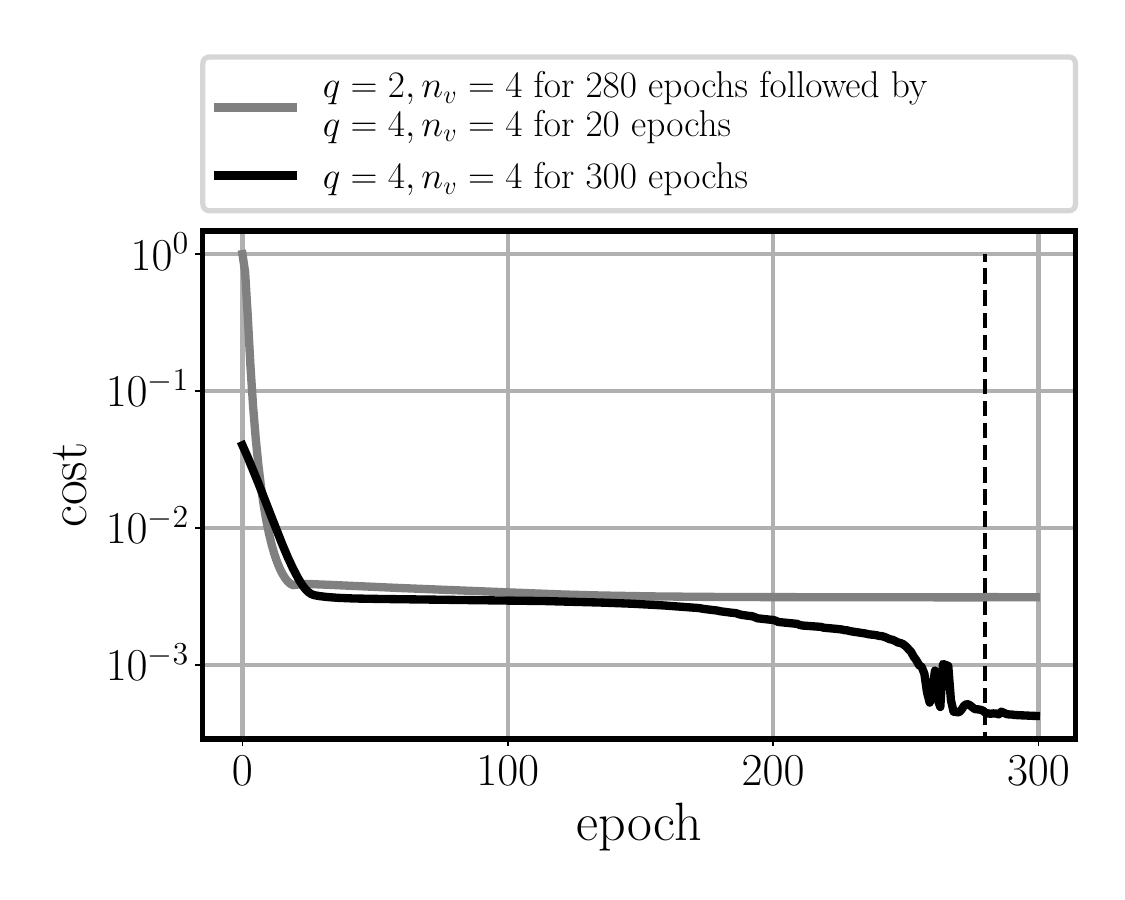}
		\caption{$f=34.39$~Hz}
    \end{subfigure}
    \hfill
    \begin{subfigure}[t]{0.49\textwidth}
        \includegraphics[width=\textwidth]{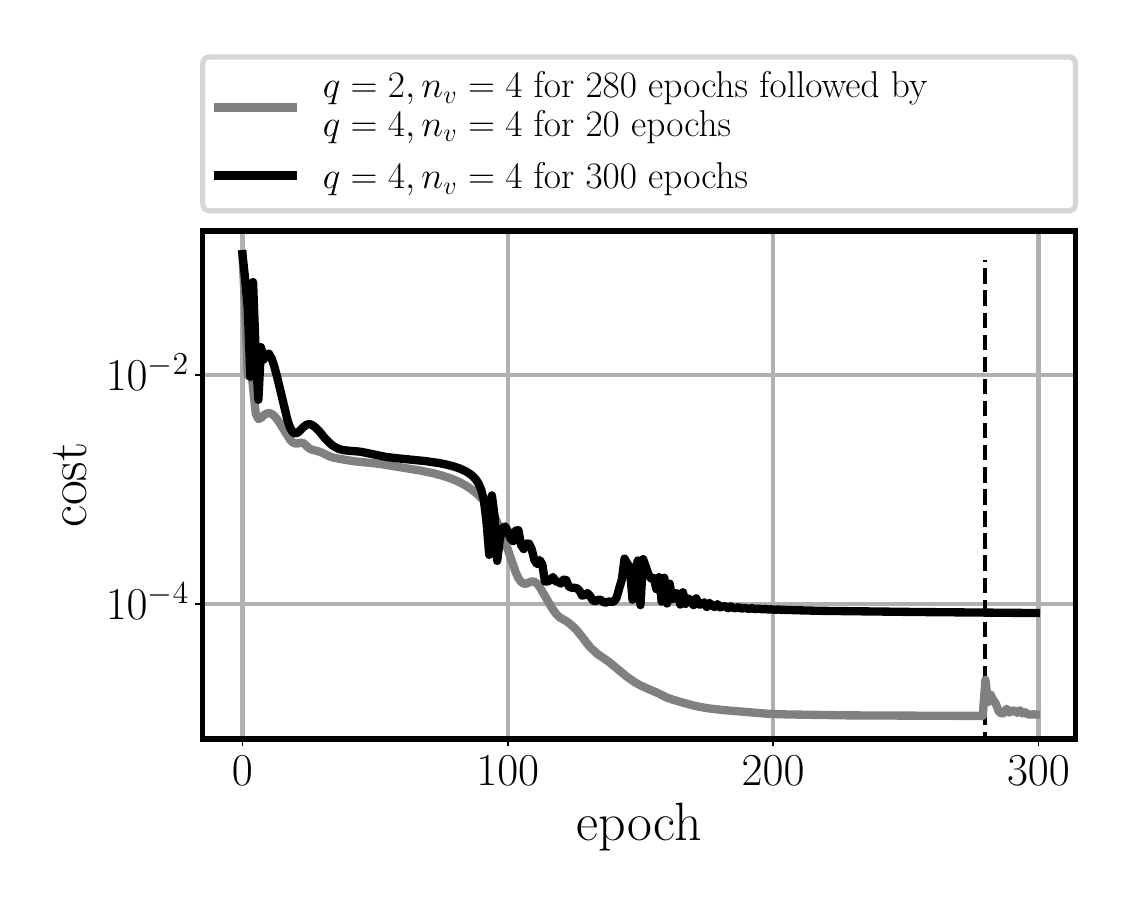}
		\caption{$f=69.43$~Hz}
    \end{subfigure}
    \caption{Comparison of cost histories in high-order multi-resolution acoustic topology optimization. Black indicates a conventional optimization obtained after 300 epochs with $q=4, n_v=4$, while gray showcases the result using the proposed two-step optimization scheme, where a switch from $q=2, n_v=4$ to $q=4, n_v=4$ is made at epoch 280.}\label{fig:costHistoryPContinuation}
\end{figure}

\Cref{fig:visualizationOfpcontinuation} provides a visualization of the proposed two-step optimization scheme for $f=34.39$~Hz and $f=69.43$~Hz using $q=2, n_v=4$. The change introduced by the second optimization is visualized in \Cref{fig:visualizationOfpcontinuation3,fig:visualizationOfpcontinuation4}, where red indicates material removal, while blue shows additional material. Thus, only minor changes are needed. The cost histories for the two cases are illustrated in \Cref{fig:costHistoryPContinuation}, where the dashed line indicates the point at which the discretization is changed from $q=2, n_v=4$ to $q=4, n_v=4$. Only a few iterations are needed to amend the errors introduced by the under-discretization.

\section{Two-Dimensional Optimization Benchmarks}\label{appendix:optimization}

\begin{figure}[H]
    \centering
    \begin{subfigure}[t]{0.32\textwidth}
		\includegraphics[width=\textwidth]{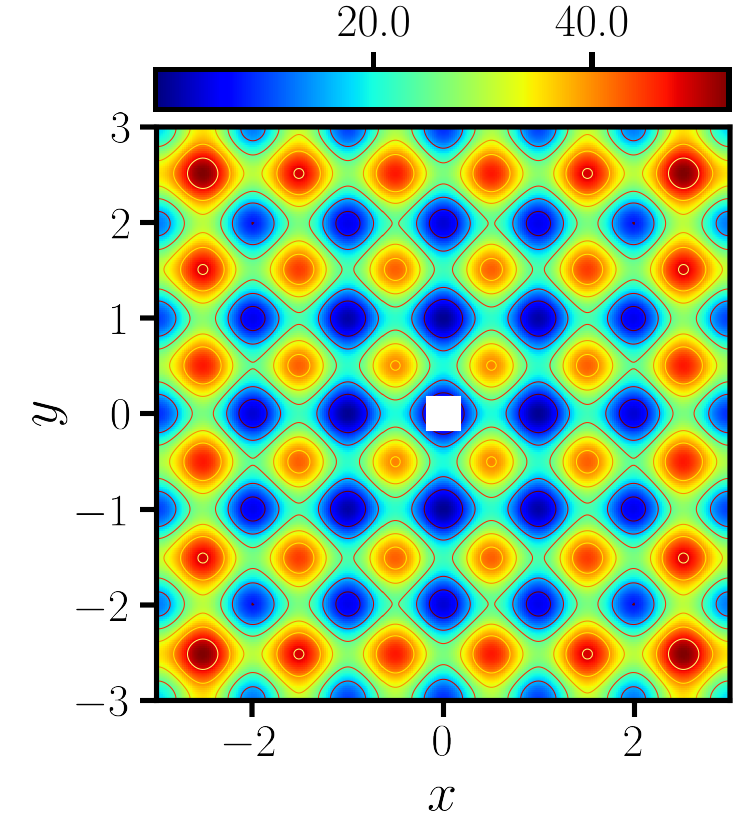}
		\caption{Rastrigrin function}
	\end{subfigure}
    \hfill
    \begin{subfigure}[t]{0.32\textwidth}
		\includegraphics[width=\textwidth]{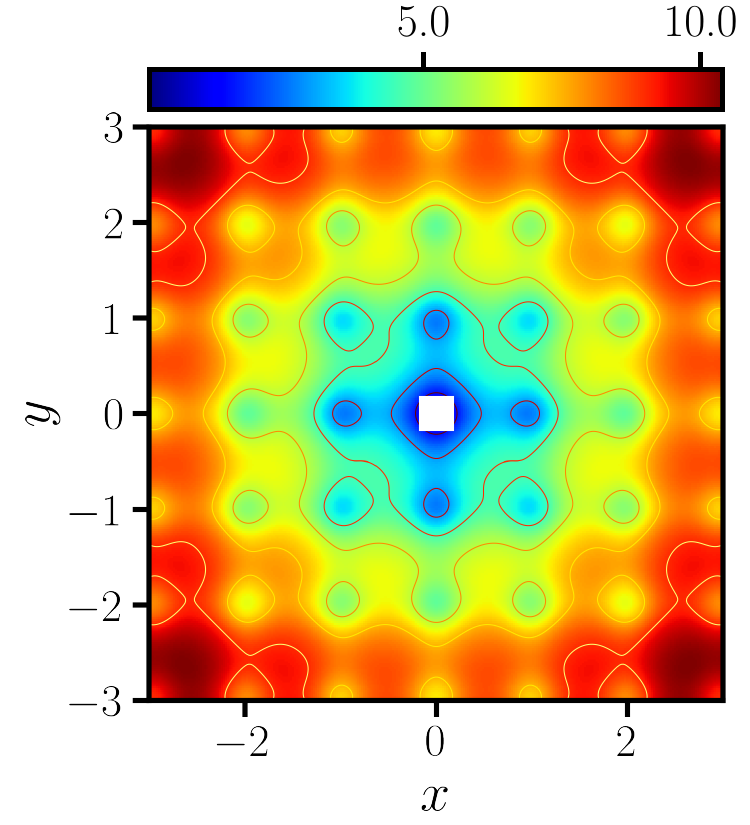}
		\caption{Ackley function}
	\end{subfigure}
    \hfill
    \begin{subfigure}[t]{0.32\textwidth}
		\includegraphics[width=\textwidth]{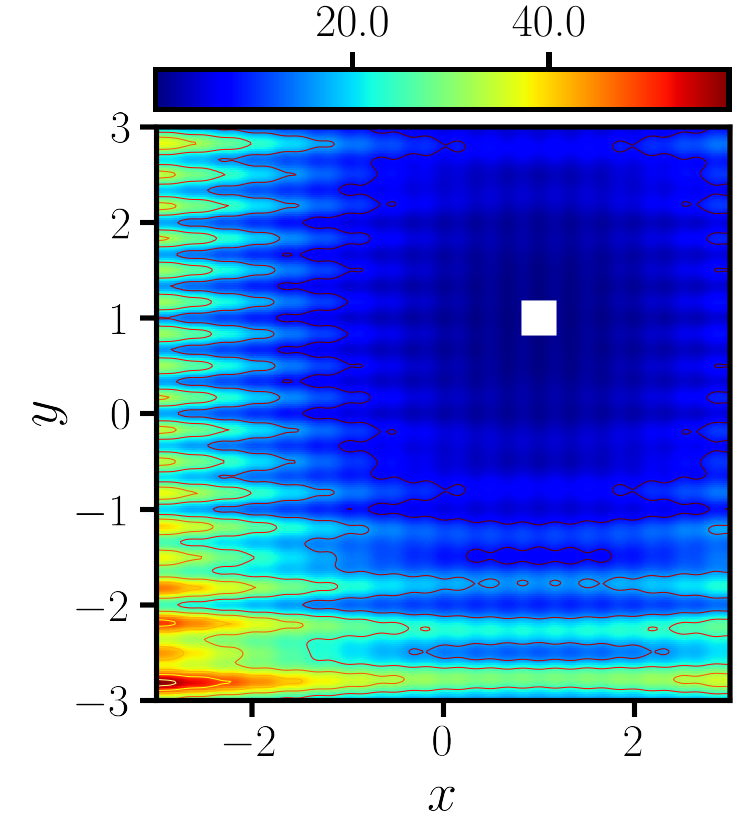}
		\caption{Levy function}
	\end{subfigure}
    \caption{Two-dimensional optimization landscapes with the global minima indicated by white squares}\label{fig:optimizationlandscapes}
\end{figure}

Aside the Rosenbrock function from~\Cref{eq:rosenbrock}, the Rastrigrin function
\begin{equation}
    f(x, y) = 20 + x^2 - 10 \cos(2\pi x) + y^2 - 10 \cos(2 \pi x),
\end{equation}
the Ackley function
\begin{equation}
    f(x, y) = - 20 e^{-0.2\sqrt{0.5 (x^2 + y^2)}} - e^{0.5(\cos(2\pi x)+\cos(2 \pi y)} + e + 20,
\end{equation}
and the Lévi function
\begin{equation}
    f(x, y) = \sin^2(3\pi x) + (x-1)^2 (1+\sin^2(3\pi y)) + (y-1)^2(1+\sin^2(2\pi y))
\end{equation}
are considered. The corresponding optimization landscapes are depicted in \Cref{fig:optimizationlandscapes} with the global minima indicated by white squares.

\section{Natural Frequency Estimation}\label{appendix:frequencies}
For the natural frequency estimation of the undisturbed domain, the $nm$th wavefield $\hat{p}(x,y)$ with the form of
\begin{equation}
    \hat{p} = \hat{p}_0 \cos\left(\frac{n\pi x}{a}\right) \cos\left(\frac{m \pi y}{b}\right)\label{eq:frequencyAnsatz}
\end{equation}
is assumed, which fulfills the homogeneous Neumann boundary conditions. Inserting \Cref{eq:frequencyAnsatz} into the homogeneous version of the Helmholtz equation from \Cref{eq:helmholtz} and assuming a homogeneous material distribution of $\tilde{\kappa}=1$ and $\tilde{\rho}=1$, i.e., assuming air everywhere, yields an expression for the $nm$th frequency $f_{nm}$.
\begin{equation}
    f_{nm} = \frac{\omega_{nm}}{2\pi} = \frac{\tilde{\omega}_{nm} c_1}{2\pi}  \frac{c_1}{2} \sqrt{\left(\frac{n}{L_x}\right)^2 + \left(\frac{m}{L_y}\right)^2}
\end{equation}
The wave speed of air $c_1$ is defined with the material parameters of air.
\begin{equation}
    c_1 = \sqrt{\kappa_1 / \rho_1}
\end{equation}

\section{Neural Network Architecture}\label{appendix:NN}
The employed NN is a U-net~\cite{ronneberger_u-net_2015} with three downsampling and three upsampling blocks. It consists of $148\,806$ trainable parameters and is summarized by \Cref{tab:2dNN}. The input to the NN is always the cost function gradient with respect to the indicator $\partial C/\partial \boldsymbol{\zeta}$, assuming $\boldsymbol{\zeta}=0$ in the entire domain $\Omega$. This input choice was first proposed in~\cite{singh_accelerating_2024} in the context of full waveform inversion. For further details, the interested reader is referred to the code made available in~\cite{herrmann_neural_2024}.

\begin{table}[H]
	\caption{Two-dimensional U-net~\cite{ronneberger_u-net_2015} constructed with convolutional NNs for the prediction of the filtered and projected indicator $\bar{\tilde{\boldsymbol{\zeta}}}$. The total number of trainable parameters is $148\,806$.} \label{tab:2dNN}
	\centering
	\begin{tabular}{lll}
		\hline
		\textbf{layer}                          & \textbf{shape after layer}        & \textbf{learnable parameters} \\ \hline
        \textbf{input}                          & $1\times 432\times 24$    & 0            \\ \hline \hline

        \textbf{downsampling} & & 
        \\ \hline \hline 
        max pooling                    & $1\times 216\times 12$    & 0                  \\ \hline
		batch norm                     & $1\times 216\times 12$    & $2$          \\ \hline
		convolution \& leaky ReLU   & $12\times 216\times 12$   & $312$          \\ \hline \hline
        max pooling                    & $12\times 108\times 6$    & 0                  \\ \hline
		batch norm                     & $12\times 108\times 6$    & $24$          \\ \hline
		convolution \& leaky ReLU   & $24\times 108\times 6$   & $7\,224$          \\ \hline \hline
        max pooling                    & $24\times 54\times 3$    & 0                  \\ \hline
		batch norm                     & $24\times 54\times 3$    & $48$          \\ \hline
		convolution    & $48\times 54\times 3$    & $28\,848$          \\
        \hline \hline

        \textbf{bottleneck} & & \\
        \hline \hline 
        batch norm                     & $48\times 54\times 3$    & $96$          \\ \hline
        convolution \& leaky ReLU   & $48\times 54\times 3$    & $57\,648$     \\ 
        \hline \hline
        
        \textbf{upsampling} & & \\
        \hline \hline
        upsample \& skip connection    & $72\times 108\times 6$   & $0$         \\ \hline
        batch norm                     & $72\times 108\times 6$   & $144$       \\ \hline
        convolution \& leaky ReLU       & $24\times 108\times 6$   & $43\,224$   \\ 
        \hline \hline
        upsample \& skip connection    & $36\times 216\times 12$   & $0$         \\ \hline
        batch norm                     & $36\times 216\times 12$   & $72$       \\ \hline
        convolution \& leaky ReLU       & $12\times 216\times 12$   & $10\,812$   \\ 
        \hline \hline
        upsample \& skip connection    & $13\times 432\times 24$   & $0$         \\ \hline
        batch norm                     & $13\times 432\times 24$   & $26$       \\ \hline
        convolution \& Sigmoid          & $1\times 432\times 24$    & $326$   \\ 
        \hline \hline

        \textbf{filtering}             & $1\times 432\times 24$    & $0$ \\
        \hline \hline
        \textbf{projection}            & $1\times 432\times 24$    & $0$ \\ 
        \hline 
    \end{tabular}
\end{table}

\section{Evaluation of Restarting Scheme}\label{appendix:restarting}
The sound pressure levels achieved with the optimized designs using the restarting scheme and the NN ansatz are summarized in \Cref{tab:RestartNN}. Comparing the results from the transfer learning scheme in \Cref{tab:NNansatz4}, transfer learning arises as the slightly superior approach. In three of the eight cases, the restarting scheme produces better designs, indicated in dark gray in \Cref{tab:RestartNN}. However, the transfer learning scheme bears the additional advantage of only requiring a single pretraining for many subsequent optimizations, while the restarting scheme needs one initial optimization for each run.

\begin{table}[H]
    \centering
    \caption{Optimized sound pressure levels with \textbf{NN ansatz} based on the \textbf{restarting scheme}. The first sound pressure level is from the initial optimization (i.e., before the restart), while the latter three are as described in \Cref{tab:NNlabels}. The gray highlights indicate that the reference from \Cref{tab:Reference} is outperformed. Additionally, the improvement over the reference designs from \Cref{tab:Reference} is quantified, where the dark gray highlights indicate better designs than those obtained with transfer learning in \Cref{tab:NNansatz4}.}\label{tab:RestartNN}
    \setlength{\tabcolsep}{4pt}
    \begin{tabular}{cccc}
        $f$ & \begin{tabular}{c}sound pressure level\end{tabular} & improvement & \begin{tabular}{c}epochs\end{tabular} \\ 
        \hline
        $21.33$ Hz & \begin{tabular}{c||c|c|c} 78.5~dB & 79.2~dB & \cellcolor{lightgray}73.5~dB & \cellcolor{lightgray}78.3~dB \end{tabular} & 3.9~dB & 150 \& 300 \& \textbf{21}\\
        \hline
        $34.39$ Hz & \begin{tabular}{c||c|c|c} 75.2~dB & \cellcolor{lightgray}72.5~dB & \cellcolor{lightgray}73.5~dB & \cellcolor{lightgray}74.5~dB \end{tabular} & 0.4~dB & 150 \& 300 \& \textbf{1}\\
        \hline
        $57.22$ Hz & \begin{tabular}{c||c|c|c} 61.6~dB & \cellcolor{lightgray}51.6~dB & \cellcolor{lightgray}52.5~dB & \cellcolor{lightgray}56.1~dB\end{tabular} & \cellcolor{gray}13.0~dB & 150 \& 300 \& \textbf{21}\\
        \hline
        $69.43$ Hz & \begin{tabular}{c||c|c|c} 60.4~dB & \cellcolor{lightgray}58.8~dB & \cellcolor{lightgray}59.9~dB & 62.3~dB \end{tabular} & \cellcolor{gray}-1.8~dB & 150 \& 280 \& \textbf{11}\\
        \hline
        $76.30$ Hz & \begin{tabular}{c||c|c|c} 78.3~dB & \cellcolor{lightgray}71.6~dB & \cellcolor{lightgray}73.3~dB & \cellcolor{lightgray}75.2~dB \end{tabular} & 4.9~dB & 150 \& 300 \& \textbf{1}\\
        \hline
        $95.37$ Hz & \begin{tabular}{c||c|c|c} 70.0~dB & \cellcolor{lightgray}63.2~dB & \cellcolor{lightgray}66.6~dB & \cellcolor{lightgray}78.3~dB \end{tabular} & 4.6~dB & 150 \& 300 \& \textbf{31}\\
        \hline
        $141.78$ Hz & \begin{tabular}{c||c|c|c} 79.6~dB & \cellcolor{lightgray}76.8~dB & \cellcolor{lightgray}77.7~dB & \cellcolor{lightgray}79.2~dB \end{tabular} & 1.3~dB & 150 \& 300 \& \textbf{1}\\  
        \hline
        $166.56$ Hz & \begin{tabular}{c||c|c|c} 78.6~dB & \cellcolor{lightgray}70.0~dB & \cellcolor{lightgray}70.1~dB & \cellcolor{lightgray}71.5~dB \end{tabular} & \cellcolor{gray}6.3~dB & 150 \& 300 \& \textbf{11}\\
        \hline
    \end{tabular}
\end{table}

\section{Additional Statistical Evaluations}\label{appendix:statistics}
\begin{figure}[htb]
	\centering
    \begin{subfigure}[t]{0.49\textwidth}
		\includegraphics[width=\textwidth]{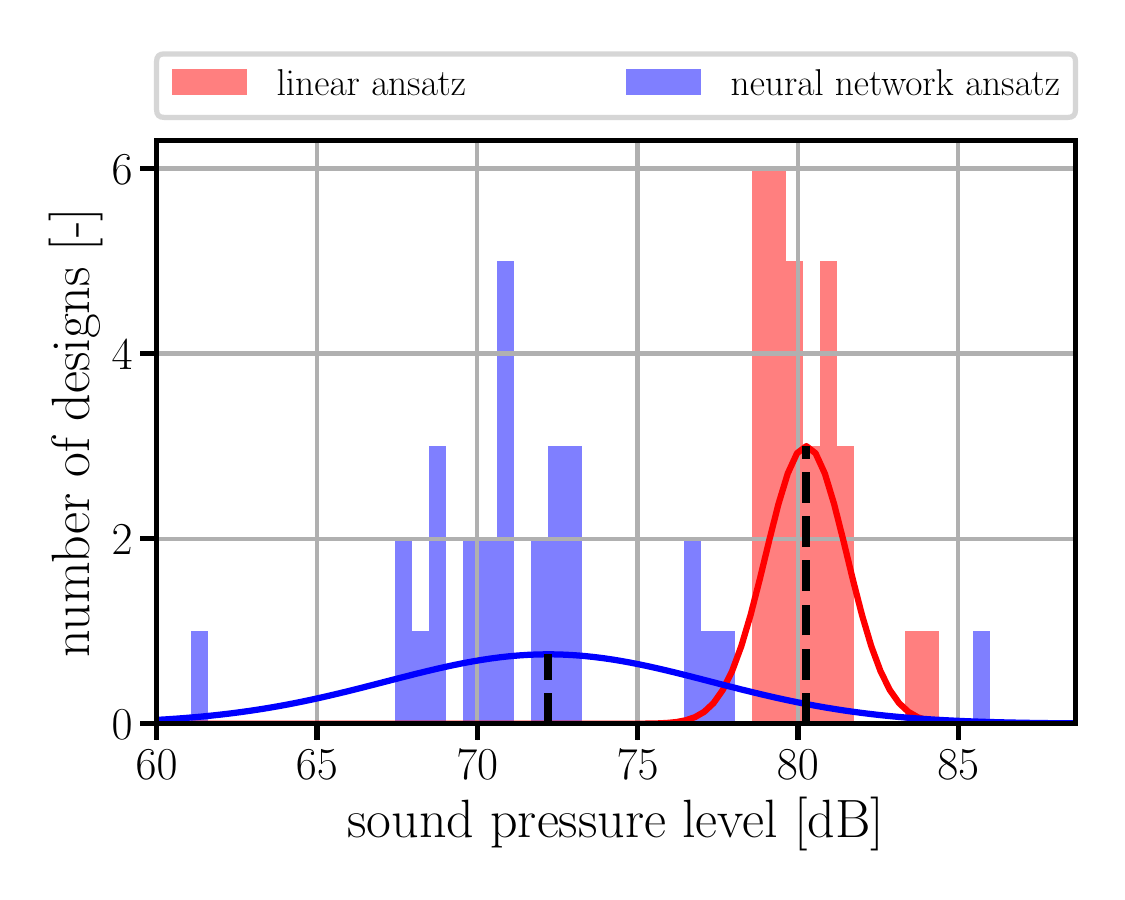}
		\caption{$f=34.39$~Hz}\label{fig:histogramspp3}
	\end{subfigure}
    \hfill
    \begin{subfigure}[t]{0.49\textwidth}
		\includegraphics[width=\textwidth]{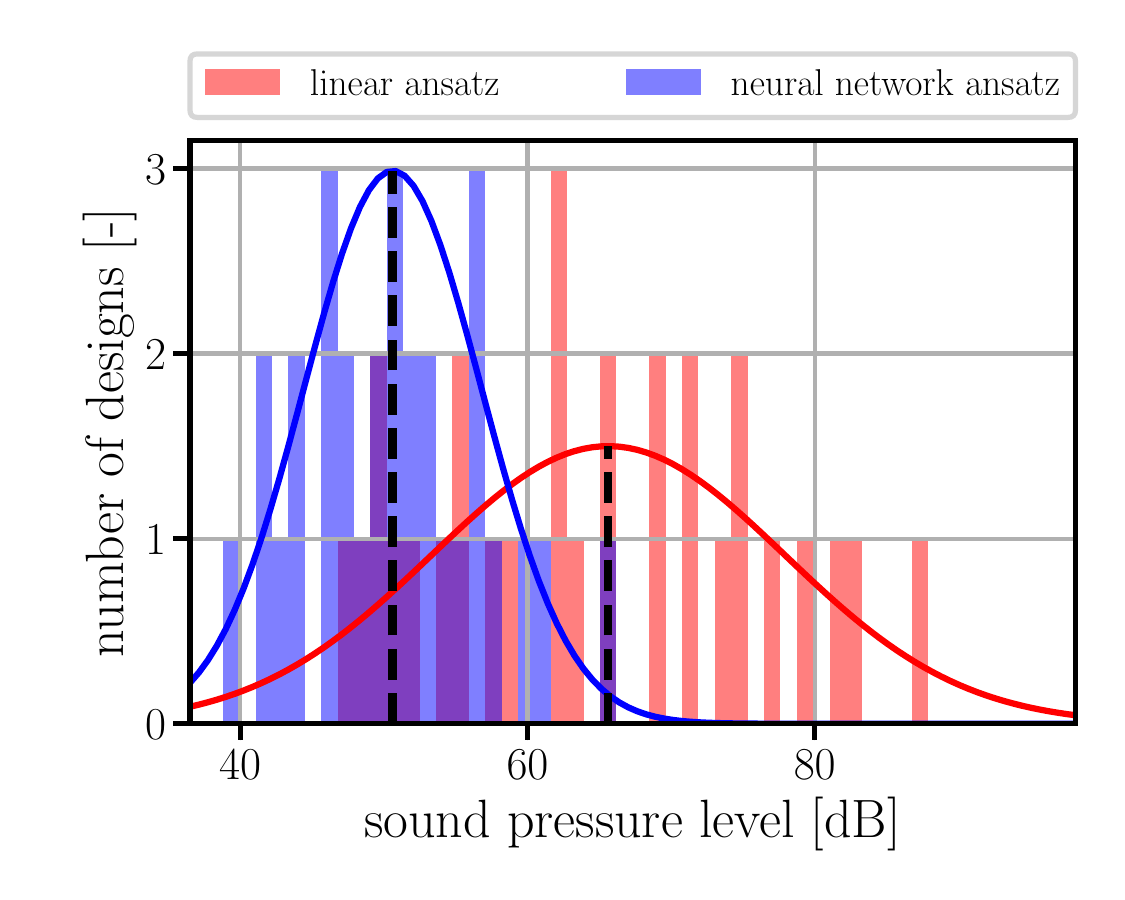}
		\caption{$f=69.43$~Hz}\label{fig:histogramspp4}
	\end{subfigure}
    \\
    \begin{subfigure}[t]{0.49\textwidth}
		\includegraphics[width=\textwidth]{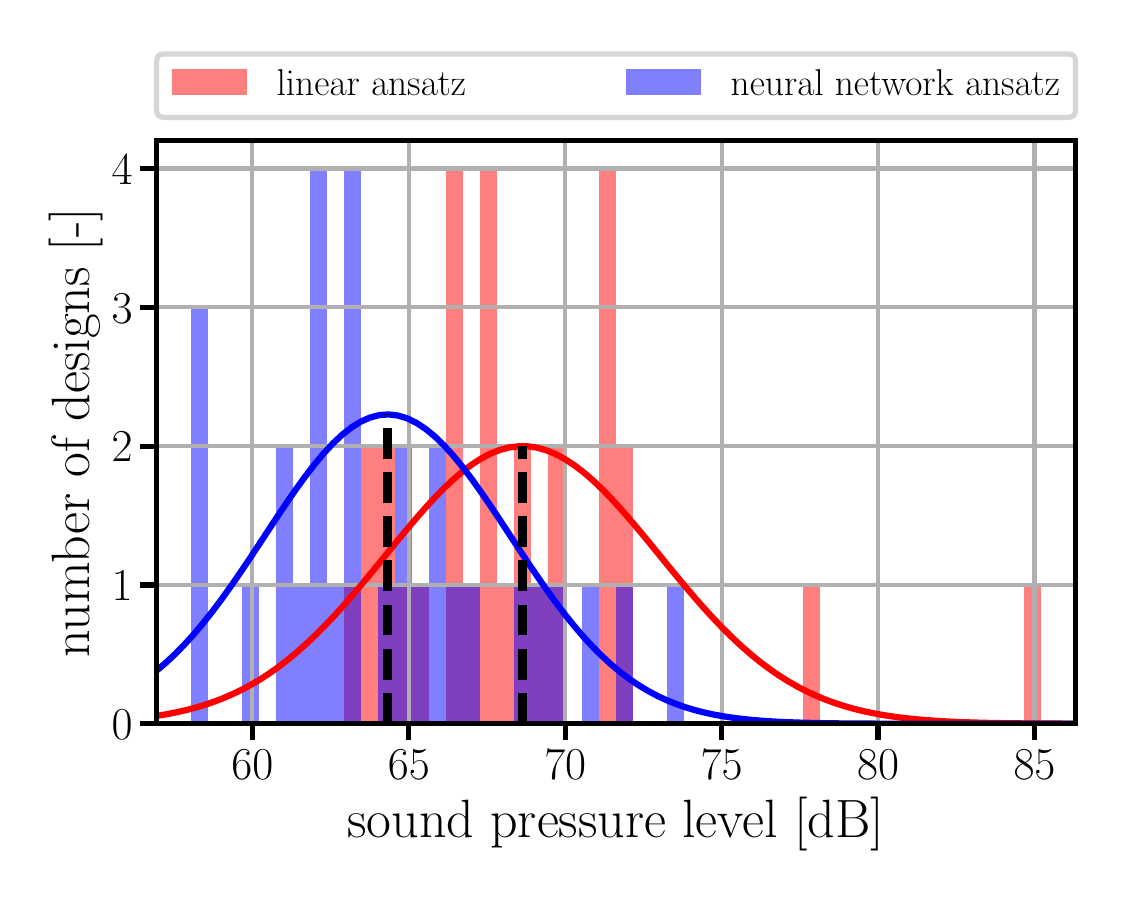}
		\caption{$f=141.78$~Hz}\label{fig:histogramspp1}
	\end{subfigure}
    \hfill
    \begin{subfigure}[t]{0.49\textwidth}
		\includegraphics[width=\textwidth]{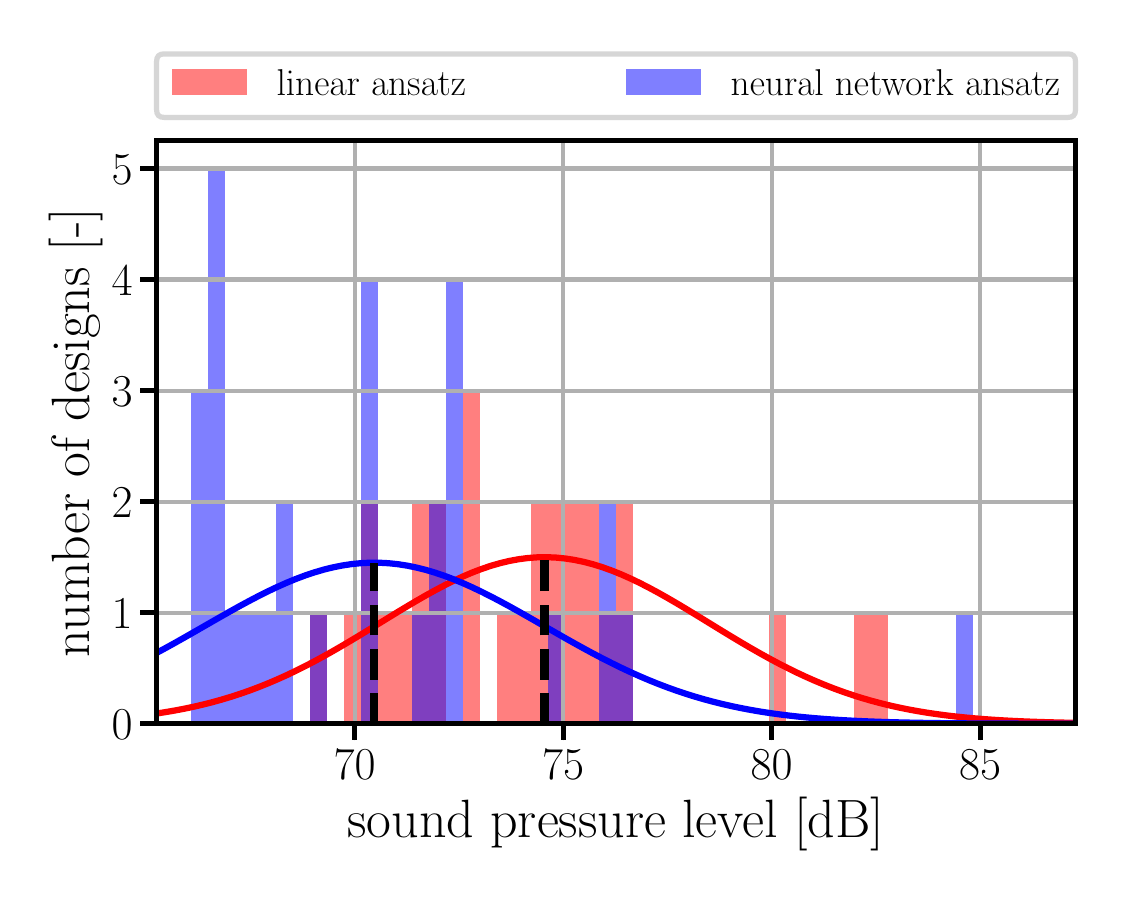}
		\caption{$f=166.56$~Hz}\label{fig:histogramspp2}
	\end{subfigure} 
    \caption{Statistical evaluation of the local optima quality with 30 samples for the linear (with homogeneous initial guesses) and the pretrained NN ansatz --- relying on different initializations. In the linear case, we used 30 homogeneous initial guesses equally spaced between 0 and 1, while the NN relies on 30 different pretrainings.}\label{fig:histogramsapp}
\end{figure}

For the sake of completeness, the remaining histograms from \Cref{ssec:statisticalEvaluation} are provided in \Cref{fig:histogramsapp} for frequencies $f=34.39$~Hz, $f=69.43$~Hz, $f=141.78$~Hz, and $f=166.56$~Hz.

\newpage

\bibliographystyle{unsrtnat}

\setlength{\bibsep}{3pt}
\setlength{\bibhang}{0.75cm}{\fontsize{9}{9}\selectfont\bibliography{2024TopOpt}}

\end{document}

%% file: configuration.tex
\usepackage[table]{xcolor}
\definecolor{lightblue}{rgb}{0,0.5,1.0}
\definecolor{linkblue}{rgb}{0,0.1,0.6}
\definecolor{citegreen}{rgb}{0,0.4,0.0}
\definecolor{linkred}{rgb}{0.8,0,0.005}
\definecolor{mailviolet}{rgb}{0.3,0,0.35}
\definecolor{tumblue}{rgb}{0,0.396,0.741}
\definecolor{darkgreen}{rgb}{0,0.4,0} 
\definecolor{darkbrown}{rgb}{0.5, 0.396, 0.09}
\usepackage[hyperindex,colorlinks=true,linkcolor=linkblue,citecolor=citegreen,urlcolor=mailviolet,filecolor=linkred]{hyperref}

\usepackage{caption, subcaption}
\usepackage{fancyhdr}

\usepackage{siunitx}
\usepackage[square,comma,nonamebreak,compress,numbers]{natbib}

\usepackage{soulpos}
\usepackage{ulem}
\usepackage{authblk} 
\usepackage{float}

\usepackage[inkscapearea=page]{svg}
\usepackage{amsmath,amsfonts,amssymb,amsthm}
\usepackage[noabbrev]{cleveref}
\usepackage{upgreek}
\usepackage{adjustbox}
\usepackage{color}
\usepackage{enumitem}

\usepackage{pgfplots}
\usepackage{pgfplotstable}
\usepackage{filecontents}
\usepackage{tikz}
\usepackage{tikzscale}
\usepackage{tikz-3dplot}

\usetikzlibrary{spy}
\usetikzlibrary{intersections}
\usetikzlibrary{arrows,shapes}
\usetikzlibrary{spy}
\usetikzlibrary{backgrounds}  
\usetikzlibrary{decorations}
\usetikzlibrary{decorations.markings}
\usetikzlibrary{positioning}
\usetikzlibrary{patterns}
\usetikzlibrary{calc} 
\usetikzlibrary{quotes} 
\usetikzlibrary{external} 
\usetikzlibrary{matrix}

\pgfplotscreateplotcyclelist{customCycleList}
{%
  {blue, mark=o},
  {red, mark=square},
  {darkgreen, mark=diamond},
  {cyan, mark=diamond},
  {darkbrown, mark=triangle*},
  {black, mark=pentagon},
}

\pgfplotsset{every axis/.append style= {
    cycle list name=customCycleList,
}}


%% file: 00_frontpage.tex
\title{Neural Networks for Generating Better Local Optima in Topology Optimization}

\author[1]{Leon Herrmann\thanks{\href{mailto:leon.herrmann@tum.de}{\texttt{leon.herrmann@tum.de}},
    Corresponding author}}
\author[2]{Ole Sigmund}
\author[1]{Viola Muning Li}
\author[3]{Christian Vogl}
\author[4]{Stefan Kollmannsberger}

\affil[1]{Chair of Computational Modeling and Simulation, Technical University of Munich, School of Engineering and Design, Arcisstraße 21, Munich, 80\,333, Germany}

\affil[2]{Department of Civil and Mechanical Engineering, Technical University of Denmark. Koppels Allé, B.404, 2800 Kgs. Lyngby, Denmark}

\affil[3]{BMW Group, Knorrstraße 147, 80937 München}

\affil[4]{Chair of Data Science in Engineering, Bauhaus-Universit\"at Weimar, Coudraystraße 13 b, 99423, Weimar, Germany}

\newcommand{\publicationDate}{\today}

\date{}
\fancypagestyle{plain}{%
\fancyhf{}%
\fancyfoot[L]{
\vspace{-0.0cm} 
\footnotesize{Preprint submitted to Structural and Multidisciplinary Optimization.  \hfill 
\publicationDate
\\
}} }